\documentclass[12pt]{article}
\usepackage{fullpage}
\usepackage{xcolor}
\usepackage{amsmath,amssymb,mathtools,bm}
\usepackage{graphicx, color}
\usepackage{url}
\usepackage[hyperfootnotes=false]{hyperref}

\newlength{\abstractwidth}
\renewcommand{\title}[1]{\vbox{\center\bf{\Large{#1}}}\vspace{5mm}}
\renewcommand{\author}[1]{\vbox{\center#1}\vspace{5mm}}
\newcommand{\address}[1]{\vbox{\center\em#1}}
\newcommand{\email}[1]{\vbox{\center\tt#1}\vspace{5mm}}

\newcommand{\be}{\begin{equation}}
\newcommand{\ee}{\end{equation}}

\newcommand{\E}{\mathbb{E}}
\newcommand{\hide}[1]{}

\numberwithin{equation}{section}

\begin{document}

\hypersetup{pageanchor=false} %
\begin{titlepage}
\rightline{MIT-CTP/5625}
\begin{center}
\hfill \\

\title{Feature Learning and Generalization in Deep Networks with Orthogonal Weights}

\author{Hannah Day,$^{ab\, \star}$, Yonatan Kahn,$^{ab\, \star}$, and Daniel A. Roberts$^{cd}$}

\address{$^{a}$ Department of Physics, University of Illinois Urbana-Champaign, \\
Urbana, IL 61801, USA

\vspace{10pt}

$^{b}$ Center for Artificial Intelligence Innovation, University of Illinois Urbana-Champaign, \\
Urbana, IL 61801, USA

\vspace{10pt}

$^{c}$ Center for Theoretical Physics {\it and} \\  Department of Physics, Massachusetts Institute of Technology \\ Cambridge, Massachusetts 02139, USA

\vspace{10pt}

$^{d}$ Sequoia Capital, Menlo Park, CA 94025, USA}

\email{hjday2@illinois.edu, yfkahn@illinois.edu, drob@mit.edu}

\end{center}

\begin{abstract}
Fully-connected deep neural networks with weights initialized from independent Gaussian distributions can be tuned to criticality, which prevents the exponential growth or decay of signals propagating through the network. However, such networks still exhibit fluctuations that grow linearly with the depth of the network, which may impair the training of networks with width comparable to depth. We show analytically that rectangular networks with tanh activations and weights initialized from the ensemble of orthogonal matrices have corresponding preactivation fluctuations which are independent of depth, to leading order in inverse width. Moreover, we demonstrate numerically that, at initialization, all correlators involving the neural tangent kernel (NTK) and its descendants at leading order in inverse width -- which govern the evolution of observables during training -- saturate at a depth of $\sim 20$, rather than growing without bound as in the case of Gaussian initializations. We speculate that this structure preserves finite-width feature learning while reducing overall noise, thus improving both generalization and training speed in deep networks with depth comparable to width. We provide some experimental justification by relating empirical measurements of the NTK to the superior performance of deep nonlinear orthogonal networks trained under full-batch gradient descent on the MNIST and CIFAR-10 classification tasks.
\end{abstract}
\let\thefootnote\relax\footnotetext{$^{\star}$ Equal contribution.}

\end{titlepage}
\hypersetup{pageanchor=true} %

\section{Introduction}

The initialization distributions for neural network parameters are key hyperparameters that can affect training and performance. A large body of work has focused on how \emph{critical initialization hyperparameters} can mitigate the exploding and vanishing gradient problem by preventing the exponential growth or decay of preactivations~\cite{poole2016exponential,schoenholz2016deep,raghu2017expressive,hanin2018neural,hanin2018start,hanin2019finite,park2019effect,yaida2020non,Roberts:2021fes,yang2020feature,yaida2022meta,dinan2023effective}. Modern developments have given rise to optimizers such as \texttt{Adam} \cite{kingma2014adam} and \texttt{AdamW}~\cite{loshchilov2017decoupled} that are relatively insensitive to initializations, as well as techniques like \texttt{LayerNorm}~\cite{ba2016layer} that explicitly enforce order-one values for propagating signals, guaranteeing critical initializations when implemented with residual connections \cite{doshi2021critical} (see also Ref.~\cite{he2022autoinit}). However, the minimal model of a wide-but-finite multi-layer perceptron (MLP) trained with full-batch gradient descent (GD) provides a theoretically-tractable playground to study the effect of initializations on the statistics and dynamics of ensembles of networks.

In such a setup, studied in detail in \cite{Roberts:2021fes} for the case of weights and biases drawn i.i.d.\ from zero-mean Gaussian distributions, MLPs of depth $L$ and width $n$ tuned to criticality still exhibit fluctuations which scale as $L/n$, characterized by the connected 4-point correlator of preactivations $V \sim \mathbb{E}[z^4]_{\rm conn.}$, where the expectation is taken over an ensemble of networks. If $L/n \sim 1$, these fluctuations grow to order-one sizes, potentially inhibiting the efficacy of training. On the other hand, finite-width effects are crucial for feature learning, as encapsulated by the evolution of the neural tangent kernel (NTK) $\widehat{H}$ which governs GD dynamics~\cite{jacot2018neural,arora2019exact,lee2019wide}. Correlators of the NTK fluctuations and derivatives of the NTK also scale like $L/n$. The competition between the ``good'' fluctuations which allow feature learning and the ``bad'' fluctuations which simply inject noise leads Ref.~\cite{Roberts:2021fes} to speculate that there exists an optimal aspect ratio $L/n \ll 1$ for Gaussian initializations, which can be calculated using techniques from information theory.

In this paper, we provide both analytical and numerical evidence that initializing the weights of an MLP from a distribution of orthogonal matrices yields drastically different behavior than Gaussian initializations for the same network architectures. In particular, orthogonal initializations yield correlators which saturate at moderate depth, rather than continuing to scale linearly with depth. Our study builds on previous work in \cite{saxe2013exact, pennington2017resurrecting, pennington2018emergence} which demonstrated the depth independence of the singular value variance of the input-output Jacobian for orthogonal networks at criticality, but only when paired with linear or tanh activation functions. Subsequent work confirmed that orthogonal initializations can also improve training in deep convolutional neural networks~\cite{xiao2018dynamical}, have provably superior training dynamics to Gaussian initializations in deep linear MLPs~\cite{hu2020provable}, and can achieve improved training in the $L/n \sim 1$ regime~\cite{huang2021neural}. Our work goes beyond these studies by showing the depth independence of the normalized single-input 4-point correlator, which measures a different facet of network fluctuations than the input-output Jacobian, while working entirely in the finite-width regime to leading order in $1/n$. Furthermore, we study for the first time the correlators of the NTK and its descendants, which govern feature learning at finite width, and empirically demonstrate that these (suitably normalized) correlators become independent of depth at $L \sim 20$. This provides a plausible theoretical link between the previously-observed superior generalization of deep orthogonal networks, and the dynamics of the NTK under GD at finite width, confirmed by our numerical experiments which show that the GD training dynamics become independent of depth for $L \sim 20$.

This paper is organized as follows. In Sec.~\ref{sec:preliminaries}, we set out our notation and review the relevant parts of the formalism of Ref.~\cite{Roberts:2021fes}. In Sec.~\ref{sec:linear}, we warm up by computing the exact single-input preactivation distribution for orthogonal MLPs with linear activation functions, finding the satisfying result that these networks preserve the norm of layer outputs for all depths. In Sec.~\ref{sec:critical}, we compute the dimensionless single-input connected 4-point correlator at leading order in $1/n$ and show that it grows as $L/n$ for ReLU activation functions, but is depth-independent for activation functions in the $K^* = 0$ universality class of~\cite{Roberts:2021fes}. In Sec.~\ref{sec:NTK}, we empirically measure the full set of single-input NTK correlators at $\mathcal{O}(1/n)$ at initialization in an ensemble of MLPs and compare their behavior for Gaussian and orthogonal weights. In Sec.~\ref{sec:MNIST}, we perform numerical experiments by training MLPs of various $L$ and $n$ on the MNIST and CIFAR-10 classification problems using full-batch GD with weight and bias learning rates scaled according to the prescriptions of~\cite{Roberts:2021fes}. Consistent with previous studies, we find that orthogonal initializations perform essentially identically to Gaussian initializations in the $L/n \ll 1$ regime, but have superior performance in the $L/n \sim 1$ regime, where we also observe suppressed NTK fluctuations during training for orthogonal networks. We conclude in Sec.~\ref{sec:conclusion} with an outlook and plans for future work. Further measurements of multi-input correlators are provided in Appendix~\ref{app:Multi}, and variations on our training experiments comprise Appendix~\ref{app:VarTraining}.

\section{Notation and preliminaries}
\label{sec:preliminaries}
Throughout this paper, we will use the notation of \cite{Roberts:2021fes}, which we briefly review here. The MLP architecture is defined by a forward iteration equation
\be
\label{eq:zforward}
z_i^{(1)} = b_i^{(1)} + \sum_{j=1}^{n_0} W_{ij}^{(1)}x_j, \qquad z^{(\ell+1)}_i = b_i^{(\ell+1)} + \sum_{j=1}^{n_\ell} W_{ij}^{(\ell+1)} \sigma(z^{(\ell)}_j), 
\ee
where $z^{(\ell)}_j$ is the preactivation at layer $\ell$, $W_{ij}^{(\ell+1)}$ and $b_j^{(\ell+1)}$ are the weights and biases at layer $\ell+1$ with $1 \leq \ell \leq L-1$, $n_\ell$ is the width of layer $\ell$, and $\sigma$ is the element-wise activation function. The input layer ($\ell = 0$) will be denoted $\mathbf{x} \equiv \mathbf{z}^{(0)}$; all boldface vectors at layer $\ell$ have $n_\ell$ components, and $||\mathbf{z}|| \equiv \sqrt{\mathbf{z} \cdot \mathbf{z}}$. Neural indices at each layer are denoted with Latin subscripts running from 1 to $n_\ell$, and we abbreviate $\sigma(z^{(\ell)}_j) \equiv \sigma^{(\ell)}_j$ and similarly with derivatives of activation functions, $\sigma'(z^{(\ell)}_j) \equiv \sigma_j'^{(\ell)}$. Greek indices will refer to samples from the dataset $\mathcal{D} \equiv \{ \mathbf{x}_\alpha \}$, with $z_{i; \alpha} \equiv z_i(\mathbf{x}_\alpha)$ the preactivations corresponding to input $\mathbf{x}_\alpha$.

When we consider statistics of networks at initialization, we will envision $\mathcal{D}$ as fixed, such that $\mathbb{E}[\cdots]$ denotes an expectation over the only source of stochasticity in the model, namely the initialization distributions. All probability distributions are taken to be conditional on the data, which may be a single input or the entire set $\mathcal{D}$. The weights and biases in each layer will be taken to have zero mean, with variances defined by
    \begin{align}
            \label{eq:weightvar}
        \E[W^{(\ell)}_{i_1 j_1}W^{(\ell)}_{i_2 j_2}] & = \delta_{i_1 i_2} \delta_{j_1 j_2}\frac{C_W}{n_{\ell - 1}}, \\
        \label{eq:biasvar}
        \E[b_{i_1}^{(\ell)}b_{i_2}^{(\ell)}] & = \delta_{i_1 i_2}C_b,
    \end{align}
where for simplicity, we assume $C_W$ and $C_b$ are constant across layers.  In particular, we will use Eq.~(\ref{eq:weightvar}) to define scaled $n \times n$ orthogonal matrices $O$ satisfying $O^T O  = C_W \mathbb{I}_{n \times n}$. In order to have square weight matrices to more easily compare orthogonal and Gaussian weights, we will take all hidden layers to have width $n$, and abbreviate $n_\ell \equiv n$ for $\ell = 1, 2, \dots, L-1$. Consequently, we will drop the layer label on $W$ and $b$ when computing expectations since their distributions are identical at each layer.

If the weight distribution were Gaussian, then Eq.~(\ref{eq:weightvar}) would fully specify the initialization hyperparameters because all higher-point moments of weights in the same layer can be expressed in terms of $C_W$ by Wick's theorem. In particular, the fourth moment would take the form
\be
\label{eq:4WGaussian}
\E[W_{i_1 j_1} W_{i_2 j_2} W_{i_3 j_3} W_{i_4 j_4}] = \frac{C_W^2}{n^2} \left[  (12)(34)_i (12)(34)_j + (13)(24)_i (13)(24)_j + (14)(23)_i (14)(23)_j \right ]
\ee
where we are using an abbreviated notation $(ab)(cd)_i \equiv \delta_{i_a i_b} \delta_{i_c i_d}$. However, for weights drawn uniformly from the Haar measure on scaled orthogonal matrices with $n > 2$, the fourth moment is instead
\be
\label{eq:4WOrthogonal}
\E[W_{i_1 j_1} W_{i_2 j_2} W_{i_3 j_3} W_{i_4 j_4}] = C_W^2
\begin{pmatrix}  (12)(34)_i \\ (13)(24)_i \\ (14)(23)_i \end{pmatrix}^T \begin{pmatrix} \mathcal{W}[1,1] & \mathcal{W}[2] & \mathcal{W}[2] \\ \mathcal{W}[2] & \mathcal{W}[1,1] & \mathcal{W}[2] \\ \mathcal{W}[2] & \mathcal{W}[2] & \mathcal{W}[1,1] \end{pmatrix}
\begin{pmatrix} (12)(34)_j \\ (13)(24)_j \\ (14)(23)_j \end{pmatrix},
\ee
where 
\begin{equation}
\label{eq:Weingarten}
\mathcal{W}[1,1] = \frac{n+1}{(n-1)n(n+2)}, \qquad \mathcal{W}[2] = \frac{-1}{(n-1)n(n+2)}
\end{equation}
are combinatorial factors computed from the Weingarten functions $\mathcal{W}$ ~\cite{Weingarten:1977ya,collins2006integration,collins2009some,Weingarten,MR4415894}. Note that for large $n$, $\mathcal{W}[1,1] \sim 1/n^2$ and $\mathcal{W}[2] \sim 1/n^3$, so that at large width, orthogonal and Gaussian weights with the same variance have identical fourth moments up to $1/n$ corrections.\footnote{Throughout this paper, we will avoid discussing the infinite-width limit whenever possible because of a rather subtle order-of-limits issue. Orthogonal matrices at all finite $n$ retain higher-point correlations between entries, but in the strict $n \to \infty$ limit, the matrix entries become independent Gaussian-distributed variables and the matrix is no longer orthogonal. In other words, the orthogonality constraint involves summing $n$ terms which becomes ill-defined as $n \to \infty$ unless $1/n$ effects are included. To avoid this complication, we will keep $n$ finite in all calculations.}

The kernel $K^{(\ell)}$ and the 4-point vertex $V^{(\ell)}$ are defined as 2- and 4-point connected correlators of preactivations at layer $\ell$ to leading order in $1/n$ across an ensemble of MLPs,\footnote{The nomenclature ``vertex'' for $V$ refers to a useful diagrammatic notation for calculations, see Refs.~\cite{dyer2019asymptotics,Banta:2023kqe} for examples.}
    \begin{align}
    \label{eq:Kdef}
        \mathbb{E}[z_{i_i; \alpha_1}^{(\ell)}z^{(\ell)}_{i_2; \alpha_2}] & = \delta_{i_1 i_2} K_{\alpha_1 \alpha_2}^{(\ell)} + \mathcal{O}(1/n), \\
        \label{eq:Vdef}
        \mathbb{E}[z^{(\ell)}_{i_1; \alpha_1}z^{(\ell)}_{i_2; \alpha_2}z^{(\ell)}_{i_3; \alpha_3}z^{(\ell)}_{i_4; \alpha_4}]_{\rm conn.} & = \frac{1}{n} \left [(12)(34)_i V^{(\ell)}_{(\alpha_1 \alpha_2)(\alpha_3 \alpha_4)} + ({\rm 2 \ sim.}) \right] + \mathcal{O}(1/n^2).
    \end{align}
In Eq.~(\ref{eq:Kdef}), the $\mathcal{O}(1/n)$ refers to the fact that $K$ is the leading part of the 2-point correlator; we will refer to the full 2-point correlator to all orders in $1/n$ as the metric, discussed further in Sec.~\ref{sec:NLOMetric}. The neural index tensor structure of Eq.~(\ref{eq:Kdef}) follows from Eq.~(\ref{eq:weightvar}) because it involves an expectation over two weights, which is identical for Gaussian and orthogonal initializations. Similarly, the tensor structure of Eq.~(\ref{eq:Vdef}) follows from Eq.~(\ref{eq:4WGaussian}) for Gaussian weights, where the two similar terms in Eq.~(\ref{eq:Vdef}) have the neural indices and sample indices matched up as in the displayed term, and we will show in Sec.~\ref{sec:linear} below that it also holds for orthogonal weights. The structure of the forward equation (\ref{eq:zforward}) allows a type of dimensional analysis (known in the physics literature as ``power counting'') where we treat the preactivations $z$ as dimensionful variables. Thus, when computing the 4-point vertex as a function of depth, we will use the normalized combination
\be
\widetilde{V}^{(\ell)} \equiv \frac{V^{(\ell)}}{(K^{(\ell)})^2},
\label{eq:VNormDef}
\ee
which is dimensionless because it has 4 powers of $z$ in the numerator and 4 in the denominator.\footnote{We have deliberately kept Eq.~\ref{eq:VNormDef} schematic by leaving off the sample indices, because in general, for $\alpha_1 \neq \alpha_2 \neq \alpha_3 \neq \alpha_4$, dimensional analysis provides no unique prescription to choose the indices in $K$. We will show in Appendix~\ref{app:MultiV} that for the purposes of determining scaling with width and depth, these choices do not matter.}

Finally, because the weight distribution $p(W)$ is still nearly-Gaussian up to $1/n$ corrections, we can follow the prescription of \cite{Roberts:2021fes} to integrate out the weights and biases and derive marginal distributions $p(\mathbf{z}^{(\ell+1)} | \mathbf{z}^{(\ell)})$ which are also nearly-Gaussian. In doing so we will encounter Gaussian expectations with covariance matrix $g$, which we denote with angle brackets,
\be
\label{eq:bracketdef}
\langle F(z_{\alpha_1}, \dots, z_{\alpha_m}) \rangle_g \equiv \int \left [ \frac{\prod_{\alpha \in \mathcal{D}} dz_\alpha}{\sqrt{{\rm det} (2\pi g)}}\right]\, \exp\left(-\frac{1}{2} \sum_{\beta_1, \beta_2 \in \mathcal{D}} (g^{-1})^{\beta_1 \beta_2}z_{\beta_1}z_{\beta_2}\right) F(z_{\alpha_1}, \dots, z_{\alpha_m}).
\ee

\section{Exact preactivation distribution for linear orthogonal networks}
\label{sec:linear}

Here we derive the fact that the probability distribution for preactivations of a linear orthogonal network of width $n$, with zero biases ($b_i^{(\ell)} = 0$) and a single input, is a uniform distribution on the sphere $S^{n-1}$ with radius equal to the norm of the input. Though this result is rather obvious from the definition of an orthogonal matrix, the calculational techniques serve as a useful warm-up for the nonlinear networks considered in Sec.~\ref{sec:critical}, where a full closed-form expression is not available. We begin by explicitly calculating the 2- and 4-point correlators, then generalize to $m$-point correlators in order to construct the characteristic function of the probability distribution, and finally Fourier-transform the characteristic function to obtain the probability distribution itself. We will find it instructive to compare this derivation to the results of Ref.~\cite{zavatone2021exact} for the identical architecture with Gaussian initializations.

For orthogonal weights, the 2-point correlator of preactivations of the first hidden layer, with a single input $\mathbf{x}$ taken to have the same width $n$ as the hidden layers, is given by
\begin{equation}
\mathbb{E}[z_{i_1}^{(1)}z_{i_2}^{(1)}] =\sum_{j_1, j_2} \mathbb{E}[W_{i_1 j_1}W_{i_2 j_2}] x_{j_1} x_{j_2} = \delta_{i_1 i_2} \frac{C_W}{n} ||\mathbf{x}||^2,
\end{equation}
where as mentioned we have set all biases to zero, and the second equality makes use of Eq.~(\ref{eq:weightvar}). Noting the index structure, we make the ansatz
\begin{equation}
\mathbb{E}[z_{i_1}^{(\ell)}z_{i_2}^{(\ell)}] = \delta_{i_1 i_2} K^{(\ell)},
\end{equation}
so that 2-point correlator for an arbitrary layer is
\begin{equation}
\mathbb{E}[z_{i_1}^{(\ell+1)} z_{i_2}^{(\ell+1)}] =\sum_{j_1, j_2} \mathbb{E}[W_{i_1 j_1} W_{i_2 j_2}] \, \mathbb{E}[z_{j_1}^{(\ell)} z_{j_2}^{(\ell)}] = \delta_{i_1 i_2} \frac{C_W}{n} \sum_j K^{(\ell)} = C_W \delta_{i_1 i_2} K^{(\ell)}.
\end{equation}
Thus we have the kernel recursion
\begin{equation}
K^{(\ell+1)} = C_W K^{(\ell)} = (C_W)^{\ell+1} K^{(0)}
\end{equation}
with initial condition
\begin{equation}
K^{(0)} \equiv \frac{1}{n}||\mathbf{x}||^2,
\end{equation}
which is identical to the recursion for Gaussian-initialized linear networks~\cite{Roberts:2021fes}. In particular, choosing $C_W = 1$ ensures criticality where $K^{(\ell)} = K^{(0)}$.

The 4-point correlator of preactivations of the first hidden layer is given by
\begin{align}
\mathbb{E}[z_{i_1}^{(1)}z_{i_2}^{(1)}z_{i_3}^{(1)}z_{i_4}^{(1)}] &= \sum_{j_1, j_2, j_3, j_4} \mathbb{E}[W_{i_1 j_1}W_{i_2 j_2}W_{i_3 j_3}W_{i_4 j_4}] x_{j_1}x_{j_2}x_{j_3}x_{j_4} \nonumber \\
&= C_W^2 \sum_{j_1, j_2, j_3, j_4} x_{j_1}x_{j_2}x_{j_3}x_{j_4} \nonumber \\
&\times \begin{pmatrix} (12)(34)_i \\ (13)(24)_i \\ (14)(23)_i \end{pmatrix}^T
\begin{pmatrix} \mathcal{W}[1,1] & \mathcal{W}[2] & \mathcal{W}[2] \\ \mathcal{W}[2] & \mathcal{W}[1,1] & \mathcal{W}[2] \\ \mathcal{W}[2] & \mathcal{W}[2] & \mathcal{W}[1,1] \end{pmatrix}
\begin{pmatrix} (12)(34)_j \\ (13)(24)_j \\ (14)(23)_j \end{pmatrix} \nonumber \\
&= C_W^2 \left((12)(34)_i + (13)(24)_i + (14)(23)_i\right) \left(\mathcal{W}[1,1] + 2\mathcal{W}[2] \right) \Big(||\mathbf{x}||^2\Big)^2 \nonumber \\
&= \frac{C_W^2}{n^2+2n} \left((12)(34)_i + (13)(24)_i + (14)(23)_i\right) \Big(||\mathbf{x}||^2\Big)^2 ,
\end{align}
where the second equality makes use of Eq.~(\ref{eq:4WOrthogonal}). Again noting the index structure, we make the ansatz
\begin{equation}
\mathbb{E}[z_{i_1}^{(\ell)}z_{i_2}^{(\ell)}z_{i_3}^{(\ell)}z_{i_4}^{(\ell)}] = \big((12)(34)_i + (13)(24)_i + (14)(23)_i\big) G_4^{(\ell)}
\end{equation}
so that the 4-point correlator of an arbitrary layer is
\begin{align}
\mathbb{E}[z_{i_1}^{(\ell+1)}z_{i_2}^{(\ell+1)}z_{i_3}^{(\ell+1)}&z_{i_4}^{(\ell+1)}] = \sum_{j_1, j_2, j_3, j_4} \mathbb{E}[W_{i_1 j_1}W_{i_2 j_2}W_{i_3 j_3}W_{i_4 j_4}] \, \mathbb{E}[z_{j_1}^{(\ell)}z_{j_2}^{(\ell)}z_{j_3}^{(\ell)}z_{j_4}^{(\ell)}] \nonumber \\
&= C_W^2 \sum_{j_1, j_2, j_3, j_4} \big((12)(34)_j + (13)(24)_j + (14)(23)_j\big) G_4^{(\ell)} \nonumber \\
&\times\begin{pmatrix} (12)(34)_i \\ (13)(24)_i \\ (14)(23)_i \end{pmatrix}^T
\begin{pmatrix} \mathcal{W}[1,1] & \mathcal{W}[2] & \mathcal{W}[2] \\ \mathcal{W}[2] & \mathcal{W}[1,1] & \mathcal{W}[2] \\ \mathcal{W}[2] & \mathcal{W}[2] & \mathcal{W}[1,1] \end{pmatrix}
\begin{pmatrix} (12)(34)_j \\ (13)(24)_j \\ (14)(23)_j \end{pmatrix} \nonumber \\
&= C_W^2 \big((12)(34)_i + (13)(24)_i + (14)(23)_i\big) \left(\mathcal{W}[1,1] + 2\mathcal{W}[2] \right) (n^2 + 2n) G_4^{(\ell)} \nonumber \\
&= C_W^2 \big((12)(34)_i + (13)(24)_i + (14)(23)_i\big) G_4^{(\ell)}.
\label{magic}
\end{align}
Note that the $1/(n^2+2n)$ factor from the sum of rows of the Weingarten matrix exactly cancels with the $n^2+2n$ from summing over the Kronecker deltas from the Wick contraction pairings, whereas the corresponding result for a Gaussian initialization would instead have a remaining factor of $(n+2)/n$. Thus, for orthogonal initializations, the $G_4$ recursion for an orthogonal distribution is independent of width:
\begin{equation}
G_4^{(\ell+1)}=C_W^2 G_4^{(\ell)}=(C_W^2)^{\ell+1} G_4^{(0)},
\end{equation}
with initial condition
\begin{equation}
G_4^{(0)} \equiv \frac{1}{n(n+2)} \Big(||\mathbf{x}||^2\Big)^2  = \frac{n}{n+2} \left(K^{(0)}\right)^2.
\end{equation}
Furthermore, the 4-point vertex (i.e.\ the connected 4-point correlator) now takes on the simple form
\begin{equation}
V^{(\ell)} \equiv G_4^{(\ell)} - \left(K^{(\ell)}\right)^2 = -\frac{2}{n+2}\left(K^{(\ell)}\right)^2 \qquad {\rm (orthogonal)}.
\end{equation}
Compared with the analogous result for Gaussian initializations~\cite{Dyer:2019uzd}, 
\begin{equation}
V^{(\ell)} = \frac{2(\ell-1)}{n} \left(K^{(\ell)}\right)^2 + \mathcal{O}(1/n^2) \qquad {\rm (Gaussian)},
\end{equation}
the 4-point vertex for orthogonal initializations is independent of depth if $K$ is tuned to a fixed point, $K^{(\ell)} = K^*$, while the same quantity grows linearly with depth for Gaussian initializations.

To generalize the above derivation to higher-point correlators, we use the fact that any $2m$-point expectation of orthogonal matrix elements can be written as an analogous Weingarten matrix~\cite{Weingarten}, whose rows or columns each sum to $[n(n+2)(n+4)\cdots(n+2m-2)]^{-1}$. The inverse of this factor is proportional to the combinatorial factor which sums Wick contractions of $m$ indices with values from 1 to $n$, defined as 
\begin{equation}
c_{2m}(n) = \frac{n(n+2)(n+4)\cdots(n+2m-2)}{n^m}
\end{equation}
in~\cite{Roberts:2021fes}. Thus, these two factors will always cancel when evaluating $2m$-point expectations of preactivations after the first layer. This somewhat magical cancellation occurs because the Weingarten matrix is, by definition, the pseudo-inverse of the Gram matrix of the vector space spanned by all $(2m-1)!!$ pair partitions of $2m$ indices~\cite{Weingarten}.

Now, defining the $2m$-point correlator $G_{2m}^{(\ell)}$ through the ansatz
\begin{equation}
\mathbb{E}[z_{i_1}^{(\ell)}z_{i_2}^{(\ell)}...z_{i_{2m}}^{(\ell)}] = \bigg ( \sum_{\rm all \ pairings} \delta_{i_{k_1} i_{k_2}} \delta_{i_{k_3} i_{k_4}} \cdots \delta_{i_{k_{2m-1}} i_{k_{2m}}} \bigg ) G_{2m}^{(\ell)},
\end{equation}
such that $G_2^{(\ell)} \equiv K^{(\ell)}$ (and $G_{2m+1}^{(\ell)} = 0$), we can compute the $2m$-point expectation for all $m$:
\begin{equation}
\begin{aligned}
\mathbb{E}[z_{i_1}^{(\ell+1)}z_{i_2}^{(\ell+1)}...&z_{i_{2m}}^{(\ell+1)}] = \sum_{j_1, j_2,..., j_{2m}} \mathbb{E}[W_{i_1 j_1}W_{i_2 j_2}...W_{i_{2m} j_{2m}}] \, \mathbb{E}[z_{j_1}^{(\ell)}z_{j_2}^{(\ell)}...z_{j_{2m}}^{(\ell)}] \\
&= C_W^m \bigg ( \sum_{\rm all \ pairings} \delta_{i_{k_1} i_{k_2}} \delta_{i_{k_3} i_{k_4}} \cdots \delta_{i_{k_{m-1}} i_{k_m}} \bigg ) G_{2m}^{(\ell)}.
\end{aligned}
\end{equation}
Thus we have the recursion 
\begin{equation}
\label{eq:G2mortho}
G_{2m}^{(\ell+1)}=C_W^m G_{2m}^{(\ell)}=C_W^{m(\ell+1)} G_{2m}^{(0)} \qquad {\rm (orthogonal}),
\end{equation}
where the Weingarten combinatorial factor appears in the initial condition:
\begin{equation}
G_{2m}^{(0)} = \frac{n^m}{n(n+2)(n+4)\cdots(n+2m-2)}\left(K^{(0)}\right)^{m}.
\end{equation}
Critical initialization with $C_W = 1$ ensures that this recursion is trivial, $G_{2m}^{(\ell)} = G_{2m}^{(0)}$, while other choices of $C_W$ lead to exponential growth or decay. The analogous recursion for Gaussian initializations, from Ch.~3 of Ref.~\cite{Roberts:2021fes}, involves a product of $c_{2m}(n)$ and thus is considerably more complicated.

Having computed all of the moments of $z^{(\ell)}$, the characteristic function $\varphi_{\mathbf{z}^{(\ell)}}(u)$ may now be constructed:
\begin{equation}
\varphi_{\mathbf{z}^{(\ell)}}(\mathbf{u}) \equiv\mathbb{E}[e^{i\mathbf{u} \cdot \mathbf{z}^{(\ell)}}]=1-\frac{1}{2!}\sum_{i_1, i_2}u_{i_1} u_{i_2} \mathbb{E}[z^{(\ell)}_{i_1}z^{(\ell)}_{i_2}]+\frac{1}{4!}\sum_{i_1, i_2, i_3, i_4}u_{i_1} u_{i_2} u_{i_3} u_{i_4}\mathbb{E}[z^{(\ell)}_{i_1}z^{(\ell)}_{i_2}z^{(\ell)}_{i_3}z^{(\ell)}_{i_4}]-...\\
\end{equation}
All terms contain only even powers of $z$ since the odd moments vanish, and each expectation yields $(2m-1)!!$ Wick pairings and a Weingarten combinatorial factor from the initial condition of the $G_{2m}$ recursion. Rearranging some factors, we obtain the power series representation
\begin{equation}
\varphi_{\mathbf{z}^{(\ell)}}(u) = \sum_{m=0}^\infty \frac{1}{(n/2)(n/2 + 1)\cdots(n/2 + (m-1))}\frac{(-n C_W^\ell K^{(0)} u^2/4)^m}{m!},
\end{equation}
where $u =||\mathbf{u}||$. Note that the characteristic function is purely radial, depending only on $u$, which makes sense because $\varphi$ should be invariant under orthogonal transformations of $\mathbf{z}$. Furthermore, the power series takes the form of a confluent hypergeometric limit function,
\begin{equation}
\varphi_{\mathbf{z}^{(l)}}(u) = ~_0F_1(;\tfrac{n}{2};-\tfrac{1}{4} u^2 n C_W^{\ell} K^{(0)}),
\end{equation}
which can also be expressed as a Bessel function,
\begin{equation}
\varphi_{\mathbf{z}^{(l)}}(u) =\Big(\frac{u}{2}\sqrt{nC_W^\ell K^{(0)}}\Big)^{-\frac{n-2}{2}}\Gamma(n/2) \, J_{\frac{n-2}{2}}\Big(u\sqrt{nC_W^{\ell}K^{(0)}}\Big).
\end{equation}
This Bessel form is useful for computing the integral in the next paragraph.

Proceeding on, we take the Fourier transform to recover the probability distribution for $\mathbf{z}^{(\ell)}$. Since the characteristic function is radial, we may perform the $n-1$ angular integrals in an identical fashion to Ref.~\cite{zavatone2021exact} using Hankel transform relations. Letting $z^{(\ell)} = ||\mathbf{z}^{(\ell)}||$, we have
\begin{equation}
\begin{aligned}
p_n(\mathbf{z}^{(\ell)}| \mathbf{x})&=(2\pi)^{-n/2}(z^{(\ell)})^{\frac{2-n}{2}}\int_0^\infty du\,  u \, u^{\frac{n-2}{2}}\, J_{\frac{n-2}{2}}(z^{(\ell)}u)\, \varphi_{\mathbf{z}^{(\ell)}}(u)\\
&=\pi^{-\frac{n}{2}}(C_W^{\ell} K^{(0)})^{-\frac{n-2}{4}} n^{-\frac{n+2}{4}}\Gamma(1+ \tfrac{n}{2}) (z^{(\ell)})^{-\frac{n}{2}} \delta\Big(z^{(\ell)}-\sqrt{nC_W^{\ell} K^{(0)}}\Big) \ {\rm (orthogonal)},
\end{aligned}
\label{eq:pOrtho}
\end{equation}
where the delta function arises from the orthogonality relation for Bessel functions of like order. Thus, the exact distribution of preactivations at layer $\ell$ is simply a uniform distribution on the sphere $S^{n-1}$ of radius $\sqrt{n C_W^\ell K^{(0)}}$, since all the factors apart from the delta function rearrange to cancel the volume factor $d\Omega_{n-1}$. If we further choose critical initialization, $C_W = 1$, then $p_n(\mathbf{z}^{(\ell)} | \mathbf{x})$ becomes independent of $\ell$ and preserves the norm of the input data, since $\sqrt{n K^{(0)}} = ||\mathbf{x}||$. Successive layers of the linear network simply apply random rotations and/or reflections to the preactivations without changing their magnitude. Of course, this is exactly what one should expect from multiplying an input vector $\mathbf{x}$ by successive random Haar-distributed orthogonal matrices, which is what this linear network does.

It is interesting to compare the simple interpretation of Eq.~(\ref{eq:pOrtho}) with the analogous result for a rectangular network with Gaussian initializations from Ref.~\cite{zavatone2021exact}, translated into our notation:
\begin{equation}
p_n(\mathbf{z}^{(\ell)}| \mathbf{x}) = \frac{1}{(\Gamma(n/2))^{\ell-1}(2^\ell \pi \, n C_W^\ell K^{(0)})^{n/2}} G^{\ell,0}_{0,\ell} \Big ( \left. \frac{(z^{(\ell)})^2}{2^\ell n C_W^\ell K^{(0)}} \right | \begin{array}{c} - \\ 0, 0, \dots, 0 \end{array} \Big ) \qquad {\rm (Gaussian)}
\end{equation}
where $G^{\ell,0}_{0,\ell}$ is a Meijer $G$-function. In this case, even at criticality with $C_W = 1$, the dependence on depth is explicit in the order of the $G$-function, and expanding perturbatively in $1/n$ yields non-Gaussianities that grow with $\ell$, as derived in Ch.~3 of Ref.~\cite{Roberts:2021fes}:
\begin{equation}
    G_{2m}^{(\ell)} = \left(c_{2m}(n)\right)^{\ell-1}\left(K^{(0)}\right)^m \qquad {\rm (Gaussian)},
\end{equation}
which should be contrasted with the depth-independent recursion for orthogonal initializations, Eq.~(\ref{eq:G2mortho}).

\section{Critical initialization and fluctuations for nonlinear orthogonal networks}
\label{sec:critical}

While deep linear networks provide a nice warmup, they are necessarily limited in that they are only able to compute linear functions of their input. To study more generally useful networks, we need to consider networks with nonlinear activation functions. In this section, we follow the logic of Ref.~\cite{Roberts:2021fes} and use the recursion relation for the kernel to determine the orthogonal matrix scaling factors $C_W$ (and bias variance $C_b$) that ensure critical behavior for orthogonal initializations with various activations. We then show that the normalized single-input 4-point vertex $\widetilde{V}$ is depth-independent for linear and tanh activations, but scales linearly with depth for ReLU activations, consistent with the results of Ref.~\cite{pennington2017resurrecting} for the singular value variance of the input-output Jacobian. We further show that at leading order in $1/n$, finite-width corrections to the kernel do not require further tunings of $C_W$ to maintain criticality with tanh activations, in contrast to the situation for Gaussian initializations.

\subsection{Kernel recursion and critical initialization hyperparameters}
To derive a recursion relation for the kernel, we compute the 2-point expectation
\begin{align}
\E[z^{(\ell+1)}_{i_1; \alpha_1} z^{(\ell+1)}_{i_2; \alpha_2}] & = \E \left [\left(b_{i_1}^{(\ell+1 )} + \sum_{j_1} W^{(\ell+1)}_{i_1 j_1 }  \sigma^{(\ell)}_{j_1; \alpha_1} \right) \left(b_{i_2}^{(\ell+1 )} + \sum_{j_2} W^{(\ell+1)}_{i_2 j_2} \sigma^{(\ell)}_{j_2; \alpha_2}\right) \right ] \\
& = C_b \delta_{i_1 i_2} + \frac{C_W}{n} \delta_{i_i i_2} \sum_j \E [\sigma^{(\ell)}_{j; \alpha_1} \sigma^{(\ell)}_{j; \alpha_2} ],
\label{eq:zzkernel}
\end{align}
where we have used Eqs.~(\ref{eq:weightvar})--(\ref{eq:biasvar}) and the statistical independence of the $(\ell+1)$-th layer weights from the $\ell$-th layer activations. Now, we recall that the non-Gaussianity of the orthogonal weight distribution is suppressed by $1/n$, as detailed in Sec.~\ref{sec:preliminaries}. Thus, the preactivation distribution $p_n(\mathbf{z}^{(\ell)} | \mathcal{D})$, which is used to compute the remaining expectation in Eq.~(\ref{eq:zzkernel}), only receives $1/n$ corrections compared to initializations with independent Gaussian weights. By the results of~\cite{Roberts:2021fes}, $p_n(\mathbf{z}^{(\ell)} | \mathcal{D})$ with Gaussian weights is also Gaussian up to $1/n$ corrections, so to leading order in $1/n$ we can replace the expectation by a Gaussian expectation with respect to the kernel $K_{\alpha_1 \alpha_2}^{(\ell)}$ in layer $\ell$. Likewise, evaluating the left-hand side to leading order in $1/n$ using Eq.~(\ref{eq:Kdef}), we obtain the recursion 
\be
\label{eq:KRecursion}
K^{(\ell+1)}_{\alpha_1 \alpha_2} = C_b + C_W \langle \sigma_{\alpha_1} \sigma_{\alpha_2} \rangle_{K^{(\ell)}},
\ee
where the brackets mean a Gaussian expectation with respect to $K^{(\ell)}$ as defined in Eq.~(\ref{eq:bracketdef}). Eq.~(\ref{eq:KRecursion}) is identical to the kernel recursion for Gaussian weights, which means the criticality conditions are also identical. We could have anticipated this result, since the kernel recursion is a statement about the infinite-width behavior, and the suppression of non-Gaussianities by $1/n$ in the orthogonal distribution means that orthogonal and Gaussian initializations are the same in the limit $n \to \infty$.

As derived in Ch.~5 of Ref.~\cite{Roberts:2021fes}, the single-input susceptibilities $\chi_\parallel(K)$ and $\chi_\perp(K)$ control the behavior of the kernel for two nearby inputs. They are defined as
\begin{align}
\chi_{\parallel}(K) & \equiv \frac{C_W}{2K^2} \langle \sigma(z) \sigma(z) (z^2 - K) \rangle_K, \\ 
\chi_\perp(K) & \equiv C_W \langle \sigma'(z) \sigma'(z) \rangle_K.
\end{align}
The criticality conditions are $\chi_{\parallel}(K) = \chi_{\perp}(K) = 1$, which imposes that the kernel neither grows nor decays exponentially, and thus nearby inputs stay close as they propagate through the network.\footnote{Ref.~\cite{Banta:2023kqe} showed that $\chi_\parallel = \chi_\perp  = 1$ is actually a sufficient condition for preventing exponential growth/decay of \emph{all} higher-point correlators for multiple inputs, in MLP architectures with Gaussian weights.} The value of $K$ which solves these conditions is the fixed point $K^*$, which implicitly satisfies
\be
K^* = C_b + C_W \langle \sigma(z) \sigma(z) \rangle_{K^*}
\ee
for any single input vector $\mathbf{x}$. 

The value of $K^*$ is activation-function dependent and, except for some edge cases, can be divided into two universality classes where the behavior of correlators is the same for all activation functions in the universality class up to order-one factors. The first class contains scale-invariant (piecewise-linear) activation functions such as linear and ReLU, for which 
\be
(C_b, C_W)^{\rm critical} = \left(0, \frac{1}{A}\right); \qquad K^* = \frac{1}{n_0 A} ||\mathbf{x}||^2 \ \qquad \textrm{(scale-invariant}),
\ee
with $A = 1$ for linear and $A = 1/2$ for ReLU. The second class contains smooth nonlinear activation functions with $\sigma(0) =0$, where 
\be
(C_b, C_W)^{\rm critical} = \left(0, \frac{1}{(\sigma'(0))^2}\right); \qquad K^* = 0
\ee
and the kernel decays to zero like $1/\ell$. In particular, $C_W = 1$ is critical for tanh activation functions.

\subsection{4-point vertex recursion}

We now follow the same logic to compute the recursion relation for the 4-point vertex $V$. Since both universality classes considered above have the biases initialized to zero $(C_b = 0)$ at criticality, we will drop the biases for simplicity in the calculations that follow. Consider the expectation of the fourth moment of preactivations:
\begin{align}
& \E[z^{(\ell+1)}_{i_1; \alpha_1}z^{(\ell+1)}_{i_2; \alpha_2}z^{(\ell+1)}_{i_3; \alpha_3}z^{(\ell+1)}_{i_4; \alpha_4}]  = \sum_{j_1, j_2, j_3, j_4} \E[W_{i_1 j_1} W_{i_2 j_2} W_{i_3 j_3} W_{i_4 j_4}] \, \E[\sigma^{(\ell)}_{j_1; \alpha_1}\sigma^{(\ell)}_{j_2; \alpha_2}\sigma^{(\ell)}_{j_3; \alpha_3}\sigma^{(\ell)}_{j_4; \alpha_4}] \nonumber \\
& = C_W^2 
\begin{pmatrix} (12)(34)_i \\ (13)(24)_i \\ (14)(23)_i \end{pmatrix} ^T \sum_{j_1, j_2, j_3, j_4}\begin{pmatrix} \mathcal{W}[1,1] & \mathcal{W}[2] & \mathcal{W}[2] \\ \mathcal{W}[2] & \mathcal{W}[1,1] & \mathcal{W}[2] \\ \mathcal{W}[2] & \mathcal{W}[2] & \mathcal{W}[1,1] \end{pmatrix}
\begin{pmatrix} (12)(34)_j \\ (13)(24)_j \\ (14)(23)_j \end{pmatrix} \E[\sigma^{(\ell)}_{j_1; \alpha_1}\sigma^{(\ell)}_{j_2; \alpha_2}\sigma^{(\ell)}_{j_3; \alpha_3}\sigma^{(\ell)}_{j_4; \alpha_4}] \nonumber \\
& = C_W^2 
\begin{pmatrix} (12)(34)_i \\ (13)(24)_i \\ (14)(23)_i \end{pmatrix} ^T\begin{pmatrix} \mathcal{W}[1,1] & \mathcal{W}[2] & \mathcal{W}[2] \\ \mathcal{W}[2] & \mathcal{W}[1,1] & \mathcal{W}[2] \\ \mathcal{W}[2] & \mathcal{W}[2] & \mathcal{W}[1,1] \end{pmatrix} \begin{pmatrix} \sum_{j,k} \E[\sigma^{(\ell)}_{j; \alpha_1} \sigma^{(\ell)}_{j; \alpha_2}\sigma^{(\ell)}_{k; \alpha_3} \sigma^{(\ell)}_{k; \alpha_4}] \\ \sum_{j,k} \E[\sigma^{(\ell)}_{j; \alpha_1} \sigma^{(\ell)}_{j; \alpha_3}\sigma^{(\ell)}_{k; \alpha_2} \sigma^{(\ell)}_{k; \alpha_4}] \\ \sum_{j,k} \E[\sigma^{(\ell)}_{j; \alpha_1} \sigma^{(\ell)}_{j; \alpha_4}\sigma^{(\ell)}_{k; \alpha_2} \sigma^{(\ell)}_{k; \alpha_3}] \end{pmatrix},
\end{align}
where in the second line we used Eq.~(\ref{eq:4WOrthogonal}) for the expectation of four orthogonal weights.
Let us now expand this expectation in powers of $1/n$, beginning with an expansion of the Weingarten factors in Eq.~(\ref{eq:Weingarten}):
\begin{align}
\mathcal{W}[2] & = \frac{-1}{(n-1)n(n+2)} = -\frac{1}{n^3} + \mathcal{O}(1/n^4); \\
\mathcal{W}[1,1] & = \frac{n+1}{(n-1)n(n+2)} = \frac{1}{n^2} + \frac{2}{n^4} + \mathcal{O}(1/n^5).
\end{align}
Since $\mathcal{W}[1,1] \sim 1/n^2$ but $\mathcal{W}[2] \sim 1/n^3$, the leading terms are on the diagonal, in which case the Weingarten matrix is proportional to the identity matrix and we have
\be \E[z^{(\ell+1)}_{i_1; \alpha_1}z^{(\ell+1)}_{i_2; \alpha_2}z^{(\ell+1)}_{i_3; \alpha_3}z^{(\ell+1)}_{i_4; \alpha_4}] = \frac{C_W^2}{n^2} \left((12)(34)_i \sum_{j,k} \E[\sigma^{(\ell)}_{j; \alpha_1} \sigma^{(\ell)}_{j; \alpha_2}\sigma^{(\ell)}_{k; \alpha_3} \sigma^{(\ell)}_{k; \alpha_4}] + ({\rm 2 \ sim.}) + \mathcal{O}(1/n) \right)
\ee
where the two remaining terms have the $i$ indices paired with the corresponding $\alpha$ samples indices. At this stage, the remainder of the calculation of the expectation (and the subtraction of the various 2-point correlators to obtain the connected 4-point correlator) is identical to~\cite{Roberts:2021fes} for Gaussian weights. 

The sole effect of orthogonal weights, to leading order, is to add an additional $1/n$ term to the connected correlator from $\mathcal{W}[2]$, as the subleading part of $\mathcal{W}[1,1]$ is higher-order. Since the tensor structure of the correlator is already identical to the 4-point correlator as defined in Eq.~(\ref{eq:Vdef}), we can specialize to the coefficient of $(12)(34)_i$, which is $\dfrac{1}{n}V^{(\ell+1)}_{(\alpha_1 \alpha_2)(\alpha_3 \alpha_4)}$. Letting $\Delta V$ refer to the additional terms in $V$ specific to orthogonal initializations, we have
\be
\label{eq:VOrthogonalPiece}
\frac{1}{n} \Delta V^{(\ell+1)}_{(\alpha_1 \alpha_2)(\alpha_3 \alpha_4)} \equiv -\frac{C_W^2}{n^3} \sum_{j,k} \left(\E[\sigma^{(\ell)}_{j; \alpha_1} \sigma^{(\ell)}_{j; \alpha_3}\sigma^{(\ell)}_{k; \alpha_2} \sigma^{(\ell)}_{k; \alpha_4}] + \E[\sigma^{(\ell)}_{j; \alpha_1} \sigma^{(\ell)}_{j; \alpha_4}\sigma^{(\ell)}_{k; \alpha_2} \sigma^{(\ell)}_{k; \alpha_3}] \right).
\ee
To leading order, $\Delta V$ is $\mathcal{O}(1)$, so we must take the off-diagonal terms in the sum with $j \neq k$, of which there are $n^2 - n \sim n^2$ terms. To this order we may also replace the expectations with Gaussian expectations~\cite{Roberts:2021fes}:
\be
\E[\sigma^{(\ell)}_{j; \alpha_1} \sigma^{(\ell)}_{j; \alpha_3}\sigma^{(\ell)}_{k; \alpha_2} \sigma^{(\ell)}_{k; \alpha_4}] = \langle \sigma_{\alpha_1} \sigma_{\alpha_3} \rangle_{K^{(\ell)}} \langle \sigma_{\alpha_2} \sigma_{\alpha_2} \rangle_{K^{(\ell)}} + \mathcal{O}(1/n),
\ee
using the bracket notation of Eq.~(\ref{eq:bracketdef}) for Gaussian expectations.
Each term in the sum is then identical, so evaluating the sums just gives a factor of $n^2$, which combines with the $1/n^3$ prefactor in Eq.~(\ref{eq:VOrthogonalPiece}) to give
\be
\Delta V^{(\ell+1)}_{(\alpha_1 \alpha_2)(\alpha_3 \alpha_4)} = 
-C_W^2 \bigg [ \langle \sigma_{\alpha_1} \sigma_{\alpha_3} \rangle_{K^{(\ell)}} \langle \sigma_{\alpha_2} \sigma_{\alpha_4} \rangle_{K^{(\ell)}} + \langle \sigma_{\alpha_1} \sigma_{\alpha_4} \rangle_{K^{(\ell)}} \langle \sigma_{\alpha_2} \sigma_{\alpha_3} \rangle_{K^{(\ell)}} \bigg ] + \mathcal{O}(1/n).
\ee
Combining this contribution from orthogonal weights with the remaining parts of the 4-point vertex arising from the non-Gaussianity of the preactivation distribution~\cite{Roberts:2021fes} gives the full recursion to leading order in $1/n$:
\begin{align}
& V^{(\ell+1)}_{(\alpha_1 \alpha_2)(\alpha_3 \alpha_4)}  = C_W^2 \bigg [ \langle 1234 \rangle_{K^{(\ell)}} - \langle  12 \rangle_{K^{(\ell)}} \langle 34 \rangle_{K^{(\ell)}} - \langle  13 \rangle_{K^{(\ell)}} \langle 24 \rangle_{K^{(\ell)}} - \langle  14 \rangle_{K^{(\ell)}} \langle 23 \rangle_{K^{(\ell)}} \bigg ] \nonumber \\
& + \frac{C_W^2}{4} \sum_{\beta_1, \dots, \beta_4 \in \mathcal{D}} V_{(\ell)}^{(\beta_1 \beta_2)(\beta_3 \beta_4)} \bigg \langle \sigma_{\alpha_1} \sigma_{\alpha_2} \left(z_{\beta_1}z_{\beta_2} - K^{(\ell)}_{\beta_1 \beta_2} \right ) \bigg \rangle_{K^{(\ell)}}  \bigg \langle \sigma_{\alpha_3} \sigma_{\alpha_4} \left(z_{\beta_3}z_{\beta_4} - K^{(\ell)}_{\beta_3 \beta_4} \right ) \bigg \rangle_{K^{(\ell)}}
\label{eq:VRecursion}
\end{align}
where $V_{(\ell)}^{(\beta_1 \beta_2)(\beta_3 \beta_4)} \equiv \sum_{\gamma_1, \dots, \gamma_4 \in \mathcal{D}} K_{(\ell)}^{\gamma_1 \beta_1} K_{(\ell)}^{\gamma_2 \beta_2} K_{(\ell)}^{\gamma_3 \beta_3} K_{(\ell)}^{\gamma_4 \beta_4} V^{(\ell)}_{(\gamma_1 \gamma_2)(\gamma_3 \gamma_4)}$ and we have used the abbreviation $\langle i j \cdots k \rangle_{K^{(\ell)}} \equiv \langle \sigma_{\alpha_i} \sigma_{\alpha_j} \cdots \sigma_{\alpha_k} \rangle_{K^{(\ell)}}$, with all activations and preactivations on the RHS at layer $\ell$.

\subsection{Depth dependence of $V$ for various activations}
\label{sec:V}

While Eq.~(\ref{eq:VRecursion}) is somewhat opaque for four general inputs, it simplifies greatly in various limits. First, for linear activations $\sigma(z) = z$, note that the term in square brackets has precisely the index structure of a \emph{connected} 4-point correlator of preactivations $z^{(\ell)}$, which vanishes in a Gaussian expectation. Therefore, for linear activations, the recursion for $V$ is homogeneous, $V^{(\ell+1)} \propto C_W^2 V^{(\ell)}$. At criticality with $C_W = 1$, $V$ is thus independent of $\ell$ and equal to its value at the first layer,
\begin{align}
V^{(\ell)}_{(\alpha_1 \alpha_2)(\alpha_3 \alpha_4)} = V^{(1)}_{(\alpha_1 \alpha_2)(\alpha_3 \alpha_4)} & = 
-\frac{1}{n^2} \bigg [ (\mathbf{x}_{\alpha_1} \cdot \mathbf{x}_{\alpha_3}) (\mathbf{x}_{\alpha_2} \cdot \mathbf{x}_{\alpha_4}) + (\mathbf{x}_{\alpha_1} \cdot \mathbf{x}_{\alpha_4}) (\mathbf{x}_{\alpha_2} \cdot \mathbf{x}_{\alpha_3}) \bigg ] \nonumber \\
& = - \bigg [ K^*_{\alpha_1 \alpha_3}  K^*_{\alpha_2 \alpha_4} + K^*_{\alpha_1 \alpha_4}  K^*_{\alpha_2 \alpha_3} \bigg ] \qquad ({\rm linear}).
\label{eq:VLinear}
\end{align}
This is simply the leading-order non-Gaussianity from the exact preactivation distribution derived in Sec.~\ref{sec:linear}; since the exact distribution is depth-independent, its connected fourth moment is necessarily depth-independent too. In fact, it is possible to make $V^{(\ell)}$ vanish identically for a finite-width linear network by drawing the first-layer weights i.i.d.\ from a \emph{Gaussian} distribution with $C_W = 1$, which we will refer to as ``mixed'' initializations. In that case, the initial condition for the $V$ recursion is $V^{(1)} = 0$~\cite{Roberts:2021fes}, and depth-independence from subsequent orthogonal weight layers ensures $V^{(\ell)} = 0$.

To see the general behavior of the $V$ recursion for nonlinear activation functions, it suffices to consider a single input $\alpha_1 = \alpha_2 = \alpha_3 = \alpha_4 = \alpha$, where we abbreviate $V_{(\alpha \alpha)(\alpha \alpha)} \equiv V$ and $K_{\alpha \alpha} \equiv K$. In that case, evaluating the remaining expectations as in~\cite{Roberts:2021fes}, the single-input recursion is
\be
V^{(\ell+1)} = C_W^2 \bigg [\langle \sigma^4 \rangle_{K^{(\ell)}} - 3 \langle \sigma^2 \rangle^2_{K^{(\ell)}} \bigg ] + \chi_\parallel^2 (K^{(\ell)}) V^{(\ell)},
\ee
which has an additional factor of 3 in the second term in brackets compared to the case for Gaussian weights. 

Consistent with previous studies~\cite{pennington2017resurrecting, pennington2018emergence}, nothing special happens for orthogonal initializations paired with ReLU activations. In that case, at the fixed point $K = K^*$, the Gaussian expectations are $\langle \sigma^2 \rangle = K^*/2$ and $\langle \sigma^4 \rangle = 3 (K^*)^2/2$. With the critical values $C_W = 2$ and $\chi_\parallel = 1$, the recursion for the normalized 4-point vertex (see Eq.~(\ref{eq:VNormDef})) has the solution
\be
\widetilde{V}^{(\ell)} = 3\ell - 5 \qquad {\rm (ReLU)},
\label{eq:VReLU}
\ee
which grows linearly with depth. The slope is 3 compared to 5 for Gaussian weights, which makes the fluctuations overall a bit smaller but does not qualitatively change any behavior during training, as we show in Appendix~\ref{app:VarTraining}.

In stark contract, the orthogonal initializations exhibit similar behavior to linear networks when paired with nonlinear activation functions in the $K^* = 0$ universality class, such as tanh. Because the fixed point of the kernel is at zero, asymptotically all of the preactivations will be small (but not exponentially small, due to the critical tuning), and we may linearize the preactivations around the fixed point to compute the expectations. Using the Taylor expansion $\tanh(z) = z - \frac{z^3}{3} +  \mathcal{O}(z^5)$, and working to $\mathcal{O}(K^3)$, we have
\begin{align}
\langle \sigma^4 \rangle_{K} & \approx \left \langle \left ( z - \frac{z^3}{3}\right)^4 \right \rangle_K  = 3K^2 - 20 K^3 +  \mathcal{O}(K^4), \\
\langle \sigma^2 \rangle_K^2 & \approx \left \langle \left ( z - \frac{z^3}{3}  \right)^2 \right \rangle_K^2 = K^2 - 4K^3 + \mathcal{O}(K^4).
\end{align}
Similarly, at criticality with $C_W = 1$, linearizing the susceptibility yields \cite{Roberts:2021fes}
\begin{equation}
\chi_\parallel(K) = 1 - 4 K + 17 K^2 + \mathcal{O}(K^3).
\end{equation}
Linearizing the kernel around the fixed point, $K^{(\ell)} = K^* + \Delta K^{(\ell)}$ with $K^* = 0$, the $V$ recursion becomes
\be
\label{eq:VRecursionTanh}
V^{(\ell +1)} = V^{(\ell)} [ 1 - 8 \Delta K^{(\ell)} + 50 (\Delta K^{(\ell)})^2] - 8 (\Delta K^{(\ell)})^3 + \mathcal{O}(\Delta K^{(\ell)})^4.
\ee
Since the kernel recursion is the same for Gaussian and orthogonal initializations, we may use the asymptotic solution for the kernel $\Delta K^{(\ell)} = 1/(2\ell) + \mathcal{O}(1/\ell^2)$ as in Ch.~5.3.3.\ of Ref.~\cite{Roberts:2021fes}. With some benefit of hindsight, we take the ansatz $V^{(\ell)} = c/\ell^2$. The left-hand side of the recursion (\ref{eq:VRecursionTanh}) is 
\[ \frac{c}{(\ell+1)^2} = \frac{c}{\ell^2} - \frac{2c}{\ell^3} + \mathcal{O}(1/\ell^4),\]
while the right-hand side of Eq.~(\ref{eq:VRecursionTanh}) reduces to 
\[ \frac{c}{\ell^2} - \frac{4c+1}{\ell^3} + \mathcal{O}(1/\ell^4).\]
Equating these to leading order in $1/\ell$ fixes $c = -1/2$, which yields a normalized correlator
\be
\widetilde{V}^{(\ell)} \equiv \frac{V^{(\ell)}}{(\Delta K^{(\ell)})^2} = -2 \qquad ({\rm tanh}).
\label{eq:Vtanh}
\ee
Remarkably, $\widetilde{V}^{(\ell)}$ for tanh is depth-independent to leading order in $\ell/n$, and takes the same value as for a linear activation function, as can be seen from the single-input limit of Eq.~(\ref{eq:VLinear}).

\begin{figure}[t!]
\centering
\includegraphics[width=0.455\textwidth]{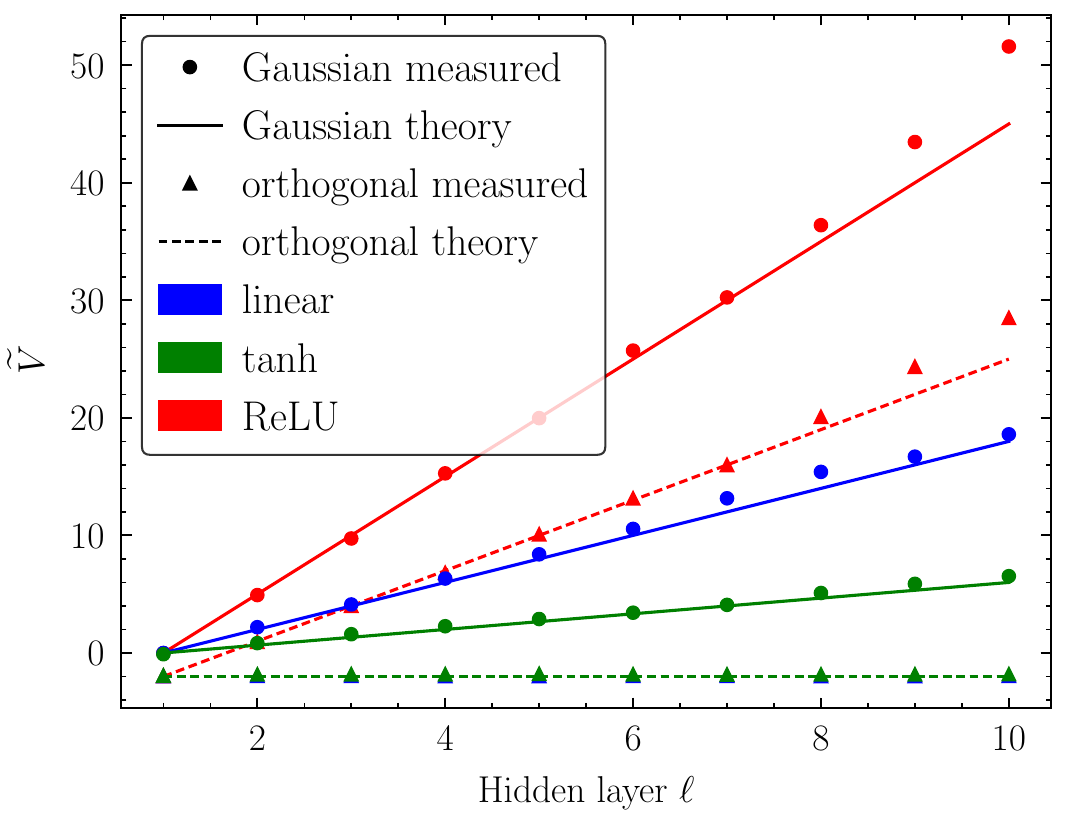}
\includegraphics[width=0.48\textwidth]{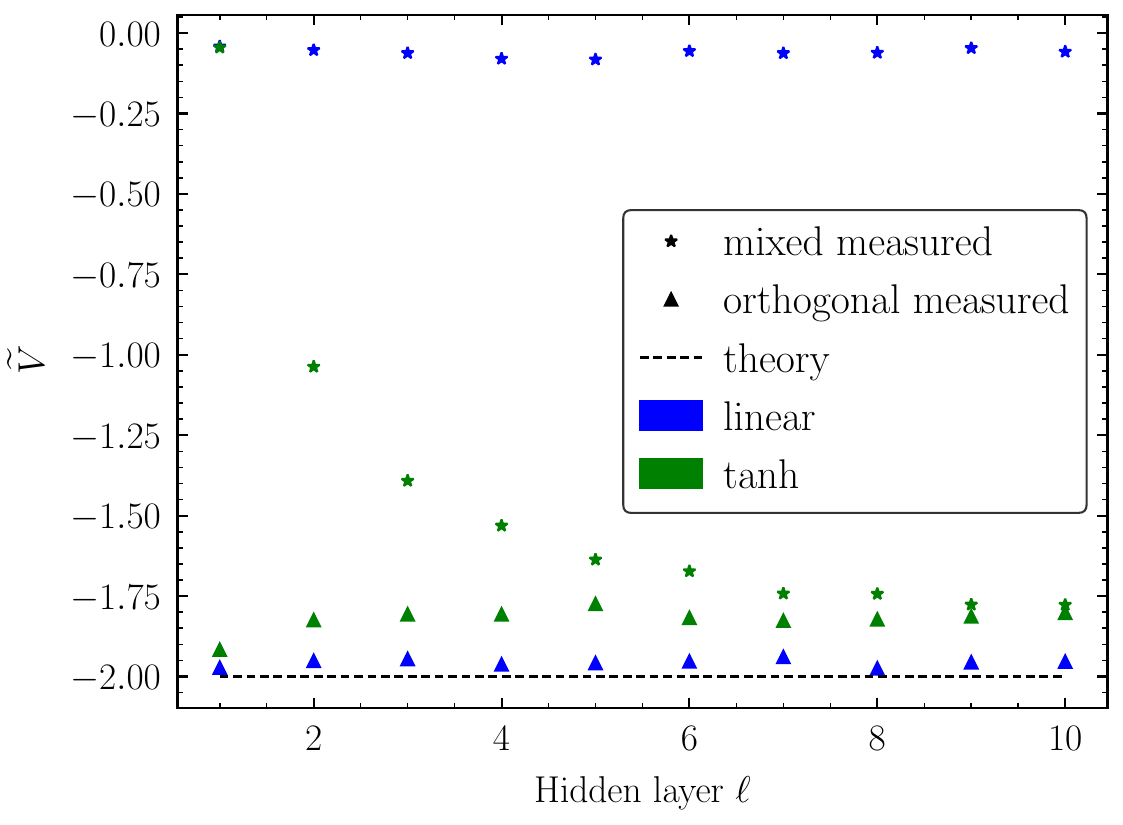}
\caption{Normalized single-input vertex $\widetilde{V}$ with $n=100$, in an ensemble of 1000 networks. Measured correlators are shown by dots, triangles, and stars; the theory lines are derived in Sec.~\ref{sec:V} for orthogonal and in Ch.\ 5 of Ref.~\cite{Roberts:2021fes} for Gaussian. \textbf{Left:} Gaussian and orthogonal initializations, with linear, ReLU, and tanh activations. \textbf{Right:} orthogonal and mixed initializations for linear and tanh activations.}
\label{fig:Vcorrelators}
\end{figure}

We have verified Eqs.~(\ref{eq:VLinear}), (\ref{eq:VReLU}), and (\ref{eq:Vtanh}) with empirical measurements over an ensemble of MLPs. We initialized an ensemble of 1000 rectangular networks of width $n = 100$ and hidden depth $L-1 = 10$ with linear, ReLU, and tanh activations, and measured the single-input $V$ for both Gaussian and orthogonal initializations. The input was taken to be a random vector of length $n$ with entries drawn uniformly from [0, 1], and we used the same input for each network configuration. The results are shown in Fig.~\ref{fig:Vcorrelators} (left) along with the theoretical results for Gaussian initializations from Ref.~\cite{Roberts:2021fes}. We find excellent agreement between measurement and theory for both initializations at small depth, with deviations at large depth arising from next-to-leading-order $\ell^2/n^2$ corrections. These corrections appear to be largest for ReLU activations, while tanh activations provide a good fit even out to $\ell = 10$. 

Fig.~\ref{fig:Vcorrelators} (right) verifies that with mixed initializations (Gaussian in the first layer, orthogonal in subsequent layers) and linear activations, $V$ can be made to vanish identically (up to order $1/n$ corrections) for all depths, as anticipated below Eq.~(\ref{eq:VLinear}). Intriguingly, we see that with tanh activations, the mixed initializations do result in a vanishing $V^{(1)}$ (as for the linear network), but in deeper layers $\widetilde{V}$ asymptotes to $-2$. This validates our linearized analysis since it verifies that $\widetilde{V} = -2$ is indeed a fixed point of the $V$ recursion, independent of initial conditions. Even for purely orthogonal initializations, $V$ has a slow decay to the fixed point at $-2$ because of higher-order $1/\ell^2$ corrections to the asymptotic analysis.

\subsection{NLO metric}
\label{sec:NLOMetric}

To complete our analytic derivations, we also compute the next-to-leading-order (NLO) metric $G^{\{1\}}$, namely the $1/n$ piece of Eq.~(\ref{eq:Kdef}):
\be
\mathbb{E}[z_{i_i; \alpha_1}^{(\ell)}z^{(\ell)}_{i_2; \alpha_2}]  = \delta_{i_1 i_2}  \left [ K_{\alpha_1 \alpha_2}^{(\ell)} + \frac{1}{n} G^{\{1\}(\ell)} + \dots \right]
\ee
Because this correlator only involves a 2-point expectation of preactivations, the recursion will be identical to that derived in Ref.~\cite{Roberts:2021fes} for Gaussian weights, with all non-Gaussianities encoded in $V$. The recursion for the single-input NLO metric and tanh activations, expanded in powers of $1/\ell$, is (see Ch.~5.4.2 of Ref.~\cite{Roberts:2021fes})
\be
G^{\{1\}(\ell+1)} = G^{\{1\}(\ell)} \left [ 1 - 4 \Delta K^{(\ell)} + \dots \right] + V^{(\ell)} \left [ -2 + \dots \right].
\ee
Similar to the analysis of $V$, we make a power law ansatz for the depth dependence of $G^{\{1\}(\ell)}$. For the case of Gaussian weights, $G^{\{1\}(\ell)}$ turns out to be constant in $\ell$ to leading order, such that $G^{\{1\}(\ell)}/K^{(\ell)} \sim \ell/n$. However, for orthogonal weights with $V^{(\ell)} = -1/(2\ell^2)$, matching powers of $1/\ell$ yields $G^{\{1\}(\ell)} = 1/\ell$ to leading order, so that $G^{\{1\}(\ell)}$ is a true perturbative correction to the metric that does \emph{not} grow with depth:
\be
\label{eq:KandG}
K^{(\ell)} + \frac{1}{n}G^{\{1\}(\ell)} = \frac{1}{2\ell} \left(1 + \frac{2}{n} + \dots \right).
\ee

Equivalently, we can consider the ``renormalized'' perspective of Ref.~\cite{Roberts:2021fes}, where NLO metric corrections imply finite-width corrections to the initialization hyperparameters:
\be
\label{eq:G1Renorm}
G^{\{1\}(\ell+1)} = c_b^{\{1\}(\ell)} + c_W^{\{1\}(\ell)} \left [ K^{(\ell)} - 2 (K^{(\ell)})^2 + \dots \right] + G^{\{1\}(\ell)}  \left [ 1 - 4 K^{(\ell)} + \dots \right] -2 V^{(\ell)} + \dots
\ee
where $C_b^{(\ell)} =c_b^{\{0\}(\ell)} + \frac{1}{n} c_b^{\{1\}(\ell)} + \dots$ and $C_W^{(\ell)} = c_W^{\{0\}(\ell)} + \frac{1}{n} c_W^{\{1\}(\ell)} + \dots$. In the case of Gaussian weights, maintaining criticality at finite width requires fine-tuning the weight initialization at all layers, $c_W^{\{1\}(\ell)} = \frac{2}{3}$. However, for orthogonal weights, after inserting the series expansions into Eq.~(\ref{eq:G1Renorm}), matching powers of $\ell$ to $\mathcal{O}(1/\ell^2)$ requires that $c_b^{\{1\}(\ell)} = c_W^{\{1\}(\ell)} = 0$. In other words, the infinite-width hyperparameters $C_W = 1, C_b = 0$ still maintain criticality at leading order in $1/n$, independent of the depth of the network. This shows that the two perspectives are the same for orthogonal weights, where depth-independence of the initialization hyperparameters is equivalent to the depth-independence of the NLO metric correction from Eq.~(\ref{eq:KandG}).

\subsection{Relation to previous work}
\label{sec:InfoThy}
The result of Eq.~(\ref{eq:Vtanh}) offers an alternative perspective on the meaning of the ``linear regime'' of critically-initialized networks. Early work in Ref.~\cite{saxe2013exact} showed that deep orthogonal networks have depth-independent training times. Ref.~\cite{pennington2017resurrecting} studied the singular value spectrum of the input-output Jacobian, and found that by tuning to criticality, as $K^* \to 0$ the fourth moment of the singular value spectrum becomes depth-independent and the maximum singular value approaches its value for a linear orthogonal network. The interpretation was that since the preactivations were so small, most of the neurons in the network were operating as if the activation function were linear; this is precisely the sense in which Eq.~(\ref{eq:Vtanh}) was derived by an asymptotic analysis which linearized the activation function about zero. Our results support this interpretation because the single-input $\widetilde{V}$ takes the same value for critical tanh networks as for critical linear networks. Indeed, as noted in Ref.~\cite{Roberts:2021fes}, the variance of the input-output Jacobian is controlled by a 2-input component of the full multi-input $V^{(\ell)}_{(\alpha_1 \alpha_2)(\alpha_3 \alpha_4)}$. In Appendix~\ref{app:MultiV}, we show measurements of $\widetilde{V}^{(\ell)}_{(\alpha_1 \alpha_2)(\alpha_3 \alpha_4)}$ for various choices of sample indices, all of which exhibit the same depth-independence. These results strongly suggest that the \emph{entire} normalized 4-input tensor $\widetilde{V}^{(\ell)}_{(\alpha_1 \alpha_2)(\alpha_3 \alpha_4)}$ is depth-independent. We plan to study this further analytically in future work.

Finally, we note that $V < 0$ is somewhat unusual in light of the discussion of optimal depth-to-width ratios from tripartite mutual information considerations in Appendix A of Ref.~\cite{Roberts:2021fes}. In Gaussian networks, $V > 0$ at all layers, which corresponds to a redundant storing of mutual information among neurons in the final layer. This leads to a competition with the growth of mutual information with depth at large width, resulting in a theoretically-optimal aspect ratio. By contrast, with $V < 0$ in orthogonal tanh networks, mutual information is stored synergistically, and maximum-entropy arguments do not lead to an optimal aspect ratio. As we will see in the measurements and numerical experiments which follow, this suggests that the aspect ratio $L/n$ does not determine the behavior of orthogonal networks, but rather that the behavior depends only weakly on $L$ and is largely independent of $n$. We leave to future work the theoretical implications of this result for feature learning.

\section{Measurement of NTK statistics}
\label{sec:NTK}

While the results of Sec.~\ref{sec:critical} provide some evidence that orthogonal weights reduce network fluctuations, they only concern the statistics of preactivations at initialization and cannot yet be directly linked to training and/or generalization. To understand the behavior of the model during training, we will study the complete set of leading-order higher-point NTK correlators which are only nonvanishing at finite width and, according to the arguments of Ref.~\cite{Roberts:2021fes}, govern the evolution of the NTK during training and thus the phenomenon of feature learning at finite width. Sec.~\ref{sec:NTKdef} reviews the relevant results from~\cite{Roberts:2021fes} in order to keep this paper self-contained, while Sec.~\ref{sec:NTKmeas} reports the results of our measurements of the relevant correlators.

\subsection{NTK and learning rate definitions}
\label{sec:NTKdef}

The NTK at layer $\ell$ is defined (in the notation of~\cite{Roberts:2021fes}) as
\be
\widehat{H}^{(\ell)}_{i_1 i_2; \alpha_1 \alpha_2} \equiv \sum_{\mu, \nu} \lambda_{\mu \nu} \frac{dz^{(\ell)}_{i_1; \alpha_1}}{d\theta_\mu} \frac{dz^{(\ell)}_{i_2; \alpha_2}}{d\theta_\nu},
\ee
where $\mu$ indexes the complete set of model parameters $\theta_\mu$ (i.e.\ weights and biases) and $\lambda_{\mu \nu}$ is the learning rate tensor. The hat on $\widehat{H}$ emphasizes that it is a stochastic variable at finite width. Under full batch gradient descent, the parameters are updated according to
\be
\theta_\mu \leftarrow \theta_\mu - \eta \sum_{\nu} \lambda_{\mu \nu} \frac{d\mathcal{L}_\mathcal{A}}{d\theta_\nu},
\label{eq:GDUpdate}
\ee
where $\eta$ is the global learning rate, and $\mathcal{L}_\mathcal{A}$ is the loss function evaluated on the training set $\mathcal{A} \subset \mathcal{D}$. We will adopt the prescription of~\cite{Roberts:2021fes} and use the learning rate tensor to scale the weight and bias learning rates differently:
\be
\label{eq:lambdadef}
\lambda_{b^{(\ell)}_{i_1} b^{(\ell)}_{i_2}} = \delta_{i_1 i_2}\lambda_b^{(\ell)}, \qquad \lambda_{W^{(\ell)}_{i_1 j_1}W^{(\ell)}_{i_2 j_2}} = \delta_{i_1 i_2}\delta_{j_1 j_2} \frac{\lambda_W^{(\ell)}}{n},
\ee
where we have also allowed for a layer dependence to the learning rates. With the learning rate tensor (\ref{eq:lambdadef}), the NTK obeys a forward iteration equation
\begin{align}
\widehat{H}^{(\ell+1)}_{i_1 i_2; \alpha_1 \alpha_2} & = \delta_{i_1 i_2} \left [ \lambda_{b}^{(\ell+1)} + \lambda_{W}^{(\ell+1)} \left ( \frac{1}{n} \sum_{j} \sigma^{(\ell)}_{j_1; \alpha_1} \sigma^{(\ell)}_{j_2; \alpha_2} \right) \right] \nonumber \\
& + \sum_{j_1, j_2} W^{(\ell+1)}_{i_1 j_1}W^{(\ell+1)}_{i_2 j_2} \sigma'^{(\ell)}_{j_1; \alpha_1} \sigma'^{(\ell)}_{j_2; \alpha_2} \widehat{H}^{(\ell)}_{j_1 j_2; \alpha_1 \alpha_2},
\label{eq:NTKforward}
\end{align} 
which can be derived by substituting the definition of $\widehat{H}$ into the MLP forward equation~(\ref{eq:zforward}) and evaluating the derivatives using the chain rule.

As is well-known in the literature~\cite{jacot2018neural,arora2019exact,lee2019wide}, at infinite width the NTK is frozen during training and becomes deterministically equal to its mean,
\be
\E [ \widehat{H}^{(\ell)}_{i_1 i_2; \alpha_1 \alpha_2} ] \equiv \delta_{i_1 i_2} \Theta^{(\ell)}_{\alpha_1 \alpha_2} + \mathcal{O}(1/n),
\ee
which depends on the initialization hyperparameters and learning rate tensor. As with the kernel, the frozen NTK takes the same value with Gaussian and orthogonal weights~\cite{huang2021neural}. To see the distinction between the two initializations, we will be interested in the fluctuations
\be
\widehat{\Delta H}^{(\ell)}_{i_1 i_2; \alpha_1 \alpha_2} \equiv \widehat{H}^{(\ell)}_{i_1 i_2; \alpha_1 \alpha_2} - \delta_{i_1 i_2} \Theta^{(\ell)}_{\alpha_1 \alpha_2},
\ee
which appear in two correlators to leading order in $1/n$:\footnote{While the tensor decomposition of these correlators was derived in~\cite{Roberts:2021fes} for the case of Gaussian weights, we have verified numerically that the same structure holds for orthogonal weights, analogous to the case of $V$ in Sec.~\ref{sec:critical}.}

\begin{align}
\label{eq:ABdef}
\E \big [\widehat{\Delta H}^{(\ell)}_{i_1 i_2; \alpha_1 \alpha_2} \widehat{\Delta H}^{(\ell)}_{i_3 i_4; \alpha_3 \alpha_4}\big ] & \equiv \frac{1}{n} \bigg [ (12)(34)_i A^{(\ell)}_{\alpha_1 \alpha_2 \alpha_3 \alpha_4} + (13)(24)_i B^{(\ell)}_{\alpha_1 \alpha_3 \alpha_2 \alpha_4} + (14)(23)_i B^{(\ell)}_{\alpha_1 \alpha_4 \alpha_2 \alpha_3} \bigg ], \\
\label{eq:DFdef}
\E \big [z^{(\ell)}_{i_1; \alpha_1} z^{(\ell)}_{i_2; \alpha_2} \widehat{\Delta H}^{(\ell)}_{i_3 i_4; \alpha_3 \alpha_4} \big] & \equiv \frac{1}{n}\bigg [ (12)(34)_i D^{(\ell)}_{\alpha_1 \alpha_2 \alpha_3 \alpha_4} + (13)(24)_i F^{(\ell)}_{\alpha_1 \alpha_3 \alpha_2 \alpha_4} + (14)(23)_i F^{(\ell)}_{\alpha_1 \alpha_4 \alpha_2 \alpha_3} \bigg ],
\end{align}
with $(ab)(cd)_i \equiv \delta_{i_a i_b} \delta_{i_c i_d}$ as in Sec.~\ref{sec:preliminaries}. Recursion relations for $A$, $B$, $D$, and $F$ are derived in~\cite{Roberts:2021fes} for Gaussian weights; we leave a parallel derivation for orthogonal weights to future work, and focus on empirical measurements of these correlators as we describe below. We may apply similar dimensional analysis reasoning as in Sec.~\ref{sec:preliminaries} to define $\widehat{H}$ as a second dimensionful quantity, and consider the dimensionless normalized correlators
\begin{align}
\widetilde{A}^{(\ell)} \equiv \frac{A^{(\ell)}}{(\Theta^{(\ell)})^2}, & \qquad \widetilde{B}^{(\ell)} \equiv \frac{B^{(\ell)}}{(\Theta^{(\ell)})^2}; \\
\widetilde{D}^{(\ell)} \equiv \frac{D^{(\ell)}}{K^{(\ell)}\Theta^{(\ell)} }, & \qquad \widetilde{F}^{(\ell)} \equiv \frac{F^{(\ell)}}{K^{(\ell)}\Theta^{(\ell)}},
\end{align}
which all have equal powers of $z$ and/or $\widehat{H}$ in the numerator and denominator.\footnote{Note that to leading order in $1/n$, these normalized correlators take the same value whether $K$ or $G$ is used in the denominator, and likewise $\Theta$ or $\widehat{H}$.}

The training dynamics of MLP architectures are governed by additional correlators at leading order in $1/n$ that involve the descendants of the NTK obtained by expanding a gradient descent update to higher order in the learning rate $\eta$. The dNTK\footnote{Since the dNTK mixes learning rates at different layers, and we are allowing the weight and bias learning rates to change with layer, we have written $\lambda_{\mu_1 \nu_1}^{(\ell_1)}$ to emphasize that the parameter indices $\mu_1$ and $\nu_1$ belong to layer $\ell_1$.}  \begin{equation}
\widehat{d H}^{(\ell)}_{i_0 i_1 i_2; \alpha_0 \alpha_1 \alpha_2} = \sum_{\ell_1, \ell_2 = 1}^{\ell}\sum_{\mu_1, \nu_1, \mu_2, \nu_2} \lambda_{\mu_1 \nu_1}^{(\ell_1)}\lambda_{\mu_2 \nu_2}^{(\ell_2)} \frac{d^2 z_{i_0; \alpha_0}^{(\ell)}}{d\theta_{\mu_1}^{(\ell_1)}d\theta_{\mu_2}^{(\ell_2)}}\frac{d z_{i_1; \alpha_1}^{(\ell)}}{d\theta_{\nu_1}^{(\ell_1)}}\frac{d z_{i_2; \alpha_2}^{(\ell)}}{d\theta_{\nu_2}^{(\ell_2)}}
\end{equation}
acts as a metafeature, updating the NTK at order $\eta$ and the preactivations at order $\eta^2$. Likewise, there are two ddNTK meta-metafeatures, $\widehat{dd_I H}$ and $\widehat{dd_{II} H}$ -- each carrying four sample indices and involving six derivatives of the preactivations\footnote{Schematically, $\widehat{dd_I H} \sim \lambda^3 \frac{d^3 z}{d\theta^3} \left(\frac{dz}{d\theta}\right)^3$ and $\widehat{dd_{II} H} \sim \lambda^3 \left(\frac{d^2 z}{d\theta^2}\right)^2 \left(\frac{dz}{d\theta}\right)^2$; the full expressions may be found in Ref.~\cite{Roberts:2021fes}.}  -- which update the dNTK at order $\eta$, the NTK at order $\eta^2$, and the preactivations at order $\eta^3$. The hierarchy terminates there, such that the ddNTKs are frozen at initialization and do not update to leading order in $1/n$. The perturbative expansion in the learning rate breaks down for activation functions with discontinuous derivatives such as ReLU, so we will focus on tanh activations in what follows, which are (fortunately) the activations which give the most interesting phenomenology for orthogonal initializations.

For our purposes, we are interested in the correlators at leading order in $1/n$,
\begin{align}
\label{eq:PQdef}
\E \big [\widehat{d H}^{(\ell)}_{i_0 i_1 i_2; \alpha_0 \alpha_1 \alpha_2} z^{(\ell)}_{i_3; \delta_3}\big ] & \equiv \frac{1}{n} \bigg [ (03)(12)_i P^{(\ell)}_{\alpha_0 \alpha_1 \alpha_2 \alpha_3} + (01)(23)_i Q^{(\ell)}_{\alpha_0 \alpha_1 \alpha_2 \alpha_3} + (02)(13)_i Q^{(\ell)}_{\alpha_0 \alpha_2 \alpha_1 \alpha_3} \bigg ], \\
\label{eq:Rdef}
\E \big [\widehat{dd_I H}^{(\ell)}_{i_1 i_2 i_3 i_4; \alpha_1 \alpha_2 \alpha_3 \alpha_4} \big] & \equiv \frac{1}{n}\bigg [ (12)(34)_i R^{(\ell)}_{\alpha_1 \alpha_2 \alpha_3 \alpha_4} + (13)(24)_i R^{(\ell)}_{\alpha_1 \alpha_3 \alpha_2 \alpha_4} + (14)(23)_i R^{(\ell)}_{\alpha_1 \alpha_4 \alpha_2 \alpha_3} \bigg ], \\
\label{eq:STUdef}
\E \big [\widehat{dd_{II} H}^{(\ell)}_{i_1 i_2 i_3 i_4; \alpha_1 \alpha_2 \alpha_3 \alpha_4} \big] & \equiv \frac{1}{n}\bigg [(12)(34)_i S^{(\ell)}_{\alpha_1 \alpha_2 \alpha_3 \alpha_4} + (13)(24)_i T^{(\ell)}_{\alpha_1 \alpha_3 \alpha_2 \alpha_4} + (14)(23)_i U^{(\ell)}_{\alpha_1 \alpha_4 \alpha_2 \alpha_3} \bigg ].
\end{align}
Notice that the index labeling for $\widehat{dH}$ is slightly different because the first index is distinguished, such that the tensor decomposition does not exactly follow that of a Wick contraction. Eqs.~(\ref{eq:ABdef}--\ref{eq:DFdef}) and (\ref{eq:PQdef}--\ref{eq:STUdef}) (along with Eq.~(\ref{eq:Vdef})) comprise the full set of MLP correlators which appear at $\mathcal{O}(1/n)$, all of which carry four neural indices and four sample indices. The dimensionless normalized correlators are
\begin{align}
\widetilde{P}^{(\ell)} \equiv \frac{P^{(\ell)}}{(\Theta^{(\ell)})^2}, & \qquad \widetilde{Q}^{(\ell)} \equiv \frac{Q^{(\ell)}}{(\Theta^{(\ell)})^2}; \\
\widetilde{R}^{(\ell)} \equiv \frac{R^{(\ell)}K^{(\ell)}}{(\Theta^{(\ell)})^3},  \qquad \widetilde{S}^{(\ell)} \equiv \frac{S^{(\ell)}K^{(\ell)}}{(\Theta^{(\ell)})^3}, & \qquad \widetilde{T}^{(\ell)} \equiv \frac{T^{(\ell)}K^{(\ell)}}{(\Theta^{(\ell)})^3}, & \qquad \widetilde{U}^{(\ell)} \equiv \frac{U^{(\ell)}K^{(\ell)}}{(\Theta^{(\ell)})^3}.
\end{align}
For Gaussian weights, all dimensionless correlators grow linearly with $\ell$ for tanh activations, except for $\widetilde{U}$ which decays as $1/\ell$.

\subsection{Measurements of NTK, dNTK, and ddNTK statistics at initialization}
\label{sec:NTKmeas}

\begin{figure}[t!]
\centering
\includegraphics[width=0.99\textwidth]{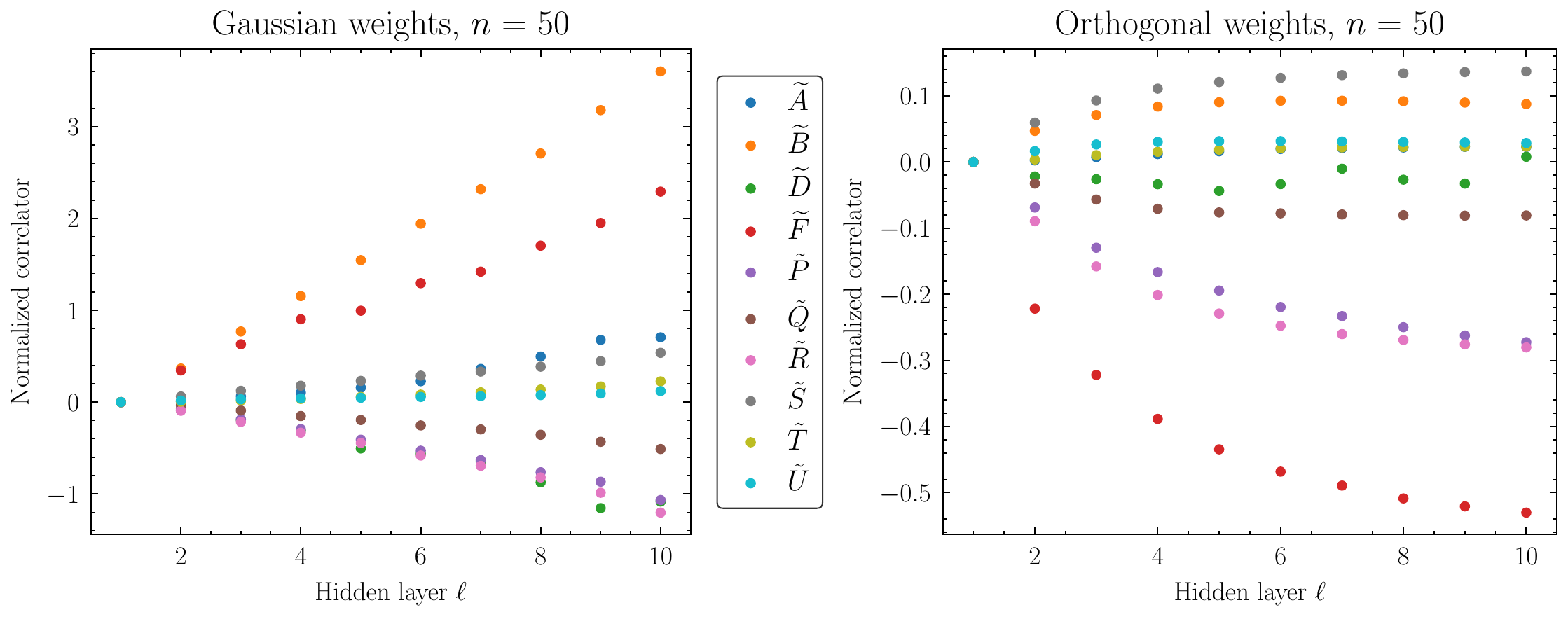}
\caption{Measurements of normalized single-input correlators for Gaussian (left) and orthogonal (right) initializations, with $n = 50$ and tanh activations, in an ensemble of 100 networks. The Gaussian correlators (except for $\widetilde{U}$) grow proportional to $\pm \ell$ as predicted by the analysis of Ref.~\cite{Roberts:2021fes}, while the orthogonal correlators are smaller in overall magnitude and begin to saturate with depth.}
\label{fig:correlatorsWide}
\end{figure}

Analogous to our measurements of $V$ in Sec.~\ref{sec:critical}, we initialized an ensemble of 100 rectangular networks of width $n = 50$ and hidden layers $L-1 = 10$ with tanh activations, and measured all single-input NTK correlators from Sec.~\ref{sec:NTKdef} for both Gaussian and orthogonal initializations. The input was taken to be a random vector of length $n$ with entries drawn uniformly from $[0, 1]$. The weight and bias learning rates which enter the definitions of $\widehat{H}$, $\widehat{dH}$, and $\widehat{dd_{I,II} H}$ were taken to be
\be
\label{eq:LRrescaling}
\lambda_b^{(\ell)} = \frac{1}{\ell}, \qquad \lambda_W^{(\ell)} = 1,
\ee
where the scaling of the bias learning rate with depth follows the ``learning rate equivalence principle'' prescription of Ref.~\cite{Roberts:2021fes} to ensure roughly equal contributions to the NTK at each layer for tanh activations.\footnote{We demonstrate in Appendix~\ref{app:VarTraining} that these prescriptions, designed to ensure that each layer yields an order-one contribution to the NTK, give essentially identical generalization at small $L/n$ compared to ordinary (non-tensorial) GD, but seem to lead to worse generalization for deep orthogonal networks. We speculate on the reasons for this behavior in Appendix~\ref{app:VarTraining}.} The results for the dimensionless correlators are shown in Fig.~\ref{fig:correlatorsWide}, using the same random input for both sets of initializations. It is clear that for the Gaussian initialization (left), all correlators except $\widetilde{U}$ are growing approximately linearly with depth, while for orthogonal initialization (right), the growth is much slower and appears to saturate at finite depth.

\begin{figure}[t!]
\centering
\includegraphics[width=0.99\textwidth]{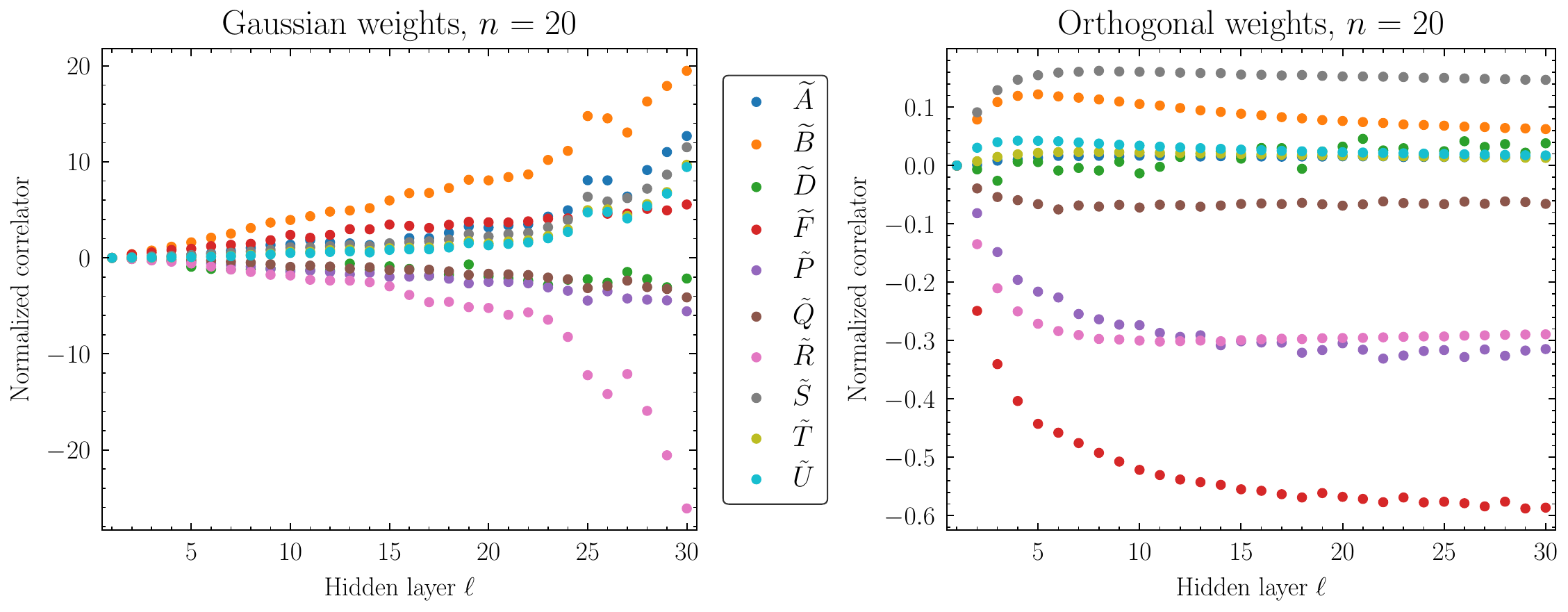}
\caption{Measurements of normalized single-input correlators for Gaussian (left) and orthogonal (right) initializations, with $n = 20$ and tanh activations, in an ensemble of 100 networks. The Gaussian correlators begin to grow exponentially in magnitude and fluctuate chaotically when $\ell \simeq n$, but the orthogonal correlators saturate at a depth $\ell \sim 20$. By $\ell = 30$, the Gaussian correlators are orders of magnitude larger than the orthogonal correlators.}
\label{fig:correlatorsNarrow}
\end{figure}

Due to memory limitations, it is difficult to measure the dNTK and ddNTK correlators at very large width and depth. However, to investigate the behavior of the orthogonal correlators further, we reduced the width to 20 and increased the number of hidden layers to $L-1 = 30$. In this regime, with $L/n \gtrsim 1$, the effective theory of the Gaussian-initialized network begins to break down and we expect to see the largest deviations between Gaussian and orthogonal initializations. The results are shown in Fig.~\ref{fig:correlatorsNarrow}. For Gaussian initializations, the correlators become noisy and begin to grow exponentially for $\ell/n \gtrsim 1$ (including $\widetilde{U}$), consistent with the effective theory arguments of Ref.~\cite{Roberts:2021fes}. However, for orthogonal weights, the correlators saturate at a depth $\ell \sim 20$, and take approximately the same asymptotic values as they do for $n = 50$ in Fig.~\ref{fig:correlatorsWide}. Intriguingly, the $A$ and $D$ correlators, whose recursions for Gaussian initializations involve the 4-point vertex $V$~\cite{Roberts:2021fes}, seem to be consistently smaller than $B$ and $F$ at all depths for orthogonal initializations, which is not the case for Gaussian initializations. While the recursions may well change for orthogonal weights (a derivation which we leave to future work), this is at least suggestive of the fact that the dominant correlators that drive feature learning (i.e. the evolution of the NTK and the dNTK) are \emph{not} the ones associated with noisy ensemble fluctuations. 

All of the above observations suggest that, unlike MLPs with Gaussian weights, orthogonal networks do \emph{not} have an $L/n$ cutoff for the perturbative behavior of correlators at initializations, but rather are only weakly sensitive to width and are roughly depth-independent for $L \gtrsim 20$ regardless of width. Empirically, the behavior of the normalized orthogonal correlators seems characteristic of a logarithmic \emph{decay} $a_0 + a_1 \log \ell/\ell$ (to be contrasted with the linear \emph{growth} of the Gaussian correlators), where a full calculation of the orthogonal initialization recursions would permit us to predict the constants $a_0$ and $a_1$ as Ref.~\cite{Roberts:2021fes} does for Gaussian initializations.

Informed by the measurements of the correlators at initialization, we conjecture that the smaller values of the ``noise'' correlators $A$ and $D$, whose recursions involve the 4-point vertex $V$, will lead to faster training for orthogonal initializations compared to Gaussian initializations for networks of the same depth and width. Furthermore, we can predict that the performance of an orthogonal network at fixed width should be roughly independent of $L$ for $L \gtrsim 20$, and generalization at the end of training should be superior to Gaussian networks in the regime $L/n \sim 1$, where Gaussian networks suffer from large fluctuations. We now proceed to test these conjectures experimentally.

\section{Training and generalization experiments}
\label{sec:MNIST}

The results of Sec.~\ref{sec:NTK} are suggestive but not persuasive, because (among other reasons) they only refer to the ensemble statistics at initialization. In this section, using the toy examples of the MNIST and CIFAR-10 classification problems, we will provide more evidence for the superior training behavior of orthogonal initializations conjectured above by comparing the final test loss on trained networks from both initializations. These experiments are very similar to those performed in Refs.~\cite{pennington2017resurrecting} and \cite{huang2021neural}, but we will make some slightly different choices in order to stay as close as possible to the theoretical framework of this paper.

\subsection{Experimental setup}
The setup of our experiments was as follows. We initialized MLP networks of hidden layer width $n=30$ or $n = 100$, tanh activations on each hidden layer, and the number of hidden layers ranging from $L-1 =1$ to $L-1 = 50$. The initial biases were set to zero and the initial weights were sampled from either Gaussian or orthogonal distributions with critical initialization $C_W = 1$. Note that an orthogonal distribution is not well-defined for the first layer because its weight matrix is not a square matrix (the input dimension is 784 for MNIST and 3072 for CIFAR-10), so instead, the first layer weights are sampled from the QR decomposition of a rectangular matrix with random Gaussian entries, as implemented by the default \texttt{Keras}~\cite{chollet2015keras} orthogonal initializer.\footnote{With an eye toward practical implications, we note that the computational complexity of QR decomposition is $\mathcal{O}(n^3)$, while filling a matrix with independent Gaussian-distributed entries is $\mathcal{O}(n^2)$, so there is a mild compute penalty for orthogonal initializations. Furthermore, as implemented in \texttt{Keras}, it is not clear if orthogonal intializations can be parallelized across GPUs for extremely large networks, though in this limit we expect Gaussian and orthogonal networks to perform very similarly.} When the weight matrix is a square matrix, this sampling algorithm results in a Haar-distributed orthogonal matrix as desired~\cite{mezzadri2007generate}. However, as discussed in Sec.~\ref{sec:V} for the case of the 4-point vertex $V$, and in Appendix~\ref{app:MultiNTK} for the NTK correlators $A$, $B$, $D$, and $F$, this choice simply affects the initial conditions of the various correlator recursions and not the large-depth asymptotics. Consequently, we expect the analysis of the previous section (in particular the depth independence for $L \gtrsim 20$) to apply. For the MNIST dataset, where most pixels are zero in the raw dataset, we performed a standard rescaling so that the entire dataset has zero mean and unit variance. The loss function was taken to be mean squared error (MSE) loss, $L = ||\mathbf{z}_\alpha^{(L)} - \mathbf{y}_\alpha||^2$ where $\mathbf{y}_\alpha$ is the 10-dimensional one-hot label encoding the class of the training point $\mathbf{x}_\alpha$.\footnote{The use of MSE loss is somewhat unconventional in the classification context, but corresponds directly to the analysis of Ref.~\cite{Roberts:2021fes} and has been shown to have essentially equivalent or even superior performance to cross-entropy loss across a wide variety of tasks~\cite{hui2020evaluation}; see Appendix~\ref{app:VarTraining} for results using cross-entropy loss.} 

For MNIST, we trained all networks with full-batch tensorial gradient descent (\ref{eq:GDUpdate}) with learning rate tensor given by Eq.~(\ref{eq:lambdadef}), weight and bias learning rate rescalings given in Eq.~(\ref{eq:LRrescaling}), and a global learning rate of $\eta = 10$.\footnote{With the conventions of Eq.~(\ref{eq:lambdadef}), this is not a parametrically large learning rate, since the weight learning rate is effectively $10/n$ which is less than 1 for both widths we consider.}  We impose implicit regularization via early stopping, using a validation set of size $10^4$, and take our final trained network to be the one corresponding to the best validation loss. We verified that for architectures of hidden depth greater than or equal to 5, $10^5$ epochs is sufficient to identify the onset of overfitting, with more training required to identify overfitting for very shallow networks: for hidden depths of 1, 2, 3, and 4, we extended training to $10^6$, $5 \times 10^5$, $3 \times 10^5$, and $3 \times 10^5$ epochs, respectively. We did not attempt to optimize the global learning rate for the best overall performance, but rather chose $\eta$ large enough for training to not exceed $10^5$ epochs for deep networks, but small enough to prevent divergences in the training loss from large step sizes. We have checked that varying $\eta$ gives qualitatively similar results for all experiments in this section. Finally, to compare the performance of trained networks across different architectures, we averaged the test loss over 10 instantiations of each network at the epoch determined by early stopping for each instantiation. For CIFAR-10, we used a validation set of size $7 \times 10^{3}$ partitioned from the training set. We used the same learning rate $\eta = 10$, but found that overfitting began later, so we trained for $3 \times 10^{5}$ epochs for hidden depths greater than or equal to 5 and $6 \times 10^6$ epochs for smaller depths. Due to the slower training times for CIFAR-10 and the compute limitations for this study, we only used 5 CIFAR-10 training runs for each architecture.

\begin{figure}[t!]
\centering
\includegraphics[width=0.53\textwidth]{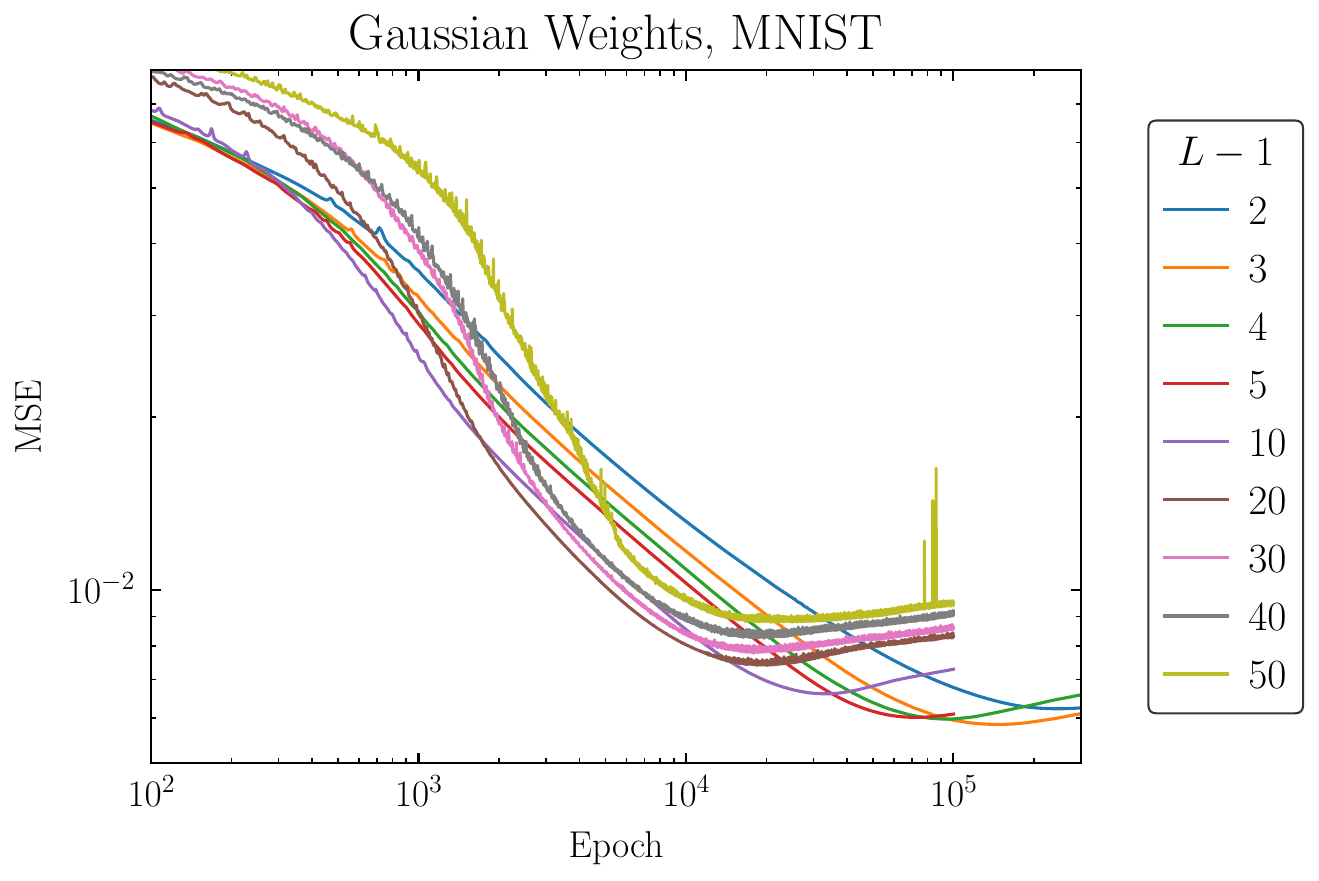}
\includegraphics[width=0.45\textwidth]{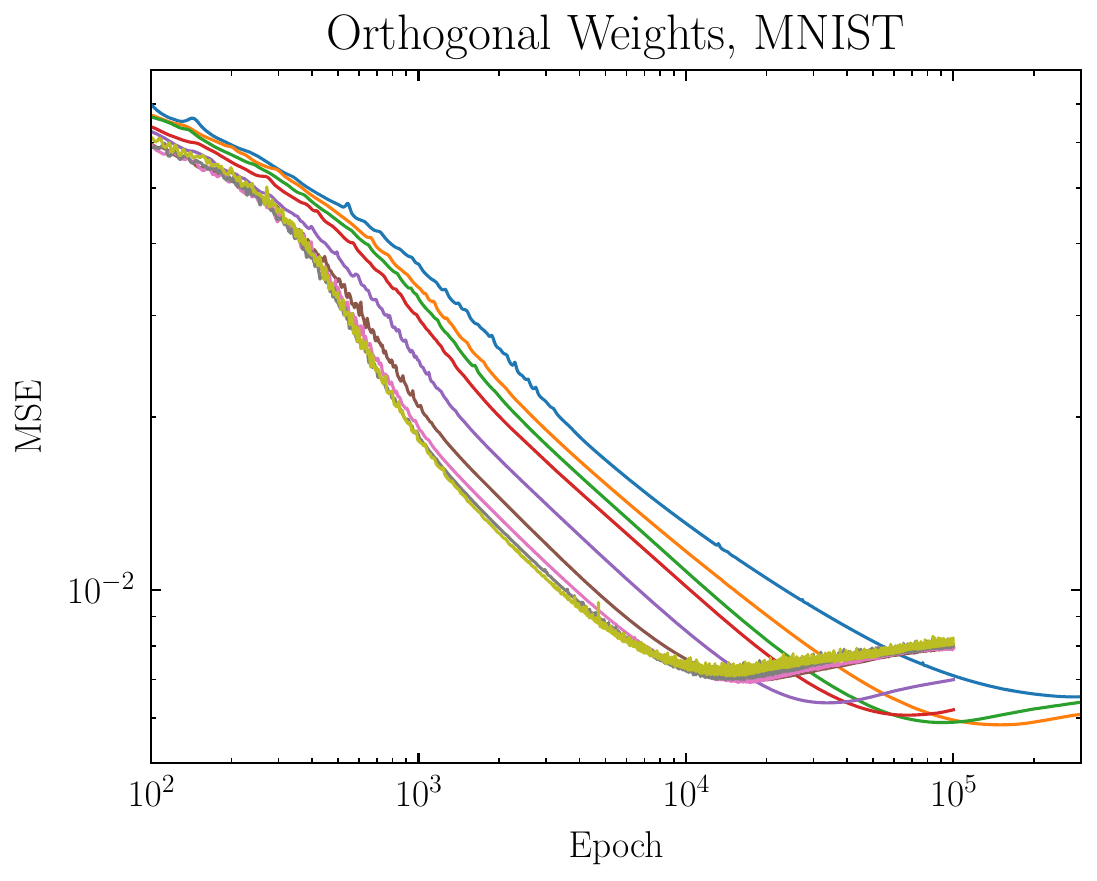}
\caption{MSE validation loss versus epoch for Gaussian (left) and orthogonal (right) initializations, with $n = 30$ and tanh activations, averaged over 10 runs on MNIST data. The Gaussian networks suffer from both slower training and worse generalization as the depth increases, while the loss curves for orthogonal networks begin to lie on top of one another for $L \gtrsim 20$.
}
\label{fig:lossesMNIST30}
\end{figure}

\begin{figure}[t!]
\centering
\includegraphics[width=0.53\textwidth]{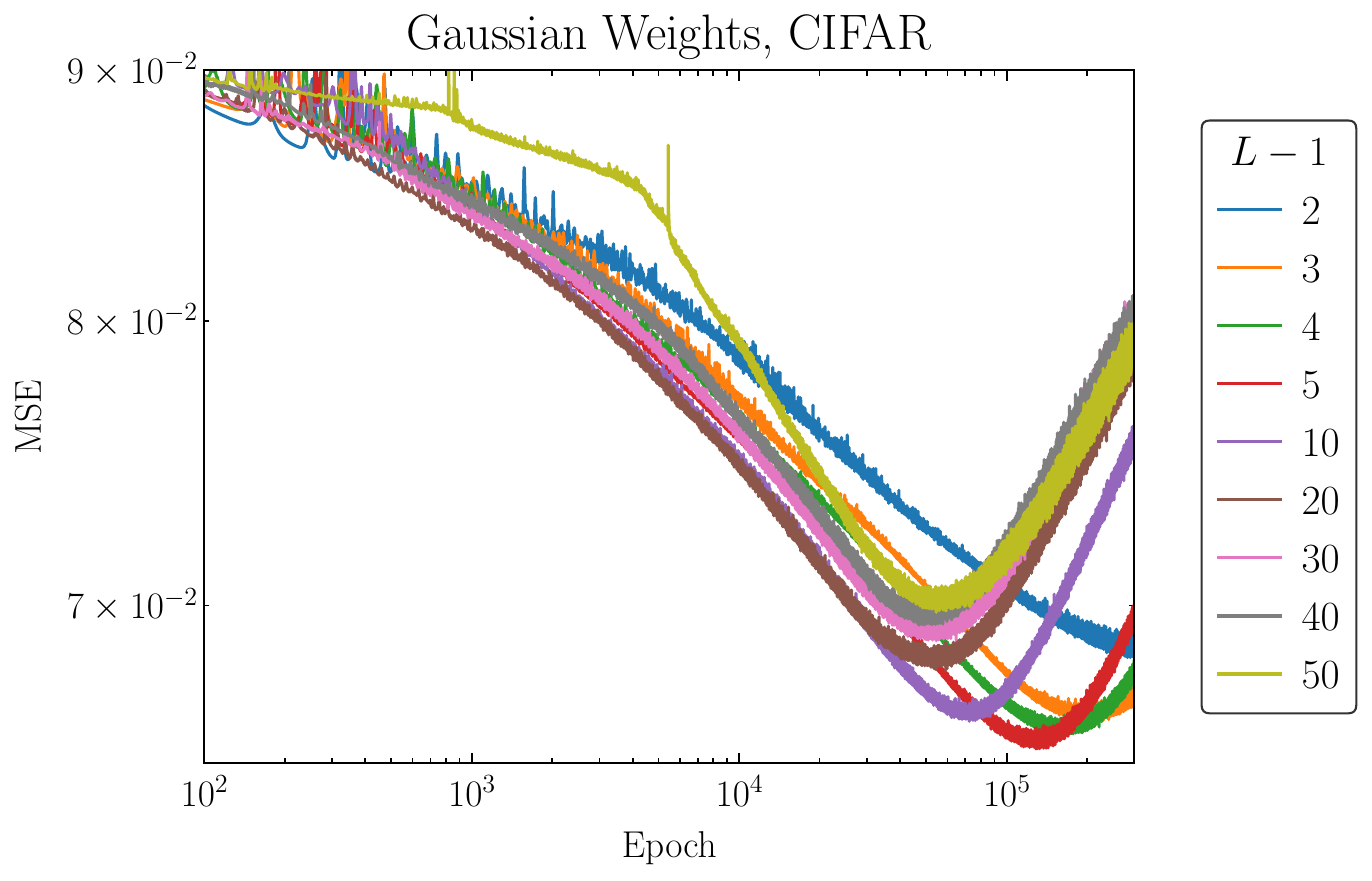}
\includegraphics[width=0.45\textwidth]{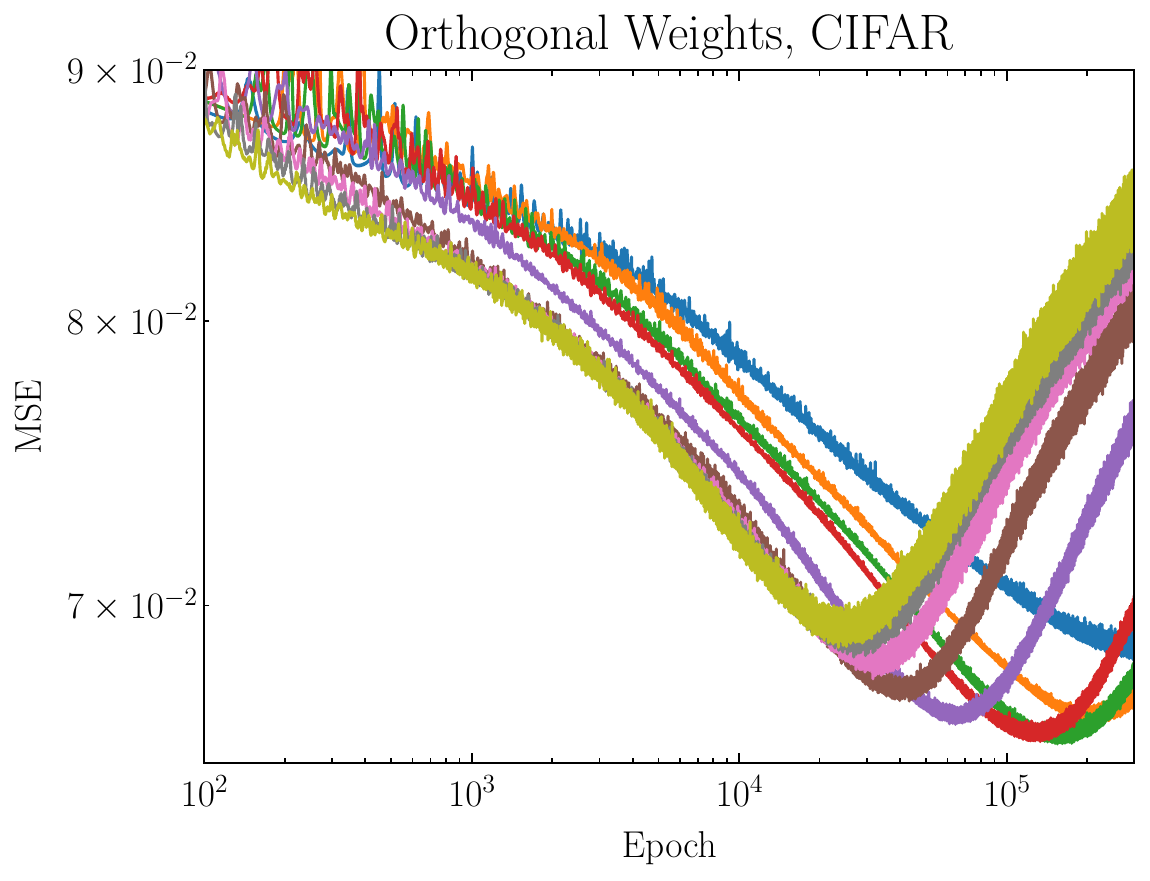}
\caption{MSE validation loss versus epoch for Gaussian (left) and orthogonal (right) initializations, with $n = 30$ and tanh activations, averaged over 5 runs on CIFAR-10 data. As with MNIST, the loss curves for Gaussian weights exhibit slower training and worse generalization at large depths compared to orthogonal initializations.}
\label{fig:lossesCIFAR30}
\end{figure}

\subsection{Results for training dynamics}
\label{sec:TrainingResults}

In Figs.~\ref{fig:lossesMNIST30} and \ref{fig:lossesCIFAR30} we show the validation loss as a function of training epoch for MNIST (averaged over 10 runs) and CIFAR-10 (averaged over 5 runs), respectively. For orthogonal initializations, the loss curves for MNIST (and to a lesser extent for CIFAR) appear to converge for depths of 20 or greater, as predicted based on the behavior of the NTK correlators as a function of depth in Sec.~\ref{sec:NTKmeas}. By contrast, for Gaussian initializations with both datasets, deeper networks appear to train more slowly and achieve a worse validation loss. Note that the non-monotonic oscillations in the loss, which are especially apparent in the CIFAR data, are characteristic of the ``edge of stability'' GD dynamics first identified in Ref.~\cite{cohen2021gradient}, and are a generic feature of full-batch GD with fixed learning rate. In general, for the same network architecture, orthogonal networks appear to train faster than Gaussian networks, as measured both by the steepness of the loss curve early in training and the epoch at which the minimum of the validation loss curve is reached. In Appendix~\ref{app:VarTraining} we verify that the superior training performance of orthogonal tanh networks holds even as the optimizer and loss function are varied, but that there only seems to be a minor improvement in performance with ReLU activations, as predicted by our analysis in Sec.~\ref{sec:V}.

\begin{figure}[t!]
\centering
\includegraphics[width=0.45\textwidth]{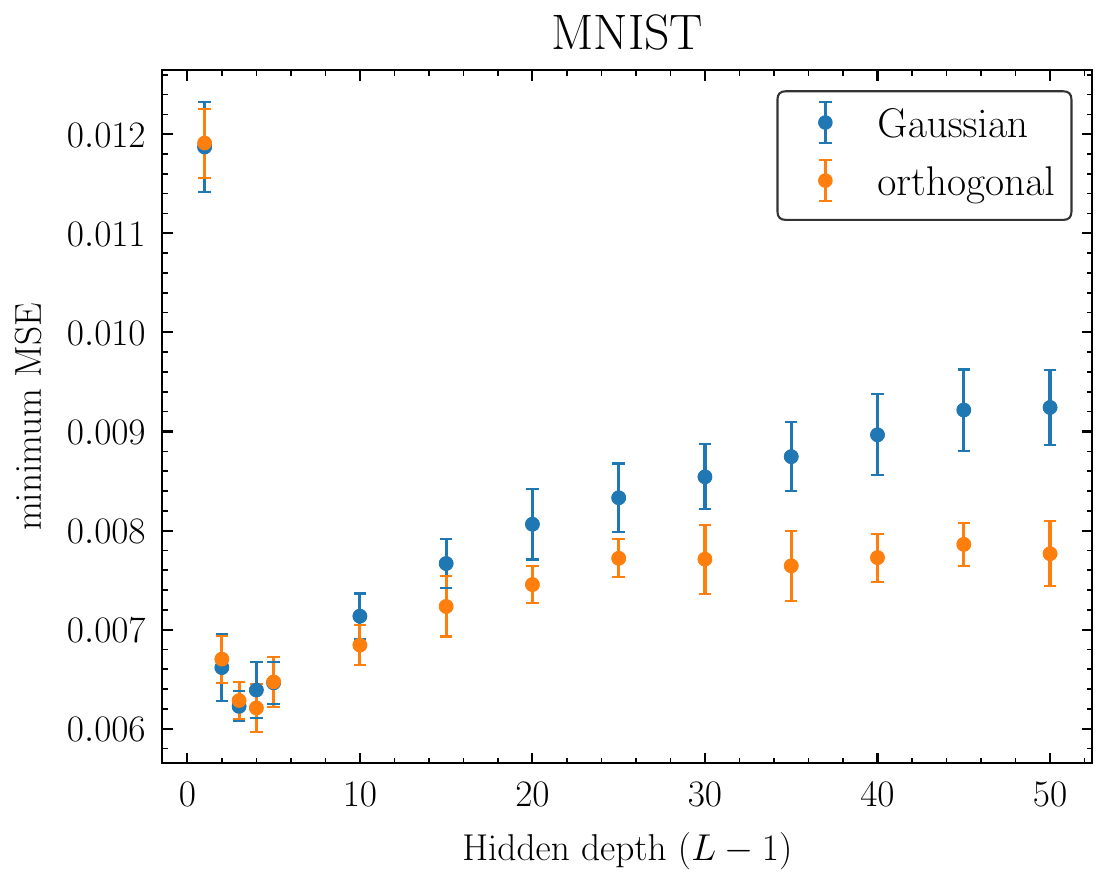}
\includegraphics[width=0.45\textwidth]{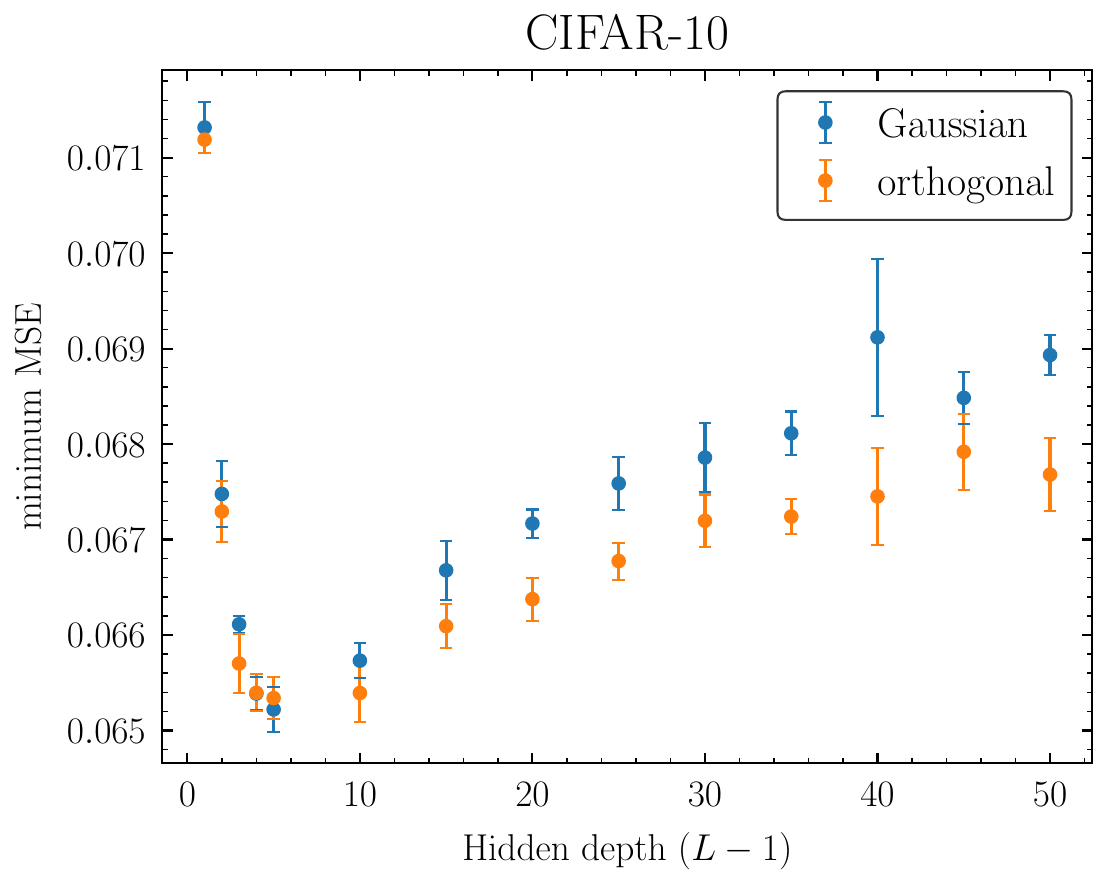}
\caption{Best MSE test loss versus depth of network for Gaussian and orthogonal initializations, with $n = 30$ and tanh activations on MNIST (left) and CIFAR-10 (right). Error bars for represent the standard deviation over 10 runs for MNIST and over 5 runs for CIFAR-10. The overall best loss is achieved at the same depth for both initializations: 3 hidden layers for MNIST, and 4 hidden layers for CIFAR-10.}
\label{fig:bestloss30}
\end{figure}

\begin{figure}[t!]
\centering
\includegraphics[width=0.45\textwidth]{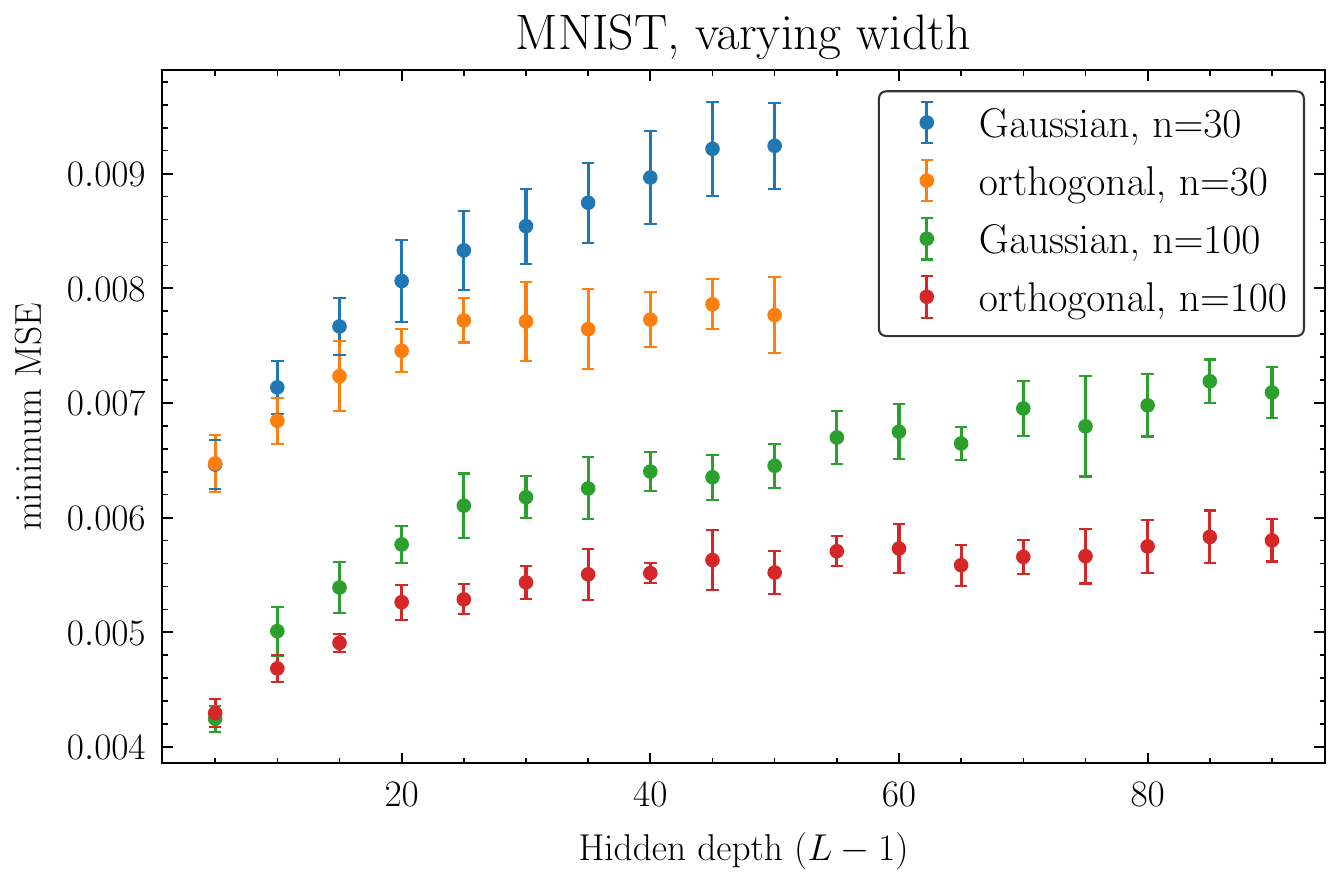}
\caption{Best MSE test loss versus depth of network for Gaussian and orthogonal initializations, with $n = 30$ and $n=100$, using tanh activations on MNIST. Error bars for represent the standard deviation over 10 runs.}
\label{fig:bestloss100}
\end{figure}

In Fig.~\ref{fig:bestloss30}, we compare the averaged best test losses for Gaussian and orthogonal initializations for MNIST (left) and CIFAR-10 (right) for a network width $n=30$, with error bars indicating the standard deviation over our runs. In Fig.~\ref{fig:bestloss100}, we compare the averaged best test losses for $n=30$ and $n=100$ for both initializations. Consistent with previous studies~\cite{pennington2017resurrecting,huang2021neural}, orthogonal initialization results in a smaller test loss (to within the quoted error bars), and hence better generalization, for \emph{all} architecture hyperparameters. Away from the infinite-width limit (when $L\gtrsim20$), the value of the best loss for MNIST is roughly independent of depth for the orthogonal initialization, despite the increasing number of model parameters, but continues to degrade with depth for Gaussian initializations. This behavior is also seen to some extent in CIFAR-10 (Fig.~\ref{fig:bestloss30}, right): the loss appears to flatten wtih depth around $L \sim 30$, though the trend at large depth is more difficult to see with the larger variance compared to MNIST. These results further validate our predictions in Sec.~\ref{sec:NTKmeas}, as the plateau in orthogonal best-loss values for MNIST begins at approximately the same depth as when the NTK correlators become depth-independent. From Fig.~\ref{fig:bestloss100}, we also see that the plateau occurs at the same depth regardless of network width. This is consistent with our observation in Sec.~\ref{sec:NTKmeas} that the magnitude of the orthogonal correlators is roughly independent of width.

For $L\lesssim20$, the results enter the regime of validity of the infinite-width approximation, so the differences between orthogonal and Gaussian initializations become increasingly small. For $n=30$, as we approach very small values of $L$, we find the best hidden depth for the MNIST classification task is $L-1=3$ for both initializations; for CIFAR-10 the best hidden depth is $L-1 = 4$ for both initializations. The information-theoretic arguments in Ref.~\cite{Roberts:2021fes} predict an optimal depth of 3.6 for width-30 tanh MLPs with an output layer of width $n_L = 10$, which is consistent with our results across both classification tasks and initializations. This is remarkable given the opposite sign of $V$ for orthogonal initializations, as discussed in Sec.~\ref{sec:InfoThy}. However, for $n=100$, the predicted optimal depth is 12, while it is clear from Fig.~\ref{fig:bestloss100} that the true optimum occurs for shallower networks. While we have demonstrated that it is possible to successfully train both orthogonal networks for $L \gtrsim 20$ (analogous to the results of Ref.~\cite{xiao2018dynamical}), since the best architecture for both widths is firmly in the $L/n \ll 1$ regime where both initializations give similar results, it is unclear if there is a practical benefit to using these very deep networks or orthogonal initializations. That said, in future work we intend to study whether the smaller correlator fluctuations in orthogonal networks lead to smaller variance in the trained network predictions, which could in fact provide a practical benefit for these initializations even in the $L/n \ll 1$ regime.

\subsection{Empirical NTK dynamics, generalization, and feature learning}

\begin{figure}[t!]
\centering
\includegraphics[width=0.45\textwidth]{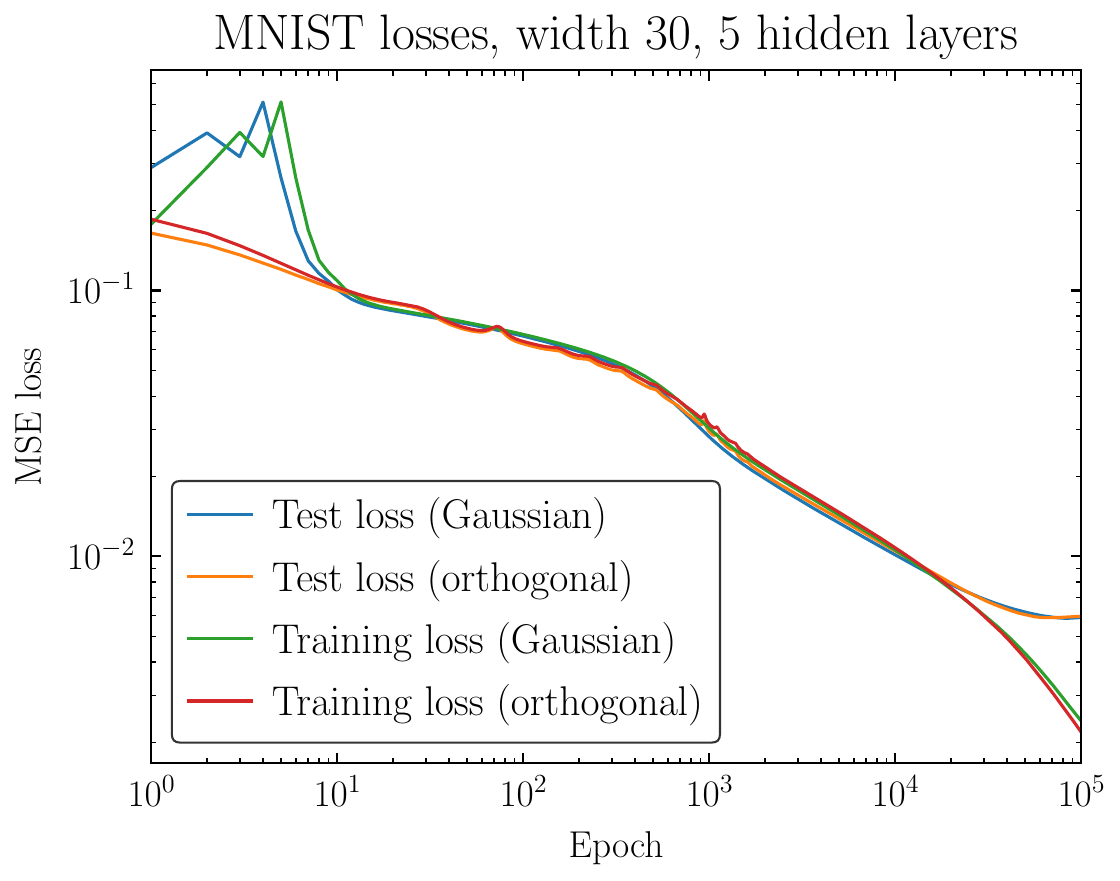}
\includegraphics[width=0.45\textwidth]{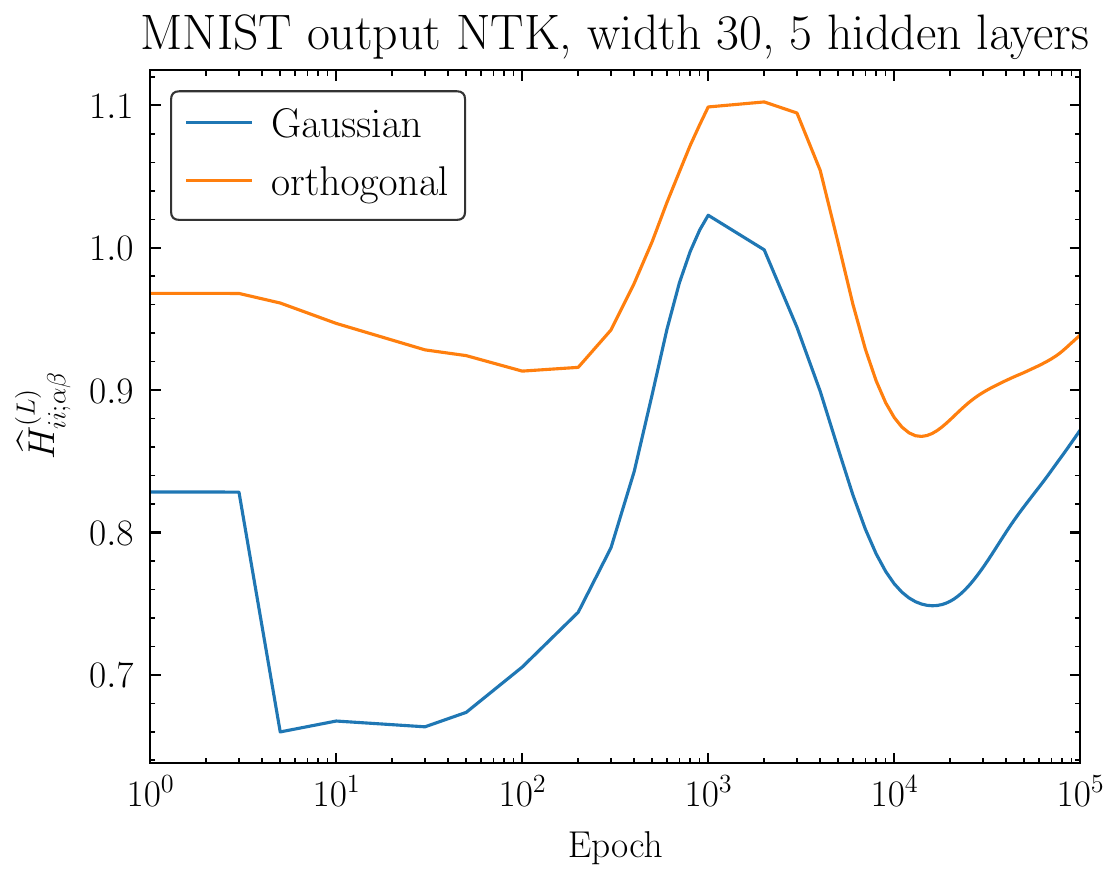} \\
\includegraphics[width=0.45\textwidth]{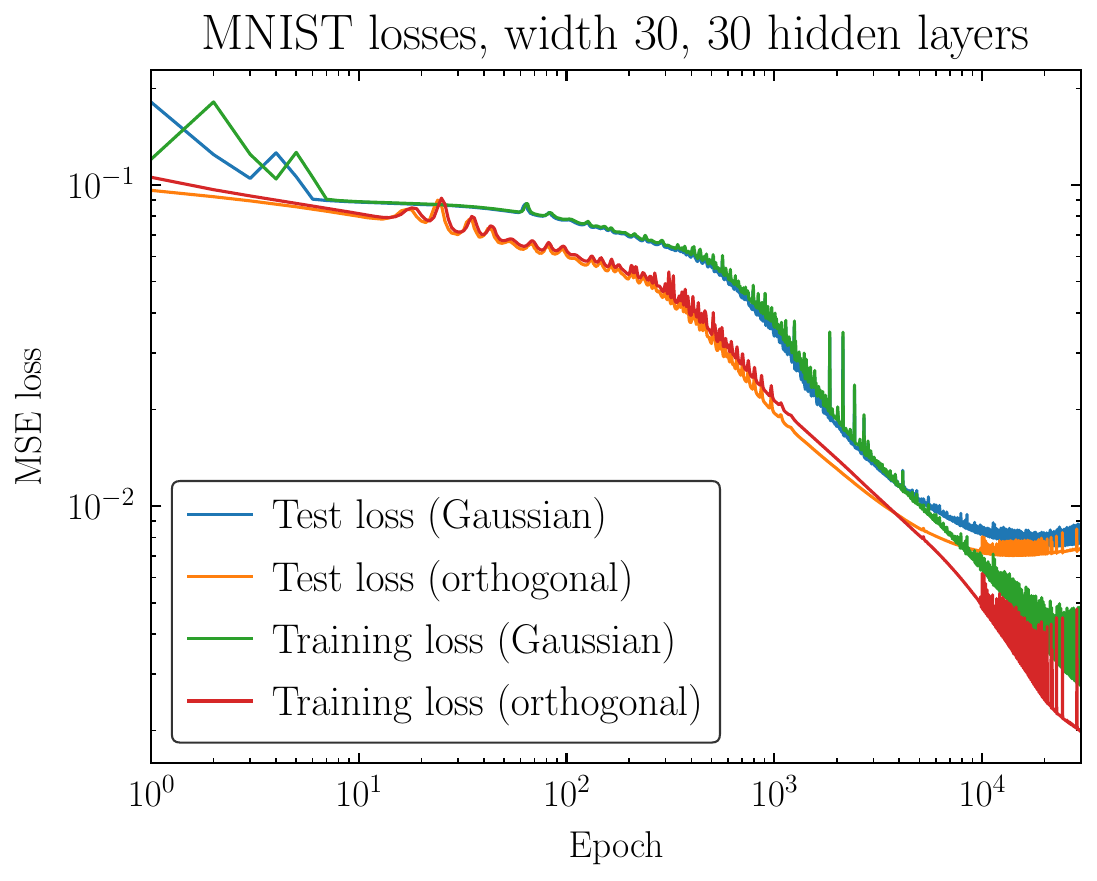}
\includegraphics[width=0.45\textwidth]{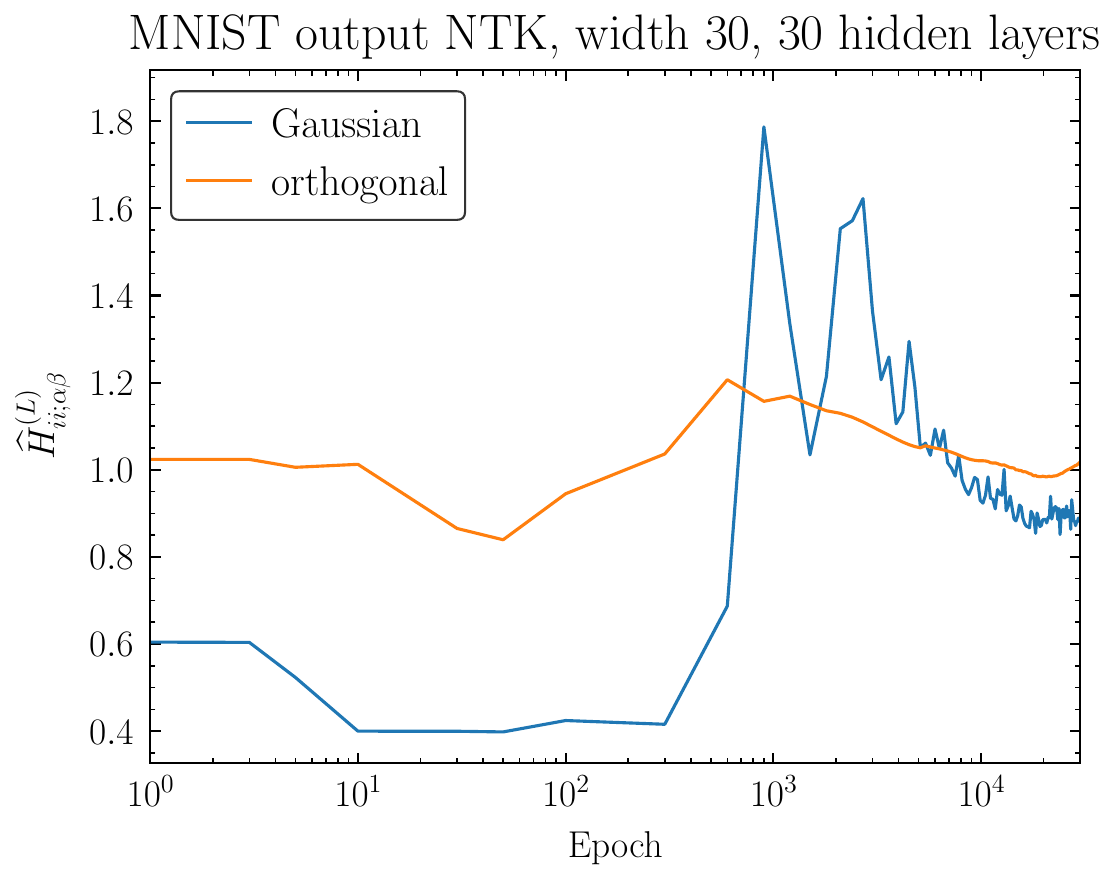}
\caption{MNIST test loss curves (left) and evolution of the principal diagonal component of the test-train NTK (right) relevant for feature learning as described in the text, for Gaussian and orthogonal networks with width $n=30$ and number of hidden layers $L-1 = 5$ (top row) and $L-1 = 30$ (bottom row). The NTK grows fastest during the epochs where the training and test losses drop fastest (around $10^3$), but the orthogonal NTK exhibits much smaller fluctuations, both during this period and at the end of training. The Gaussian and orthogonal NTKs have similar evolution for $L/n \ll 1$, consistent with the near-identical behavior of their loss curves, but the Gaussian NTK undergoes large oscillations for $L/n \simeq 1$ while the orthogonal NTK has similar evolution at both depths. For the deeper network, the orthogonal NTK also peaks earlier, consistent with the faster approach to the test loss minimum.}
\label{fig:NTKevol}
\end{figure}

Using the same network architectures and training algorithms described above, we also measured components of the NTK relevant for generalization and feature learning over the course of training. To illustrate the typical behavior, we trained single instantiations of width-30 networks with hidden depths $L -1 = 5$ and $L - 1 = 30$, for both choices of initialization. First, we measured the component of the output NTK $\widehat{H}_{ii; \alpha \beta}^{(L)}$ which is diagonal in neural indices but off-diagonal in sample indices, by taking one sample index $\alpha$ to be a random member of class ``6'' from the MNIST test set, and averaging the second sample index $\beta$ over all members of class ``6'' in the training set.\footnote{There is nothing special about the number 6, but at least from the perspective of topology, the handwritten representation of the digit 6 has more structure than the more obvious choices  of 0 or 1.} This observable is related to generalization because it updates how training examples in a given class affect the prediction of a test example in the same class. In Fig.~\ref{fig:NTKevol}, we plot the evolution of the principal diagonal component in neural indices, namely $\widehat{H}^{(L)}_{ii; \alpha \beta}$ with $i = 7$, since this is the output node which labels class ``6'' (in the standard MNIST convention with $i = 1$ labeling class ``0'') and thus should be most strongly correlated with the chosen data points. To elucidate the relationship between the NTK evolution and feature learning, we also plot the test and training loss curves for the same architectures.\footnote{In this experiment, in order to directly compare with the analogous plots in Sec.~\ref{sec:TrainingResults} and because we are not invoking early stopping, we use the validation set of size $10^4$ as our test set.}

\begin{figure}[t!]
\centering
\includegraphics[width=0.45\textwidth]{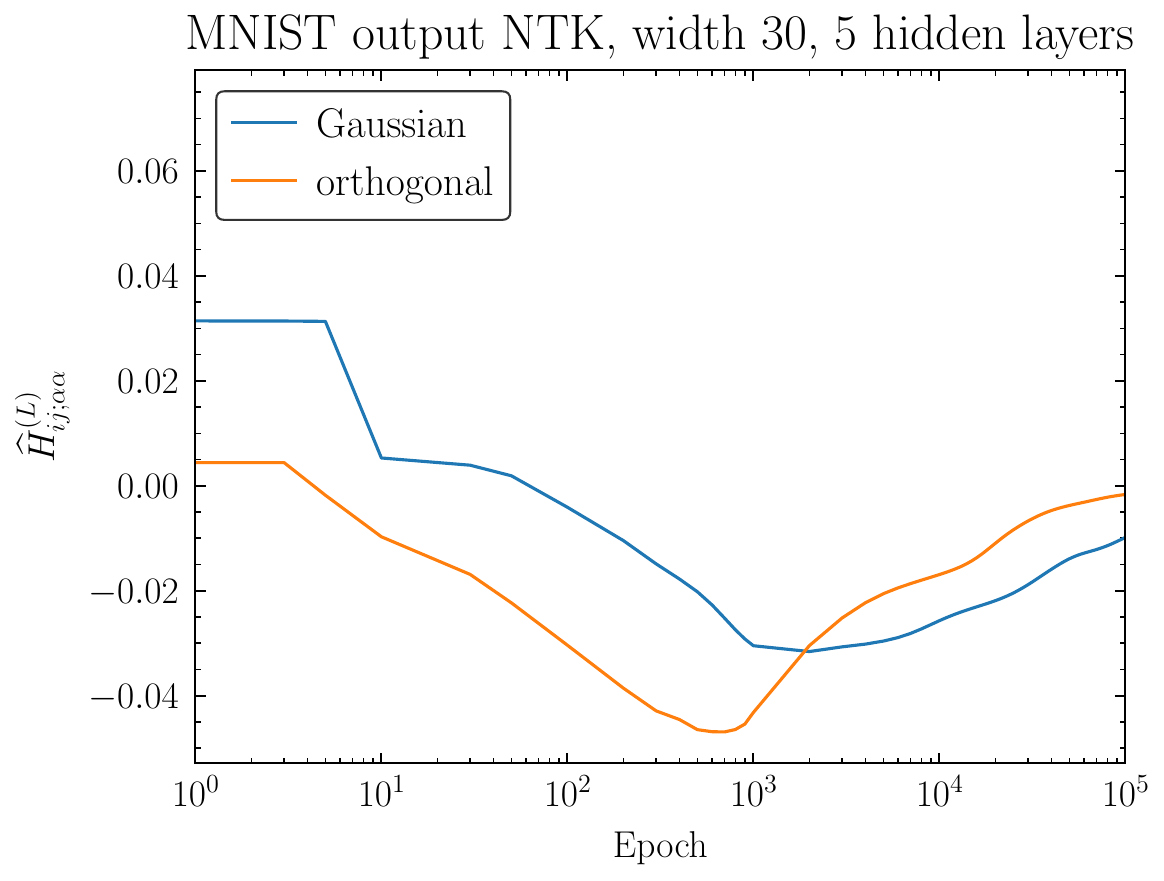}
\includegraphics[width=0.45\textwidth]{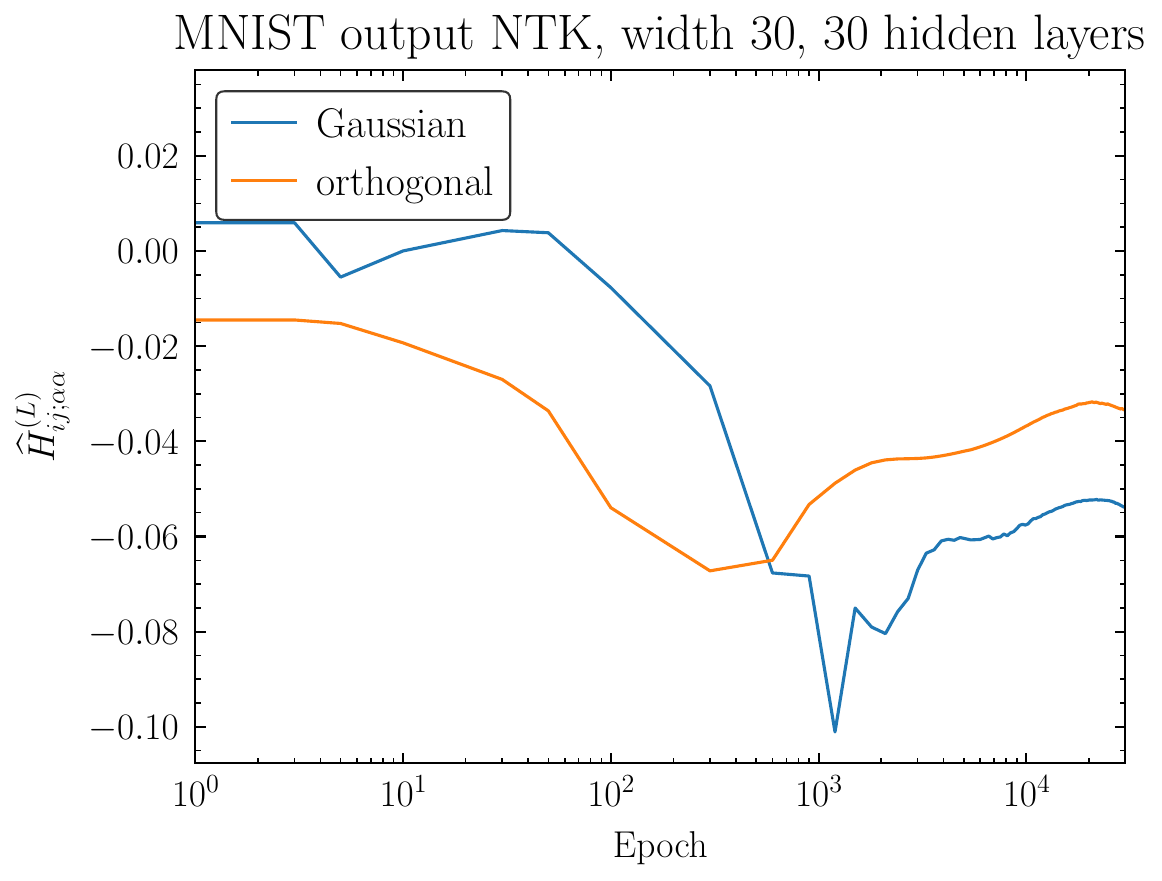}
\caption{Evolution of the single-input training NTK at the output layer, averaged over off-diagonal pairs of neural indices, for hidden depths $L-1 = 5$ (left) and $L-1 = 30$ (right). As in Fig.~\ref{fig:NTKevol}, the orthogonal NTK is consistent across depths, while the Gaussian NTK exhibits fluctuations in the $L/n \simeq 1$ regime.}
\label{fig:NTKevolSingleSample}
\end{figure}

Several features are immediately apparent in Fig.~\ref{fig:NTKevol}. First, for shallow networks with $L/n \ll 1$ (top row), the loss curves for Gaussian and orthogonal initializations are essentially identical, as are the shapes of the NTK curves. The NTK does evolve over training at the $\simeq 10\%$ level, consistent with the finite-width expectation, but the magnitude of the variation is larger for Gaussian than for orthogonal, due to the $L/n$ growth of the dNTK correlators which drive the NTK evolution. The NTK peaks roughly around the epoch where the loss curve is steepest, and peaks slightly earlier for this instantiation of the Gaussian network than for the orthogonal network; the relative speed of training is also visible in the loss curves. This similarity between the two initializations stands in sharp contrast to the case of $L/n \simeq 1$ (bottom row). There, the orthogonal network trains much faster (reflected in an NTK which peaks earlier) and exhibits markedly smaller variation over training than the Gaussian NTK, which has large oscillations around the epoch of fastest learning, followed by continued small oscillations as overfitting sets in, which are completely absent in the orthogonal case. Furthermore, the overall magnitude of the orthogonal NTK in the deep network is within $\sim 10\%$ of the analogous values for the shallow network, while the Gaussian NTK varies by factors of $\sim 2$. All of these observations support our conjecture that large NTK fluctuations inhibit both training and generalization, leading to superior performances for orthogonal networks compared to Gaussian networks with the same architecture.

Finally, to illustrate feature learning in the same models, we measure the output NTK which is diagonal in sample indices but off-diagonal in neural indices, $\widehat{H}_{ij; \alpha \alpha}^{(L)}$. Specifically, we take the sample index to be a random member of class ``6'' from the training set, and average over all off-diagonal pairs of neural indices. The results are shown in Fig.~\ref{fig:NTKevolSingleSample}. The overall size of the NTK is smaller than the diagonal component in Fig.~\ref{fig:NTKevol}, because the infinite-width NTK $\Theta_{\alpha \alpha}$ is purely diagonal in neural indices, so all deviations from zero are finite-width effects indicative of feature learning. Analogous to Fig.~\ref{fig:NTKevol}, the Gaussian and orthogonal NTKs behave similarly for $L/n \ll 1$. For $L/n \simeq 1$, the Gaussian NTK exhibits fluctuations around the epoch the loss curve is steepest, while the orthogonal NTK is very similar to its behavior for small depths. 

\section{Conclusion and outlook}
\label{sec:conclusion}

In this paper, we used the formalism of neural network correlation functions to help establish a firmer link between previous studies on the advantages of orthogonal initializations, and the important practical phenomena of feature learning and generalization. We identified the empirical depth-independence of the normalized NTK, dNTK, and ddNTK correlators -- and the corresponding suppression of $L/n$ fluctuations -- as possible explanations for the faster training and superior generalization of orthogonal networks with tanh activations, and corroborated those explanations with measurements of the NTK during training. In these examples, the advantages of orthogonal initializations are small (improvements in test loss at the percent level, and improvements in training speed by an order-one factor in epochs), but very consistent across a wide variety of measurements. To our knowledge, there has been relatively little follow-up work in this direction since Ref.~\cite{huang2021neural}, perhaps because the theoretical connection to feature learning away from the infinite-width limit was lacking. We hope this work rekindles some interest in the benefits of alternative initialization schemes to improve neural network performance.

There are numerous aspects of our analysis which we intend to pursue in future work. Most importantly, we plan to derive the full expressions for the NTK and dNTK recursions, analogous to the Gaussian versions in Ref.~\cite{Roberts:2021fes}, in order to demonstrate the depth independence of the correlators analytically. Such a derivation might also be able to shed light on the origin of the depth saturation scale $L \sim 20$: for instance, the components of the architecture or data which determines this number. Our numerical results seem to suggest that orthogonal networks have no $L/n$ cutoff; while we worked to leading order in $1/n$ in this paper, it may be possible to prove or at least motivate depth-independence to all orders in $1/n$, as we calculated for deep linear networks in Sec.~\ref{sec:linear}. Furthermore, since the same NTK correlators we studied control the variance of the trained network predictions, we intend to study whether orthogonal networks exhibit reduced variance compared to Gaussian networks, which could be of use in applications in physics where uncertainty quantification is paramount~\cite{Chen:2022pzc}. It would also be interesting to see if a theoretically-optimal global learning rate for MLPs can be derived analytically from studying the NTK correlators, and perhaps relate this to the $\eta = \mathcal{O}(1/L)$ prescription and the training time $\tau = \mathcal{O}(\sqrt{L})$ under (non-tensorial) SGD from Ref.~\cite{pennington2017resurrecting}. Finally, we intend to generalize our work to architectures and optimization algorithms of more practical use in modern applications. In particular, in light of the results of Refs.~\cite{doshi2021critical,he2022autoinit}, studying MLPs with \texttt{LayerNorm} and residual connections would be the next step in understanding the effect of orthogonal initializations on transformer architectures.

\vspace{1cm}

\noindent \textbf{Acknolwedgments.} We thank Yasaman Bahri, Nathaniel Craig, Jeffrey Pennington, Sho Yaida, and Kevin Zhang for stimulating conversations. This material is based upon work supported by the U.S. Department of Energy, Office of Science, Office of High Energy Physics, under Award Number DE-SC0023704. This work utilizes the computing resources of the HAL cluster~\cite{10.1145/3311790.3396649} supported by the National Science Foundation’s Major Research Instrumentation program, grant \#1725729, as well as the University of Illinois Urbana-Champaign. YK and DR thank the Aspen Center for Physics, which is supported by National Science Foundation grant PHY-2210452, for hospitality during the completion of this work. DR acknowledges support from the National Science Foundation under Cooperative Agreement PHY-2019786 (the NSF AI Institute for Artificial Intelligence and Fundamental Interactions, \url{http://
iaifi.org/}) and appreciates the sanction of Sequoia Capital.

\begin{appendix}

\section{Measurements of multi-input correlators}
\label{app:Multi}

Since the behavior of practical neural networks during training relies on correlations between multiple inputs from the training set, we measured the multi-input 4-point vertex $V$ and the multi-input NTK correlators $A$, $B$, $D$, and $F$ on random input data to verify that they exhibit the same behavior with depth as the single-input correlators studied in the main text. In particular, these NTK correlators are leading order in the global learning rate $\eta$ and govern the leading finite-width corrections to the mean trained network predictions~\cite{Roberts:2021fes}.

\subsection{Measurements of $V$}
\label{app:MultiV}

\begin{figure}[t!]
\centering
\textbf{Linear Activation}\par
\includegraphics[width=0.32\textwidth]{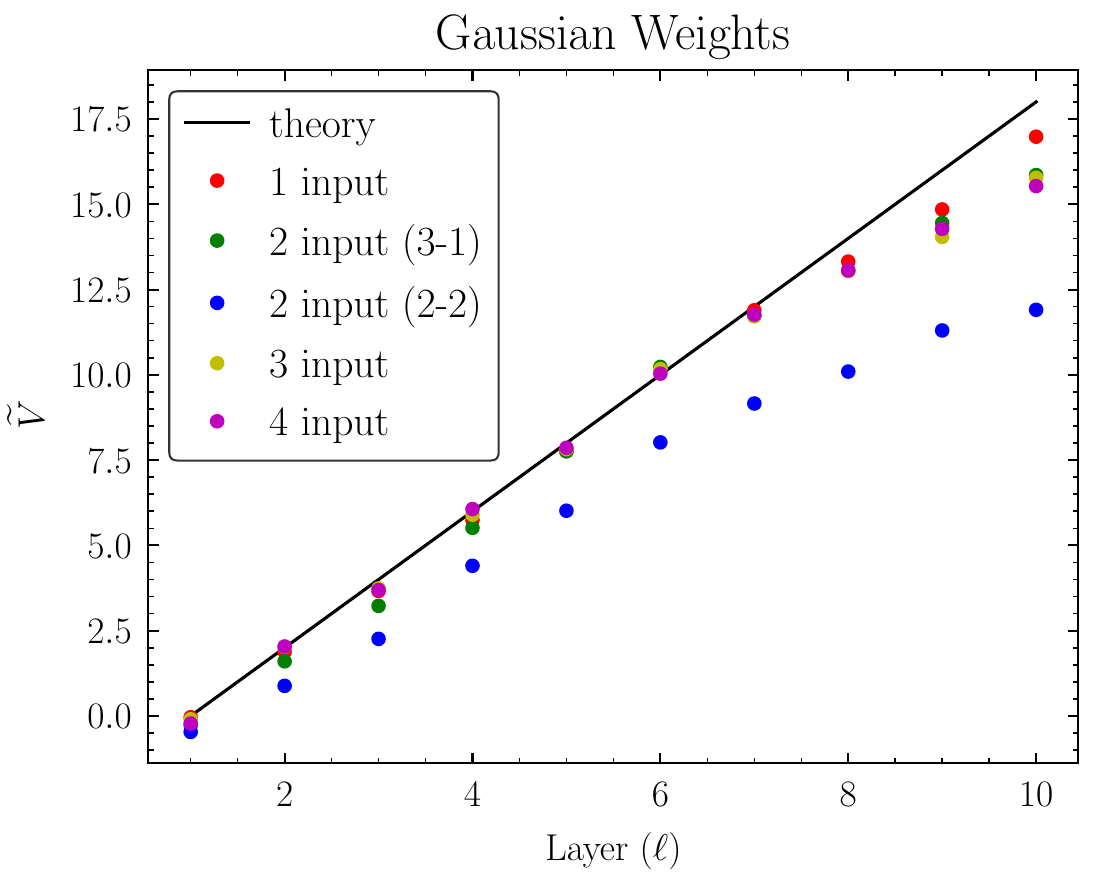}
\includegraphics[width=0.32\textwidth]{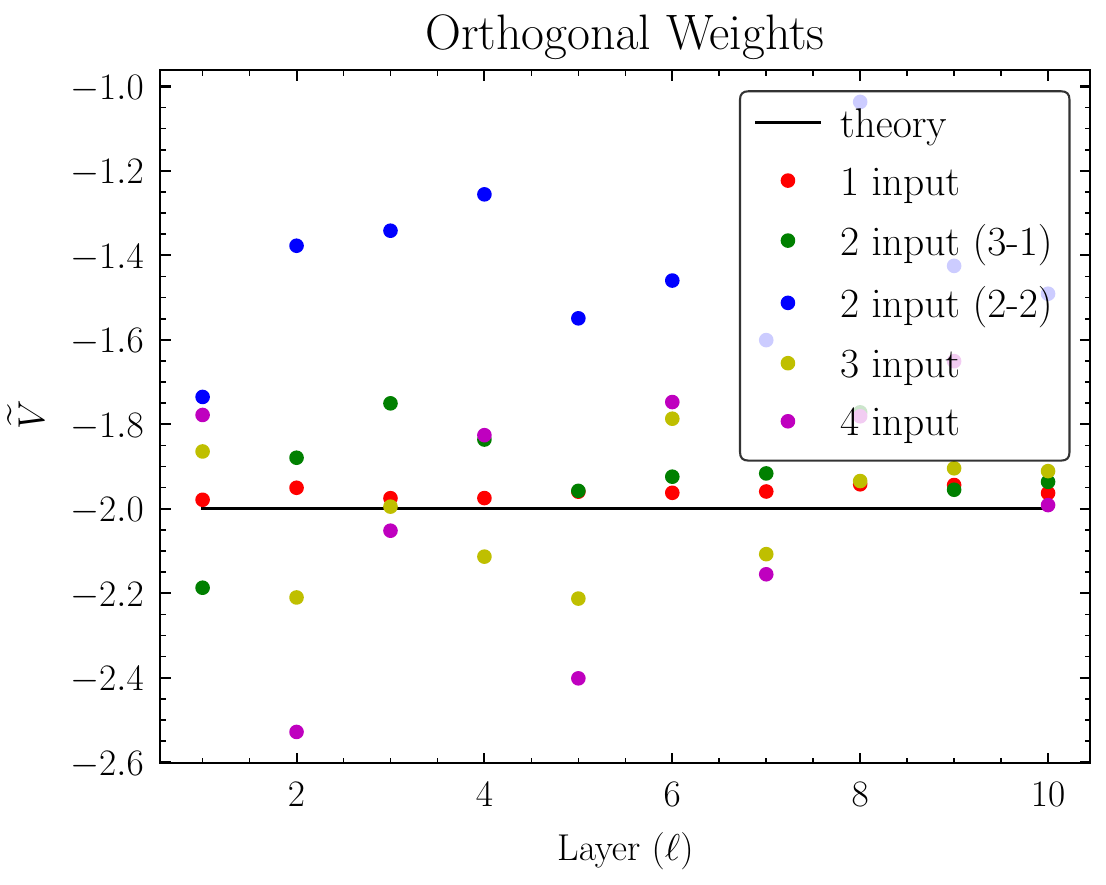}
\includegraphics[width=0.32\textwidth]{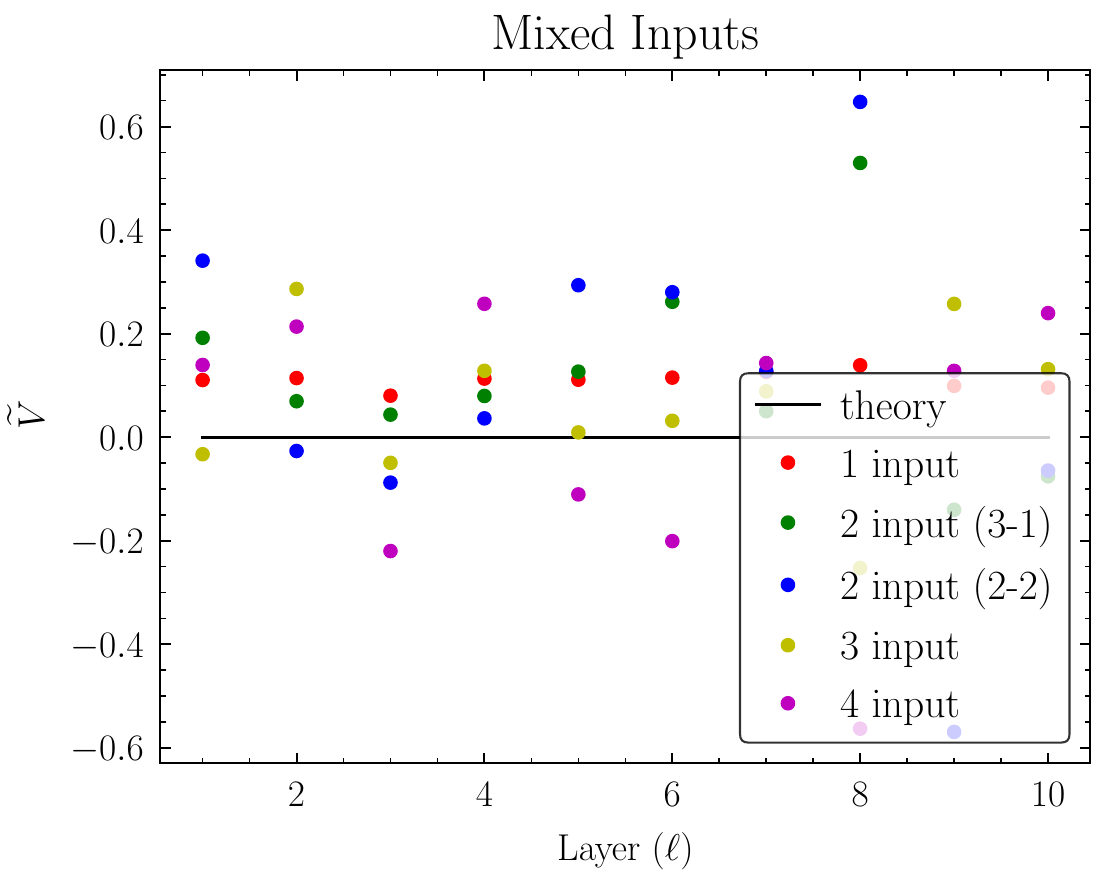}
\caption{Normalized 4-point correlator $\widetilde{V}$ for Gaussian (left), orthogonal (middle), and mixed (right) initializations, with $n = 100$ and linear activations applied to random data. Gaussian initializations grow with depth, while orthogonal and mixed initializations are approximately constant.}
\label{fig:linearVcorrelator}
\end{figure}

\begin{figure}[t!]
\centering
\textbf{ReLU Activation}\par
\includegraphics[width=0.32\textwidth]{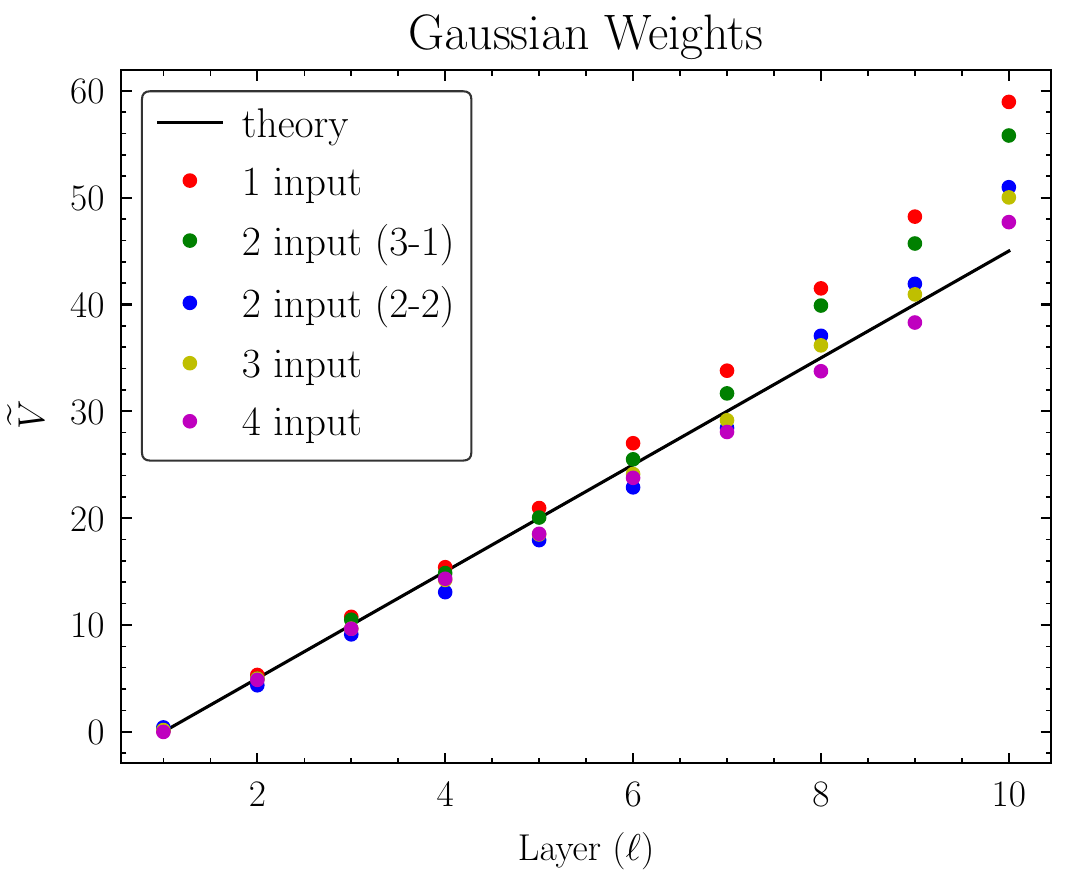}
\includegraphics[width=0.32\textwidth]{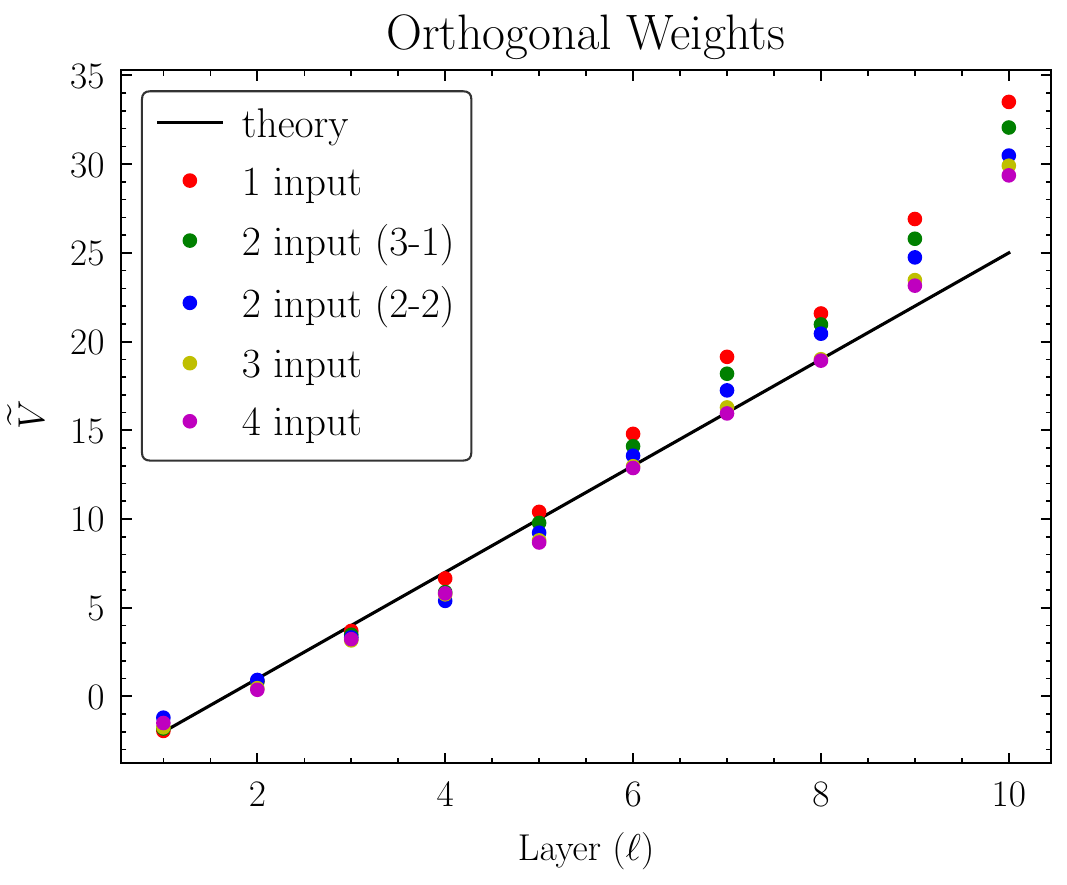}
\includegraphics[width=0.32\textwidth]{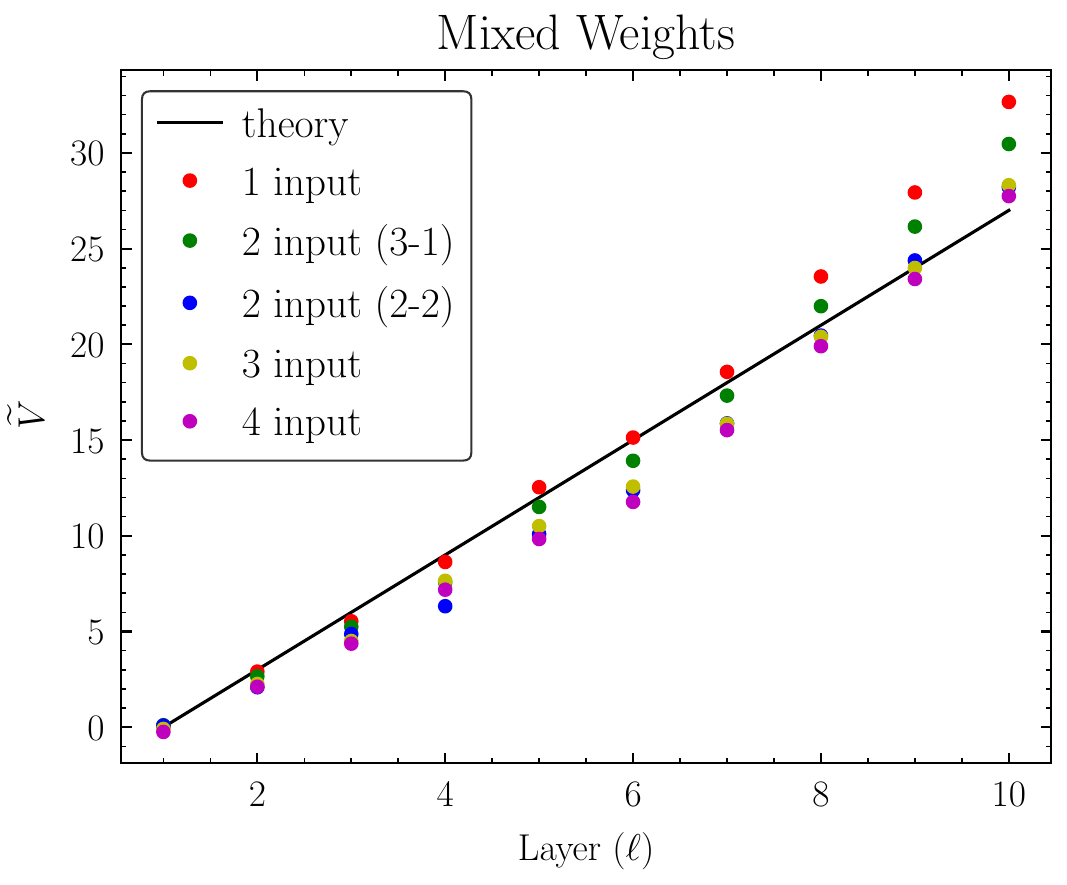}
\caption{Normalized 4-point correlator $\widetilde{V}$ for Gaussian (left), orthogonal (middle), and mixed (right) initializations, with $n = 100$ and ReLU activations applied to random data. All initializations grow with depth.}
\label{fig:ReLUVcorrelator}
\end{figure}

We initialized an ensemble of 1000 rectangular networks of width $n = 100$ and depth $L = 10$ with linear, ReLU, and tanh activations, and measured $V$ for both Gaussian and orthogonal initializations, as well as a mixed initialization with a Gaussian first layer and all other layers orthogonal. The input was taken to be a random vector of length $n$ with entries drawn uniformly from [0, 1], and we used the same input for each network configuration. We studied correlations between one, two, three, and four inputs; since $V$ accepts four inputs as arguments, for the case of two distinct inputs, we considered both the case of three identical arguments and one different (3-1) and the case of two pairs of identical arguments (2-2). For two distinct arguments with sample indices $\alpha$ and $\beta$ we measured
\begin{equation}
\label{eq:V2input}
    \widetilde{V}_{(\alpha \alpha)(\beta \beta)} \equiv \frac{V_{(\alpha \alpha)(\beta \beta)}}{\frac{1}{2}\left(K_{\alpha \alpha}K_{\beta \beta} + (K_{\alpha \beta})^2\right)}, \qquad \widetilde{V}_{(\alpha \alpha)(\alpha \beta)} \equiv \frac{V_{(\alpha \alpha)(\alpha \beta)}}{K_{\alpha \alpha} K_{\alpha \beta}}.
\end{equation}
For three distinct arguments $\alpha, \beta, \gamma$, we measured
\begin{equation}
\label{eq:V3input}
    \widetilde{V}_{(\alpha \beta)(\gamma \alpha)} \equiv \frac{V_{(\alpha \beta)(\gamma \alpha)}}{\frac{1}{2}\left(K_{\alpha \beta}K_{\alpha \gamma} + K_{\alpha \alpha}K_{\beta \gamma}\right)},
\end{equation}
and for four distinct arguments $\alpha, \beta, \gamma, \delta$, we measured
\begin{equation}
\label{eq:V4input}
    \widetilde{V}_{(\alpha \beta)(\gamma \delta)} \equiv \frac{V_{(\alpha \beta)(\gamma \delta)}}{\frac{1}{3}\left(K_{\alpha \beta}K_{\gamma \delta} + K_{\alpha \gamma}K_{\beta \delta} + K_{\alpha \delta}K_{\beta \gamma} \right)}.
\end{equation}

\begin{figure}[t!]
\centering
\textbf{Tanh Activation}\par
\includegraphics[width=0.32\textwidth]{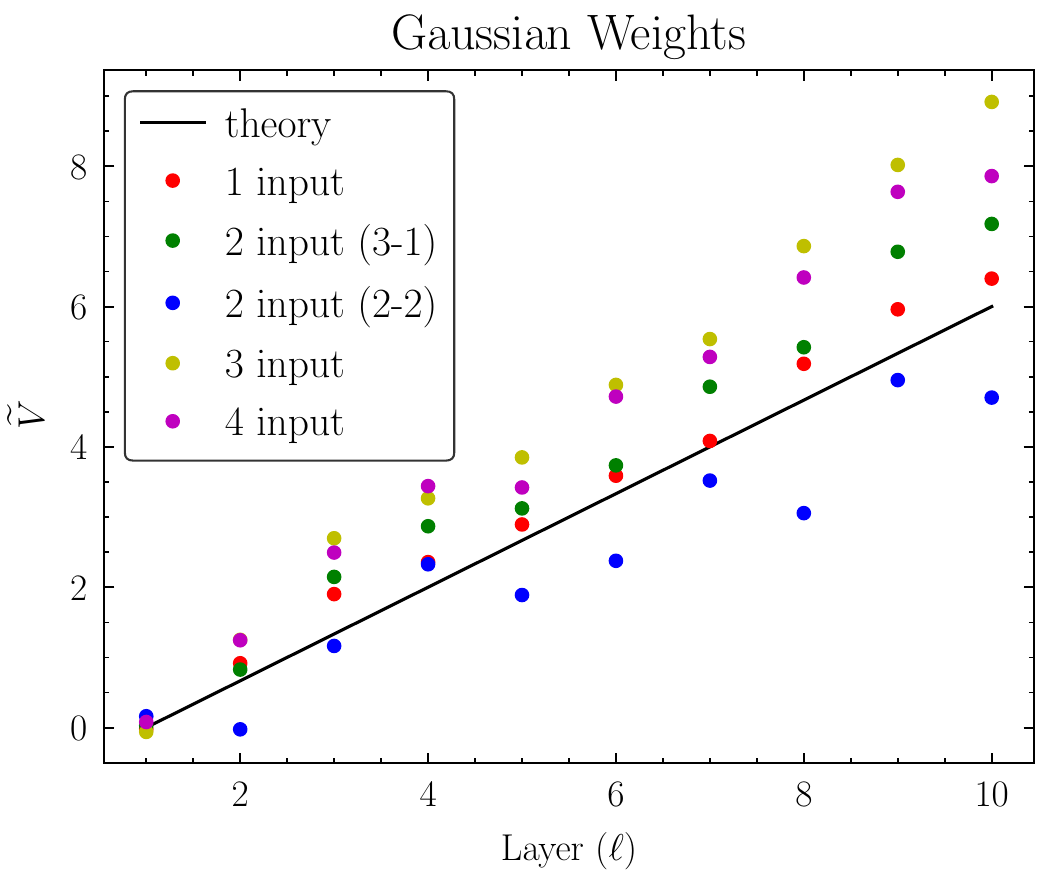}
\includegraphics[width=0.32\textwidth]{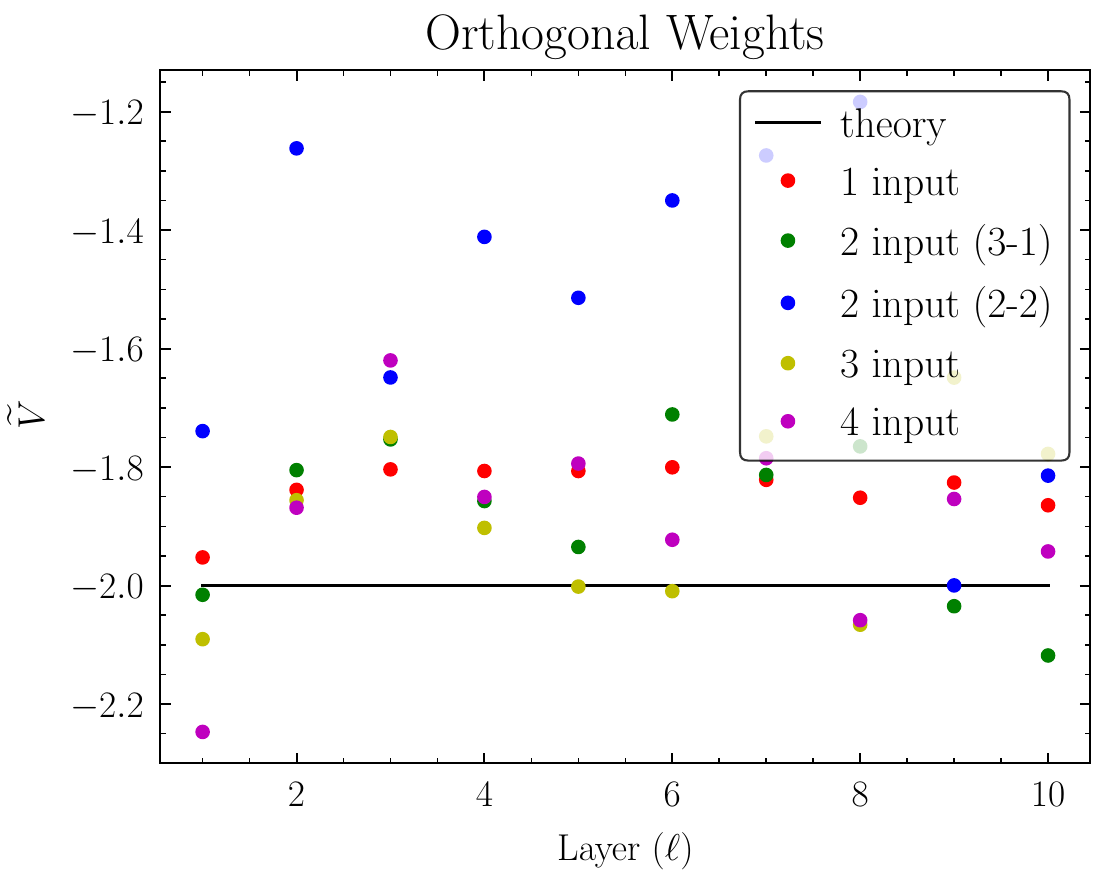}
\includegraphics[width=0.32\textwidth]{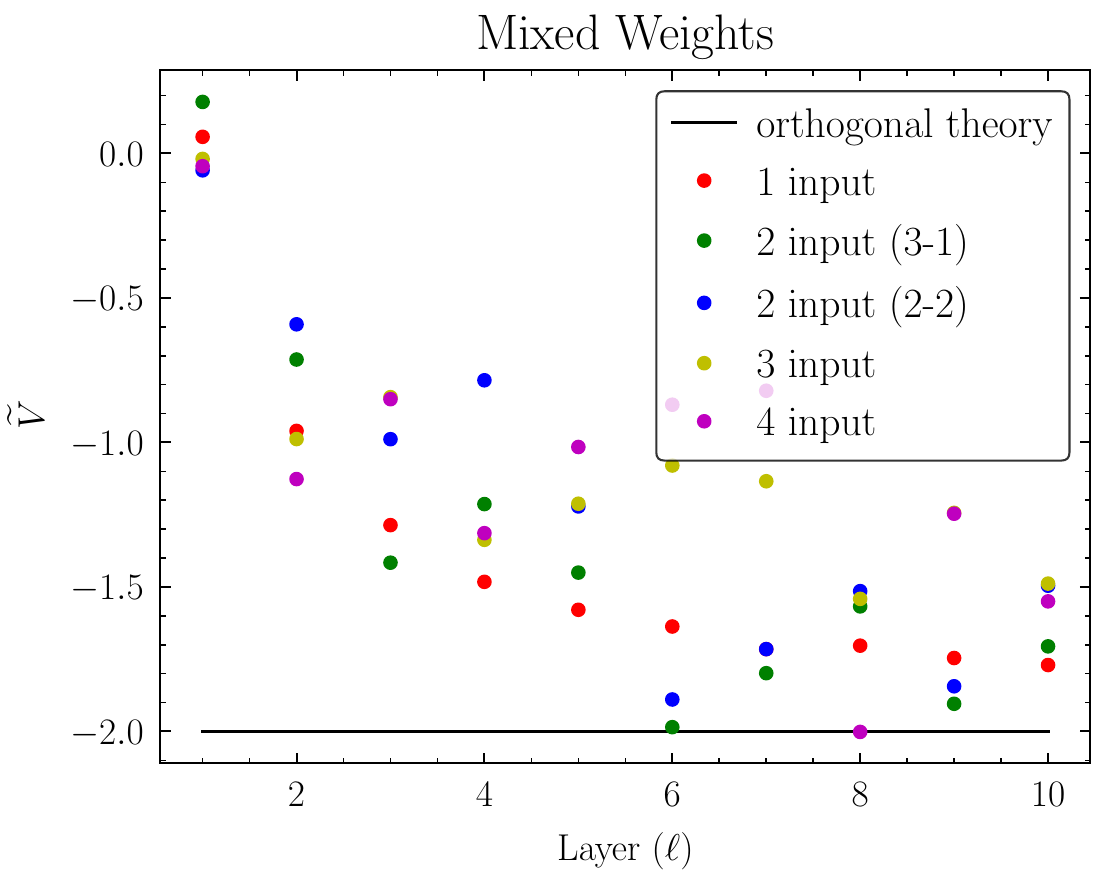}
\caption{Normalized 4-point correlator $\widetilde{V}$ for Gaussian (left), orthogonal (middle), and mixed (right) initializations, with $n = 100$ and tanh activations applied to random data. The behavior is similar to the linear activations in Fig.~\ref{fig:linearVcorrelator}; Gaussian initializations grow with depth, while orthogonal initializations are approximately constant and mixed initializations asymptote to a constant.}
\label{fig:tanhVcorrelator}
\end{figure}

The results are shown in Figs.~\ref{fig:linearVcorrelator}--\ref{fig:tanhVcorrelator}, where for reference, we also show the theoretical values for the single-input correlators in solid black. Consistent with the single-input behavior, the multi-input correlators grow approximately linearly with depth for Gaussian initializations with all activations. The orthogonal correlators also mirror the single-input behavior: they grow linearly for ReLU activations, are approximately constant for tanh and linear activations, and asymptote to a constant for mixed initializations. The larger scatter for the multi-input orthogonal correlators is likely due to our choice of normalization; the kernel $K_{\alpha \beta}$ for $\alpha \neq \beta$ is typically quite small since random high-dimensional vectors are approximately orthogonal, and a more detailed analysis using the decomposition of the multi-input correlator into irreducible representations of its symmetries would likely lead to more stable behavior.

\subsection{Measurements of NTK correlators}
\label{app:MultiNTK}

\begin{figure}[t!]
\centering
\includegraphics[width=0.53\textwidth]{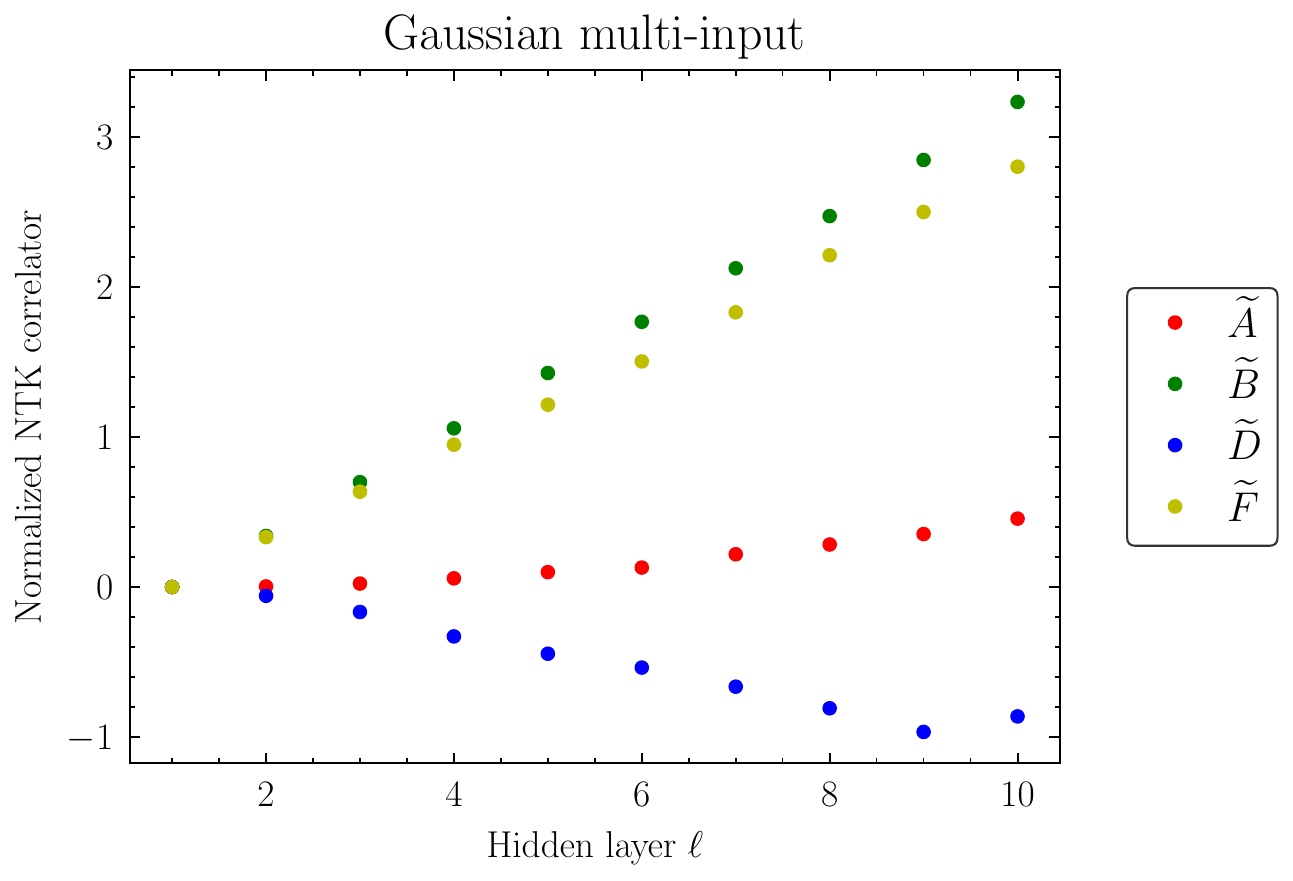}
\includegraphics[width=0.45\textwidth]{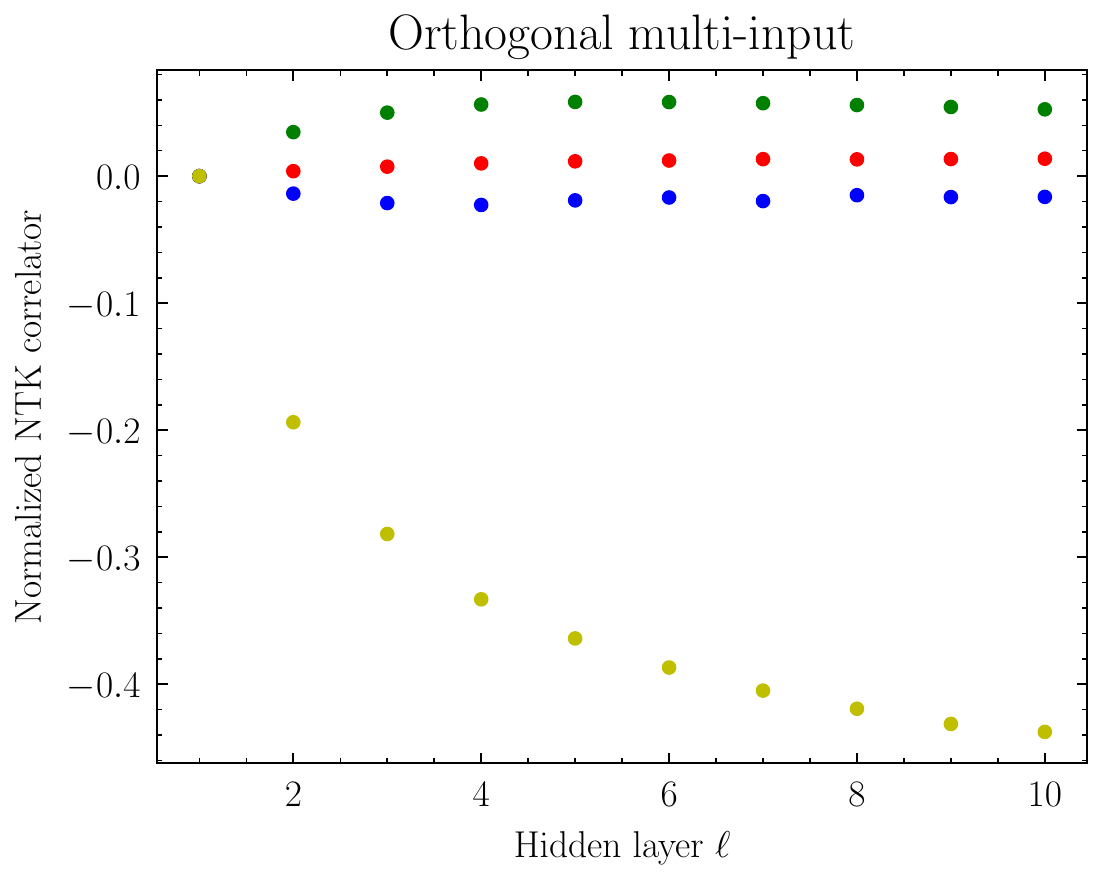}
\caption{Normalized four-input correlators for Gaussian (left) and orthogonal (right) initializations, with $n = 50$ and tanh activations, in an ensemble of 500 networks. The qualitative shapes of these curves match those of the single-input correlators.} 
\label{fig:TanhMultiABDF}
\end{figure}

We initialized an ensemble of 500 rectangular networks of width $n=50$ and depth $L=10$ with tanh activations, and measured the normalized correlators of Eq.~\ref{eq:ABdef} and Eq.~\ref{eq:DFdef} with four distinct random inputs $\alpha_1, \alpha_2, \alpha_3, \alpha_4$, with entries uniform on $[0,1]$. As with the 4-input $V$, we normalized the correlator by $\Theta^2$ or $\Theta K$ as appropriate with sample index structure analogous to Eq.~(\ref{eq:V4input}): 
\begin{equation}
\label{eq:AB4input}
    (\widetilde{A},\widetilde{B})_{\alpha \beta \gamma \delta} \equiv \frac{(A,B)_{\alpha \beta \gamma \delta}}{\frac{1}{3}\left(\Theta_{\alpha \beta}\Theta_{\gamma \delta} + \Theta_{\alpha \gamma}\Theta_{\beta \delta} + \Theta_{\alpha \delta}\Theta_{\beta \gamma} \right)}
\end{equation}
and
\begin{equation}
\label{eq:DF4input}
   ( \widetilde{D},\widetilde{F})_{\alpha \beta \gamma \delta} \equiv \frac{(D,F)_{\alpha \beta \gamma \delta}}{\frac{1}{6}\left(\Theta_{\alpha \beta}K_{\gamma \delta} + \Theta_{\alpha \gamma}K_{\beta \delta} + \Theta_{\alpha \delta}K_{\beta \gamma} + \Theta_{\gamma \delta}K_{\alpha \beta} + \Theta_{\beta \delta}K_{\alpha \gamma} + \Theta_{\beta \gamma}K_{\alpha \delta} \right)}.
\end{equation}
The results are shown in Fig.~\ref{fig:TanhMultiABDF} for Gaussian and orthogonal initializations. As was the case with $V$, we find that these multi-input NTK correlators at initialization have same qualitative behavior as the single-input correlators, with Gaussian initializations leading to linearly-growing normalized correlators, but orthogonal initializations saturating to a finite value. We intend to explore the behavior of these multi-input correlators through analytical calculations in future work.

\begin{figure}[t!]
\centering
\includegraphics[width=0.55\textwidth]{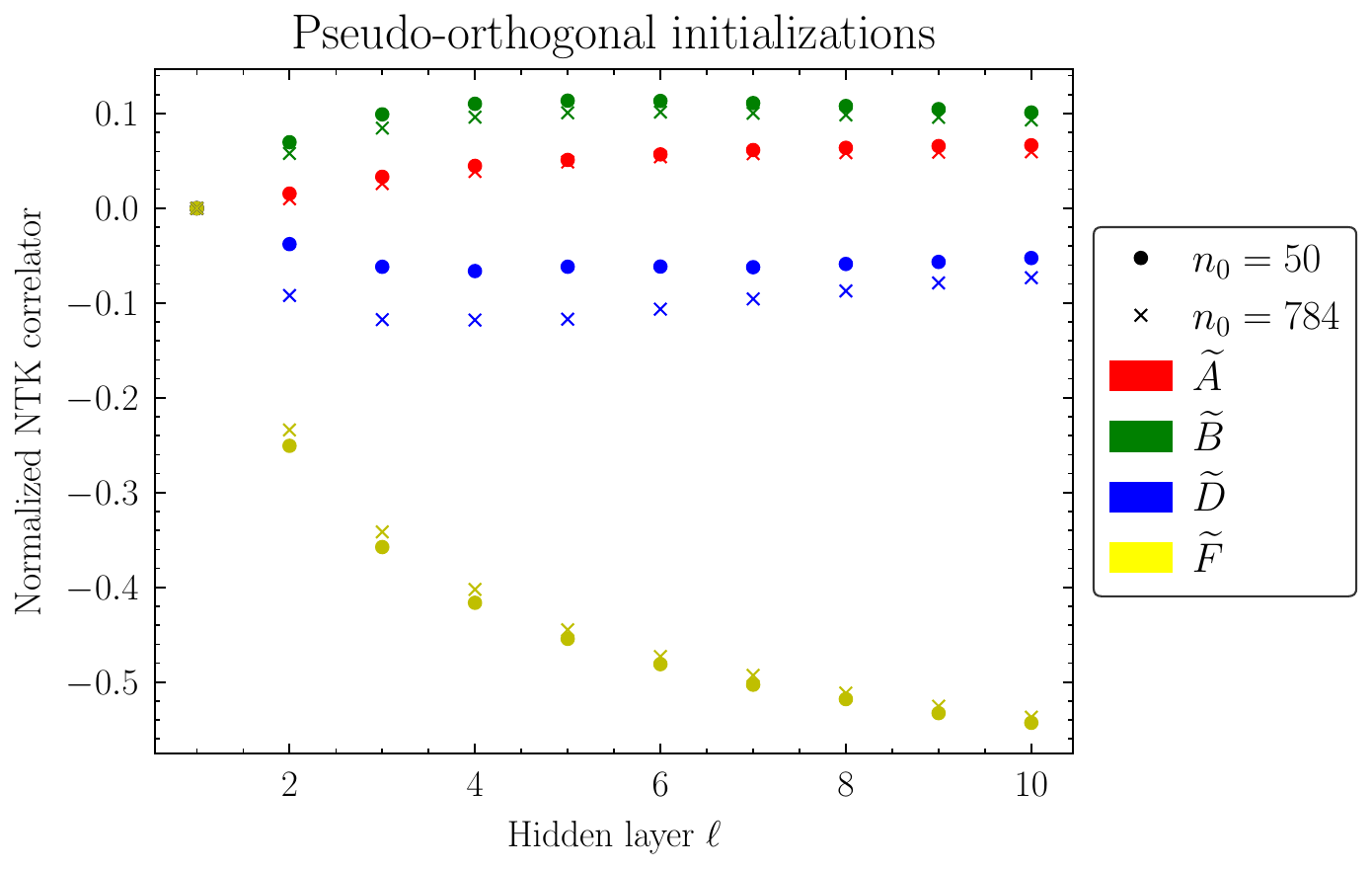}
\caption{Normalized $A$, $B$, $D$, and $F$ correlators with input width $n_0=50$ and $784$ and hidden width $n = 50$, for (pseudo)-orthogonal initialization and tanh activations. The pseudo-orthogonal initialization ($n_0\neq n$) asymptotes to the orthogonal initialization ($n_0=n$) as depth increases.} 
\label{fig:OrthoMixABDF}
\end{figure}

Finally, we verified that the asymptotic behavior of the NTK correlators is independent of the particular choice of mixed initialization in the first layer. Specifically, the QR decomposition used in our training experiments does not lead to a rectangular matrix with i.i.d.\ Gaussian entries, because the rows or columns are orthogonal. Fig.~\ref{fig:OrthoMixABDF} compares the correlators in a rectangular network with $n = 50$ to those from random inputs with size 784 (analogous to the MNIST dataset), with first-layer weights in both cases given by the QR algorithm. The differences are largest in the first layer, as expected, but both initializations asymptote to the same values at large depth.

\section{Variations on training experiments}
\label{app:VarTraining}

In this Appendix we make various perturbations to our choices of hyperparameters to confirm that our conclusions about the training and generalization advantages of orthogonal networks are robust. All networks are trained on MNIST with hidden width 30.

\begin{figure}[t!]
\centering
$\mathbf{C_W=0.1}$\par
\includegraphics[width=0.45\textwidth]{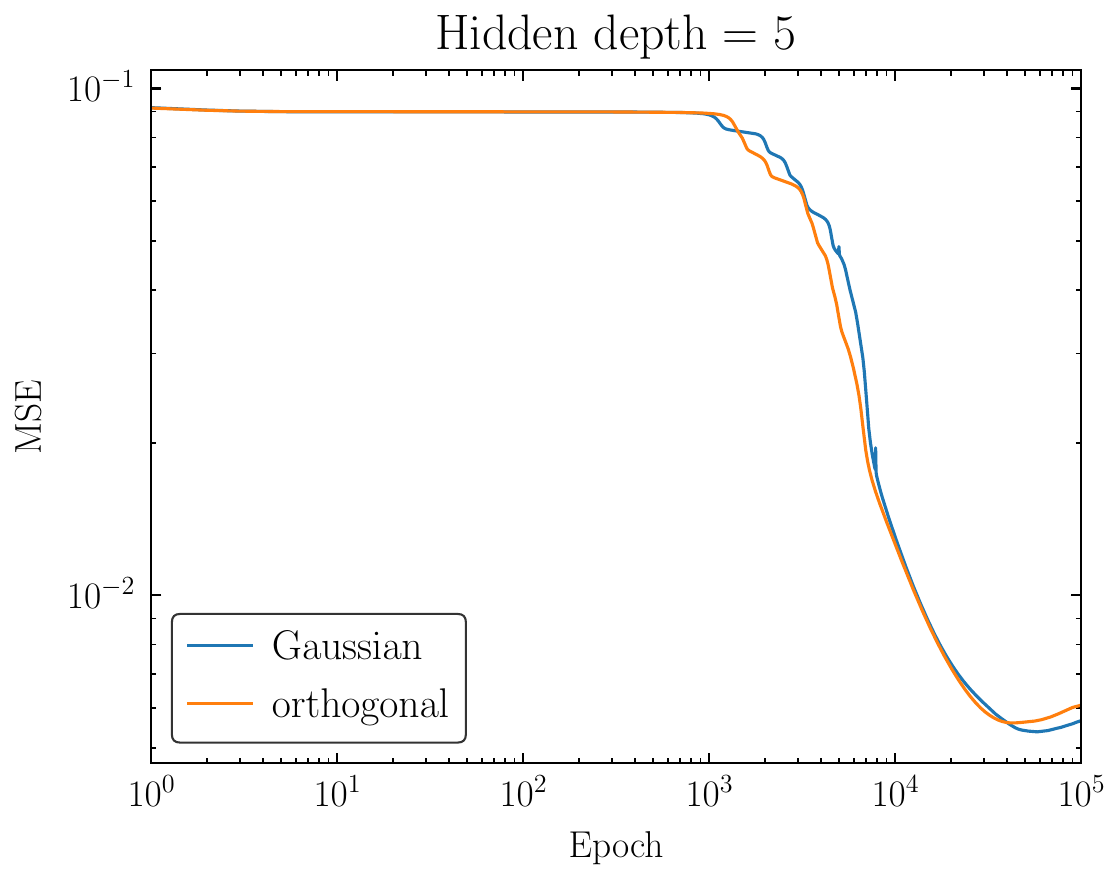}
\includegraphics[width=0.485\textwidth]{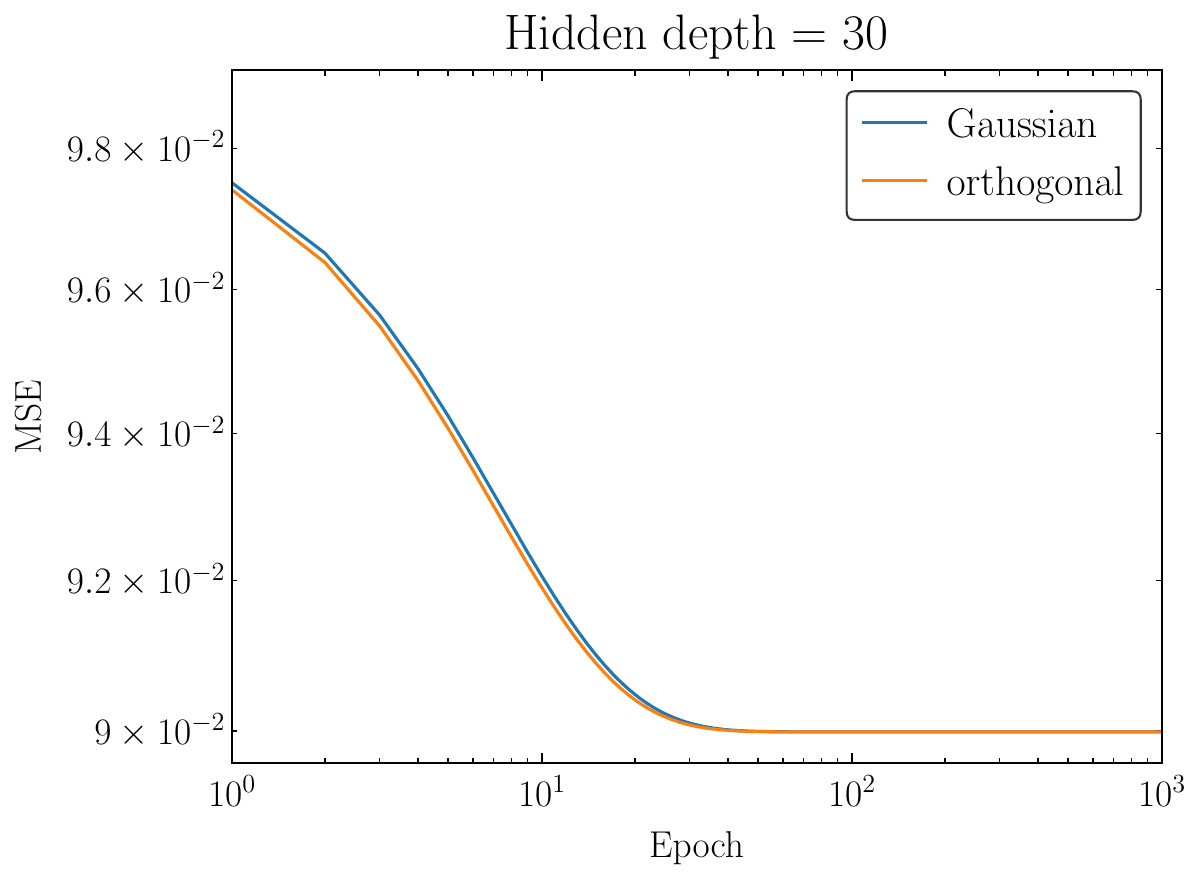}
\caption{MSE validation loss versus epoch for networks of width 30 and hidden depth 5 (left) and 30 (right), with tanh activation and $C_W=0.1$ on MNIST data. The behavior of both initializations is essentially identical for both depths. Note that the $y$-axis scale on the depth-30 plot is highly compressed; the deeper network barely trains over $10^5$ epochs due to exponentially vanishing gradients.}
\label{fig:lossesCW0p1}
\end{figure}

\begin{figure}[t!]
\centering
$\mathbf{C_W=10}$\par
\includegraphics[width=0.45\textwidth]{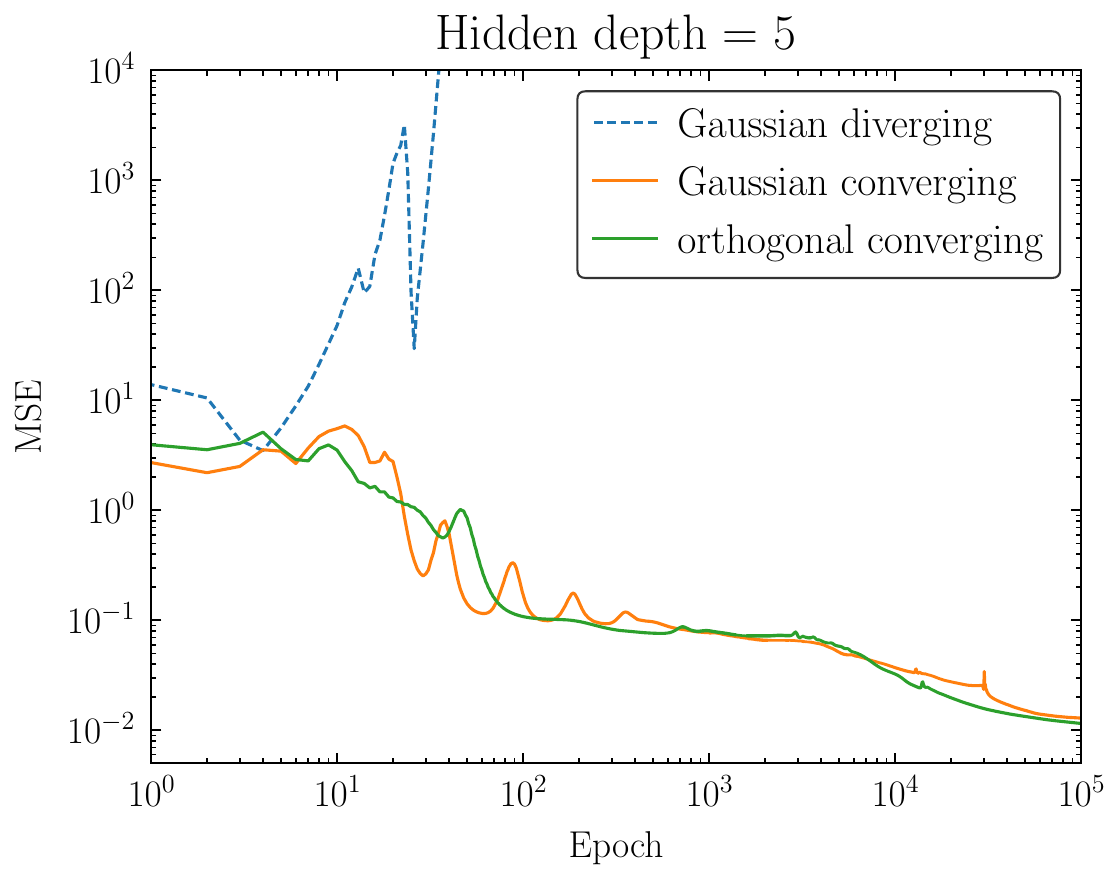}
\includegraphics[width=0.45\textwidth]{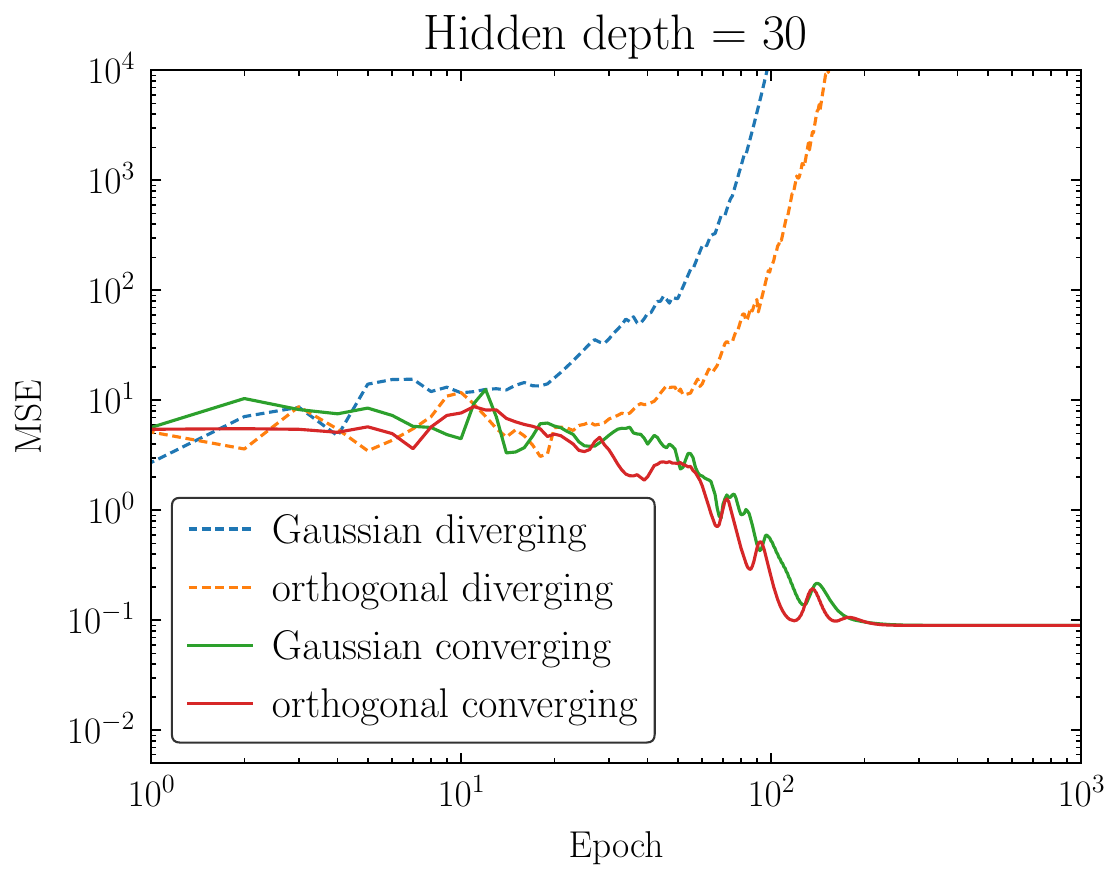}
\caption{MSE validation loss versus epoch for networks of width 30 and hidden depth 5 (left) and 30 (right), with tanh activation and $C_W=10$ on MNIST data. This configuration consistently results in a diverging loss, as shown by the dotted lines. For a hidden depth of 5, the Gaussian network had a diverging loss approximately every one in four runs, while the orthogonal network never diverged over the 20 runs we attempted. For a hidden depth of 30, both network initializations produced diverging losses approximately every other run.}
\label{fig:lossesCW10}
\end{figure}

\begin{figure}[t!]
\centering
\textbf{ReLU activation}\par
\includegraphics[width=0.45\textwidth]{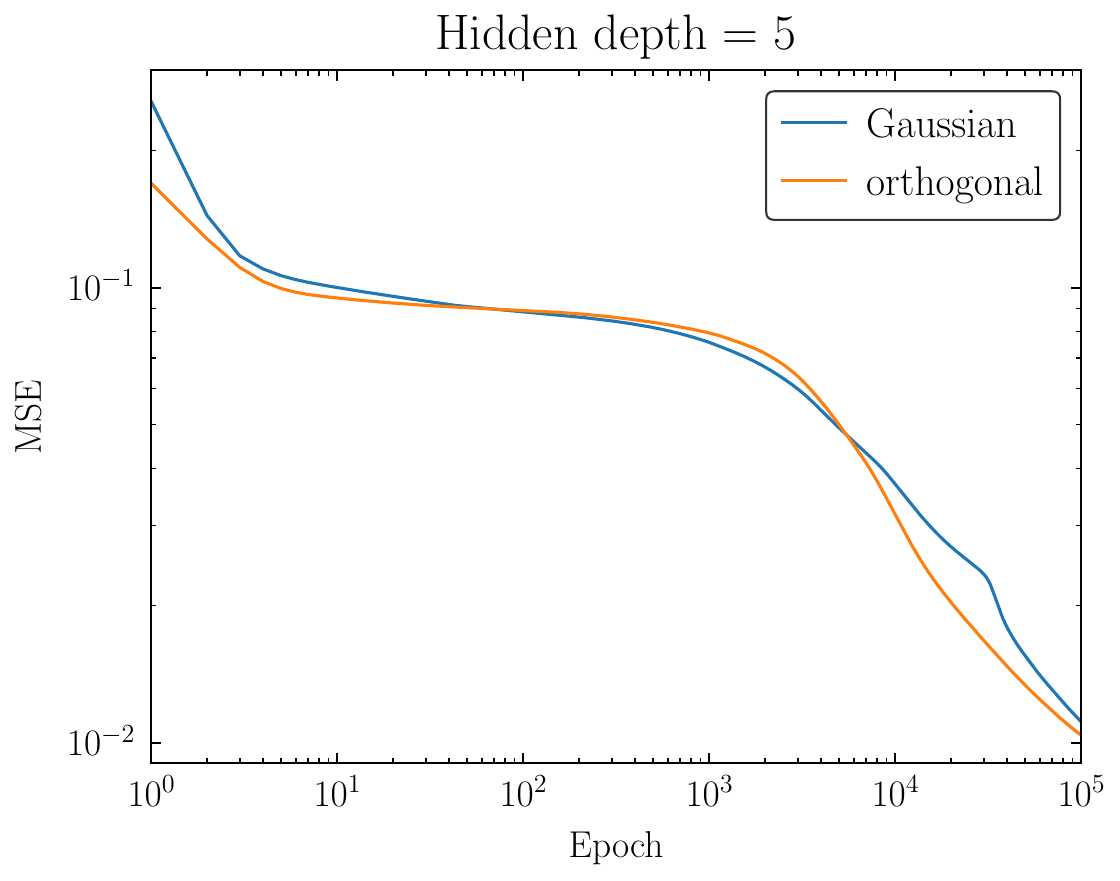}
\includegraphics[width=0.47\textwidth]{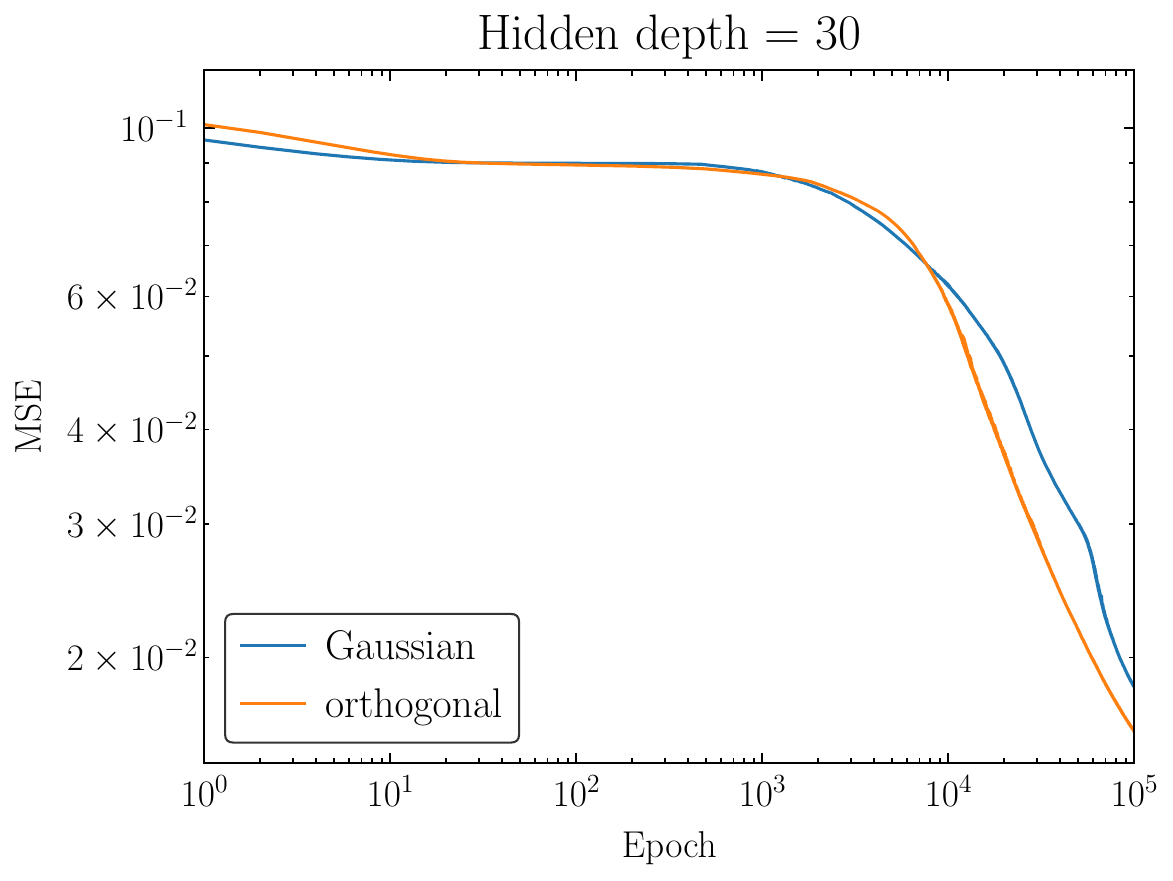}
\caption{MSE validation loss versus epoch for networks of width 30 and hidden depth 5 (left) and 30 (right), with ReLU activation and $C_W=2$ (the critical value) on MNIST data. The weight and bias learning rates are set equal to $1/L$ rather than what is used in the main text, Eq.~\ref{eq:LRrescaling}; this allows for a direct comparison between the results because both prescriptions preserve equal contributions to the NTK at each layer. The behavior of both initializations is very similar for both depths, with perhaps a slight increase in training speed from orthogonal initializations.}
\label{fig:lossesReLU}
\end{figure}

First, Figs.~\ref{fig:lossesCW0p1} and~\ref{fig:lossesCW10} show the effect of choosing $C_W$ smaller or larger than the critical value of $C_W = 1$. As expected by the analysis of Ref.~\cite{Roberts:2021fes}, $C_W < 1$ (Fig.~\ref{fig:lossesCW0p1}) leads to a vanishing gradient problem where networks either begin training much later (left, $L/n \ll 1$) or fail to train altogether (right, $L/n \simeq 1$). Similarly, $C_W > 1$ (Fig.~\ref{fig:lossesCW10}) leads to an exploding gradient problem where an order-one fraction of runs give a diverging loss. Crucially, orthogonal initializations suffer from exactly the same issues as Gaussian initializations, confirming that the benefits of orthogonal initializations are only seen at networks tuned to criticality. Similarly, Fig.~\ref{fig:lossesReLU} shows the results of changing the activation function to ReLU, now taking $C_W = 2$ to maintain criticality. In addition, we make the associated change in the weight and bias learning rates from Ref.~\cite{Roberts:2021fes} to preserve equal contributions to the NTK at each layer: specifically, the bias learning rates are now independent of depth, but both weight and bias learning rates have an overall scaling with the total depth $1/L$. Curiously, both shallow and deep networks take considerably longer to train than for tanh activations (overfitting has not occured by epoch $10^5$), but orthogonal initializations offer only a mild improvement over Gaussian, if any. This is consistent with the analysis in the main text, where orthogonal ReLU networks still have correlators that grow linearly with depth, though at a slightly reduced slope compared to Gaussian ReLU networks.

\begin{figure}[t!]
\centering
\textbf{Ordinary gradient descent}\par
\includegraphics[width=0.45\textwidth]{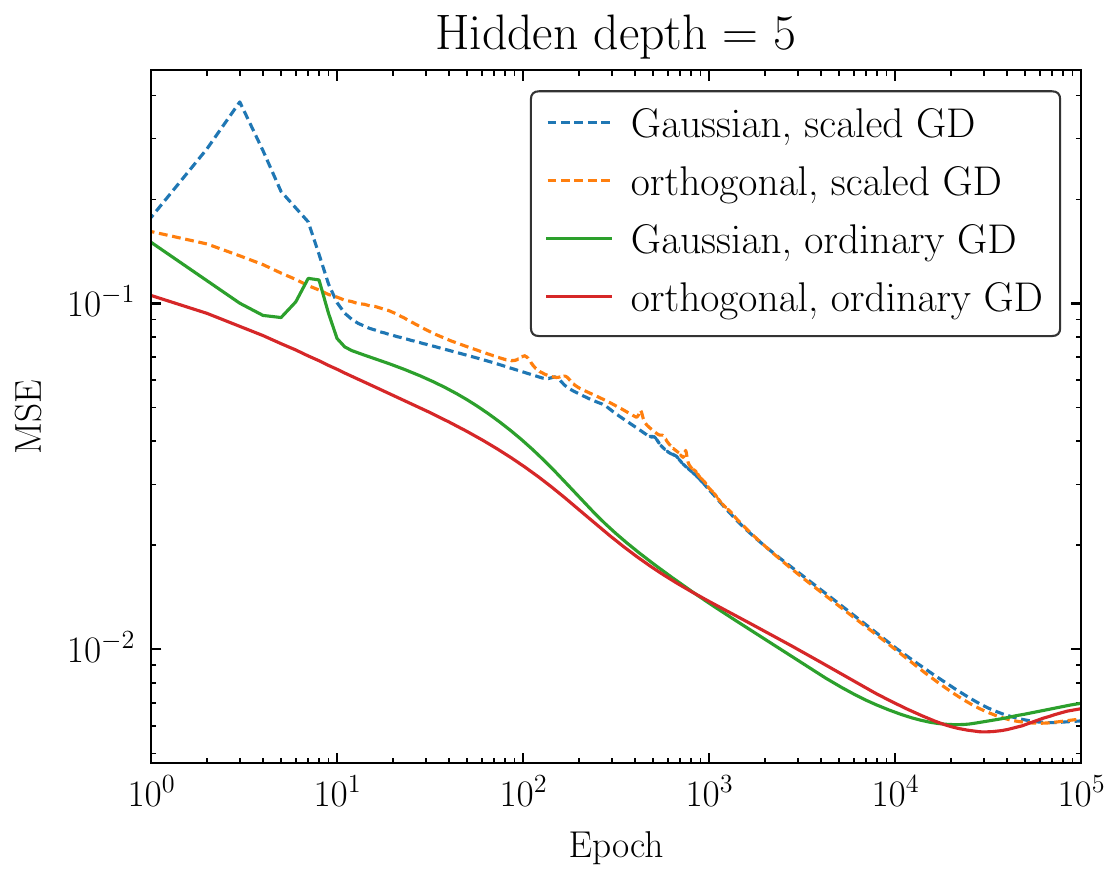}
\includegraphics[width=0.45\textwidth]{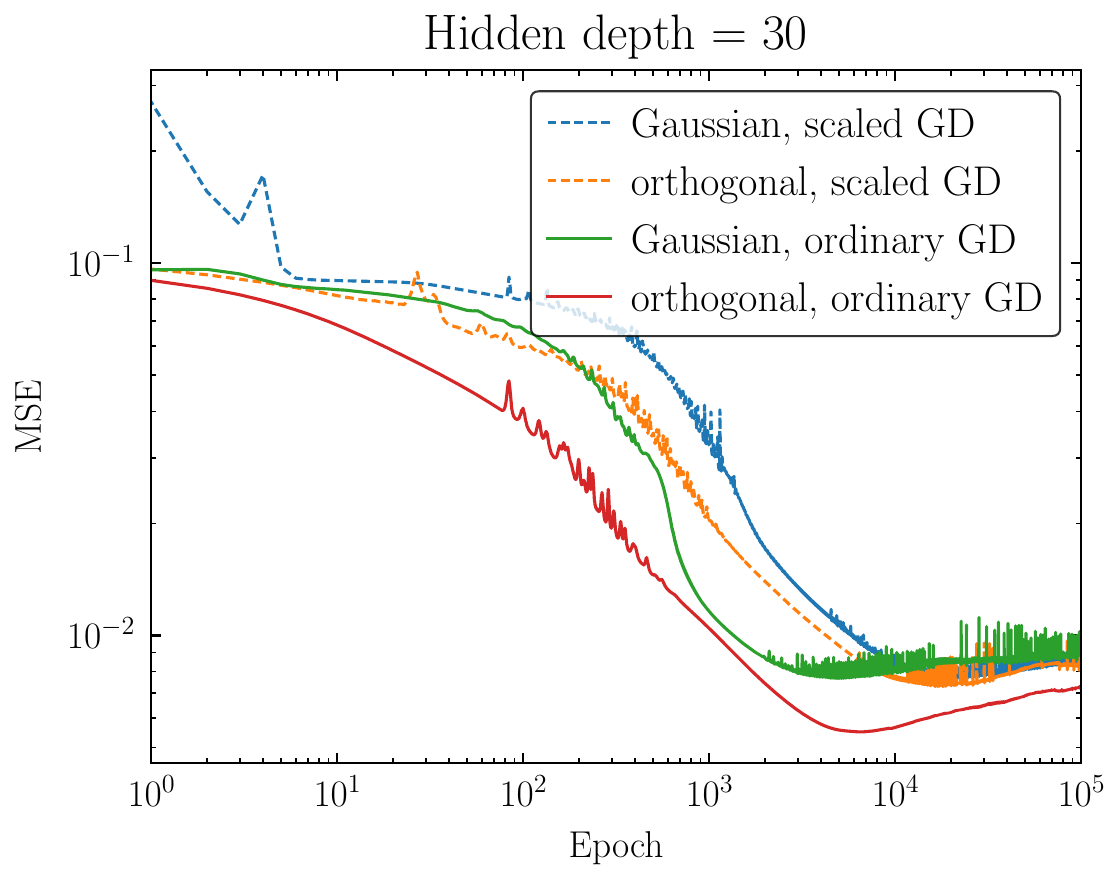}
\caption{MSE validation loss versus epoch for networks of width 30 and hidden depth 5 (left) and 30 (right), with tanh activations. We compare ordinary GD optimization (solid) and scaled GD (dashed) on MNIST data. Compared to what is used in the main text for scaled GD, the global learning rate $\eta$ for ordinary GD is divided by the width of the network; this allows for a direct comparison between the results because the scaled GD optimizer divides the weight learning rate by the width. Ordinary GD trains faster and achieves better generalization for deep orthogonal networks compared to scaled GD. The loss curves for ordinary GD are nearly identical for the two initializations with $L/n \ll 1$ but orthogonal outperforms Gaussian for $L/n \simeq 1$.}
\label{fig:lossesSGD}
\end{figure}

\begin{figure}[t!]
\centering
\textbf{Adam optimization}\par
\includegraphics[width=0.45\textwidth]{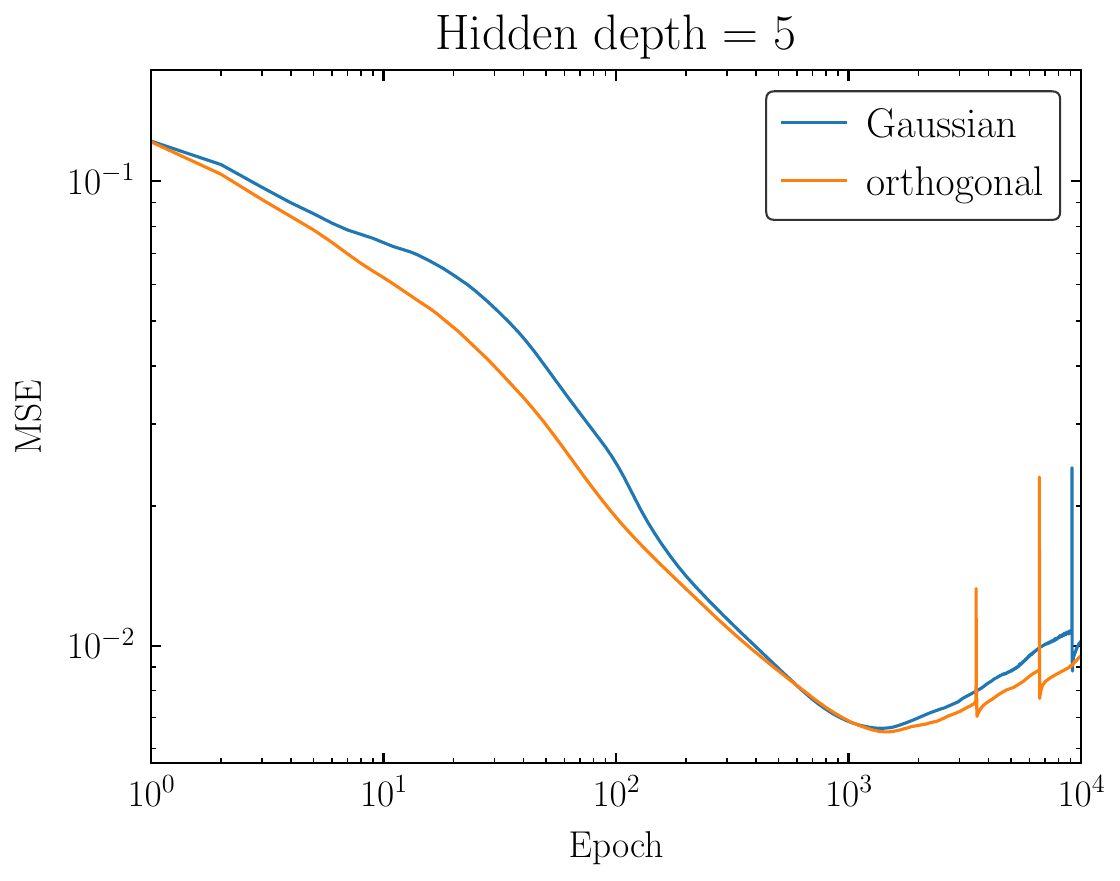}
\includegraphics[width=0.45\textwidth]{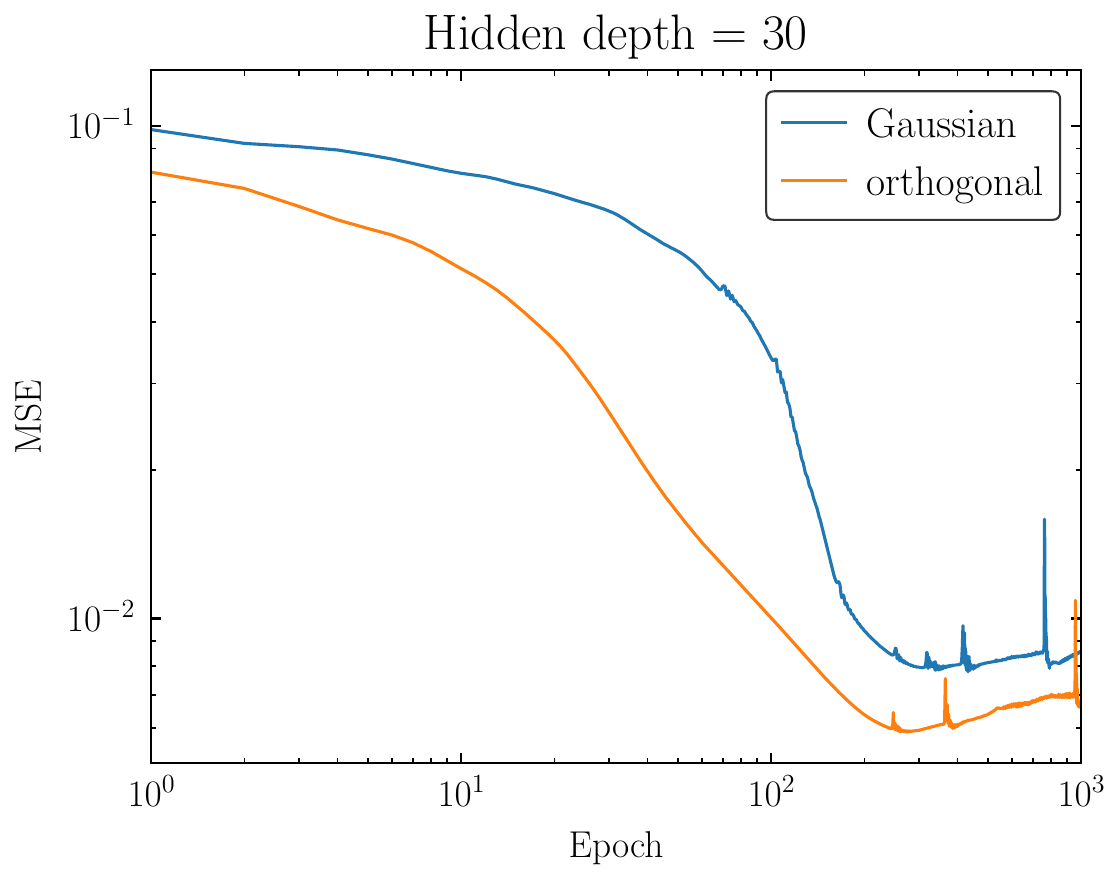}
\caption{MSE validation loss versus epoch for networks of width 30 and hidden depth 5 (left) and 30 (right), with tanh activation and \texttt{Adam} optimization on MNIST data. The global learning rate $\eta$ is set to $10^{-3}$ in order to achieve comparable best losses compared to those in the main text. Both initializations have nearly identical performance for depth 5, but orthogonal outperforms Gaussian in both training speed and generalization for depth 30.}
\label{fig:lossesAdam}
\end{figure}

Next, we consider varying the optimizer. Fig.~\ref{fig:lossesSGD} shows the result of performing ordinary full-batch GD compared to the tensorial scaled GD defined in the main text. The global learning rate for ordinary GD was taken to be $10/n$ so that the effective weight learning rate is the same as for scaled GD, c.f.\ Eq.~(\ref{eq:lambdadef}). Comparing Gaussian and orthogonal initializations for the same network architectures, the performance is nearly identical for $L/n \ll 1$ but orthogonal is clearly superior for $L/n \simeq 1$. On the other hand, comparing the two optimizers, there seems to be a clear advantage of ordinary GD compared to scaled GD for orthogonal initializations away from the infinite-width limit, while for $L/n \ll 1$, ordinary GD trains faster but does not achieve a marked improvement in best validation loss for either initialization. It is possible that the depth-independence of orthogonal NTK correlators reduces the importance of equal contributions to the NTK at each layer, which motivated the scaled GD prescription in Ref.~\cite{Roberts:2021fes} for Gaussian initializations. Relatedly, it is possible that having \emph{some} correlators growing with depth is in fact advantageous for feature learning, which may affect the optimal choice of $\eta$ for each optimizer and initialization. In future work, where we will compute the NTK correlators for orthogonal initializations with both GD choices, we will investigate this phenomenon further. Fig.~\ref{fig:lossesAdam} shows the result of using the \texttt{Adam} optimizer, which is much more common in practical applications than full-batch GD. To achieve reasonable performance, we changed the learning rate to $\eta = 10^{-3}$ (the default for \texttt{Adam}), and we caution that performance should not be directly compared with scaled GD. However, the comparison of the two initialization choices for the same optimizer shows once again that orthogonal outperforms Gaussian away from the infinite-width limit.

\begin{figure}[t!]
\centering
\textbf{Cross-entropy loss}\par
\includegraphics[width=0.45\textwidth]{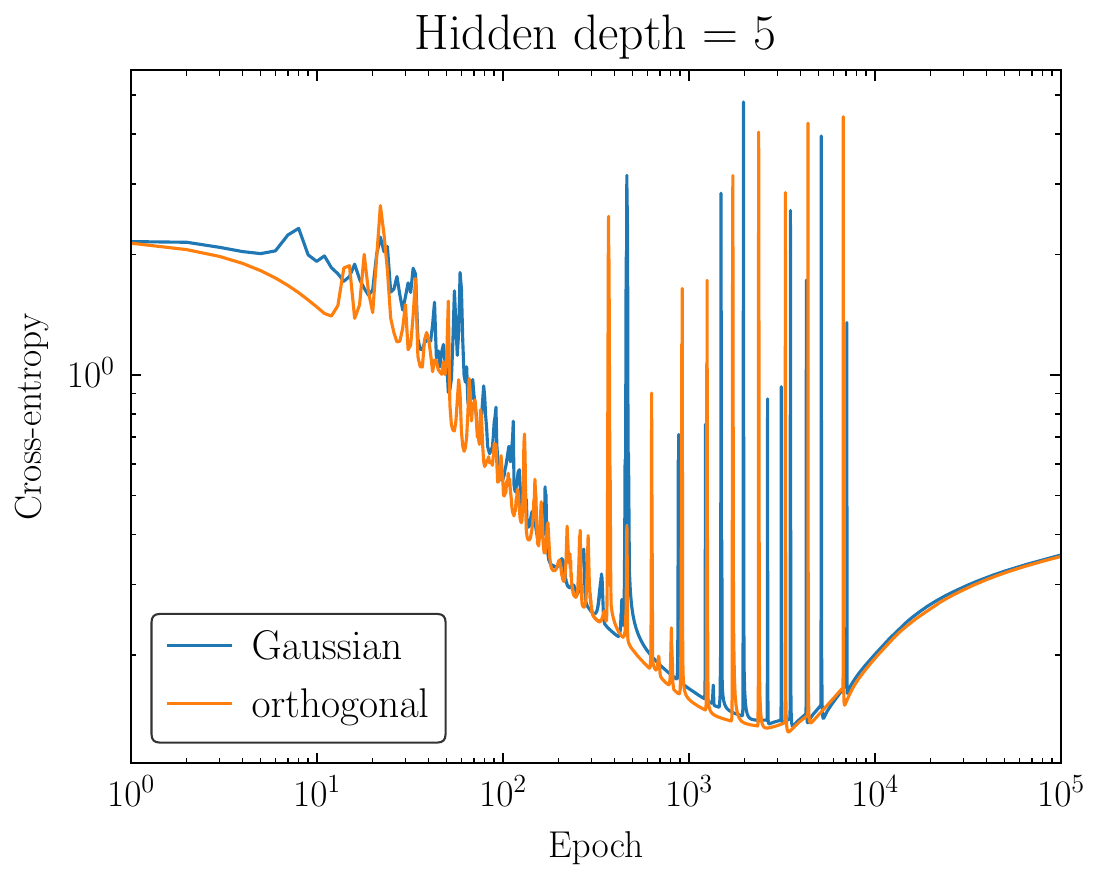}
\includegraphics[width=0.475\textwidth]{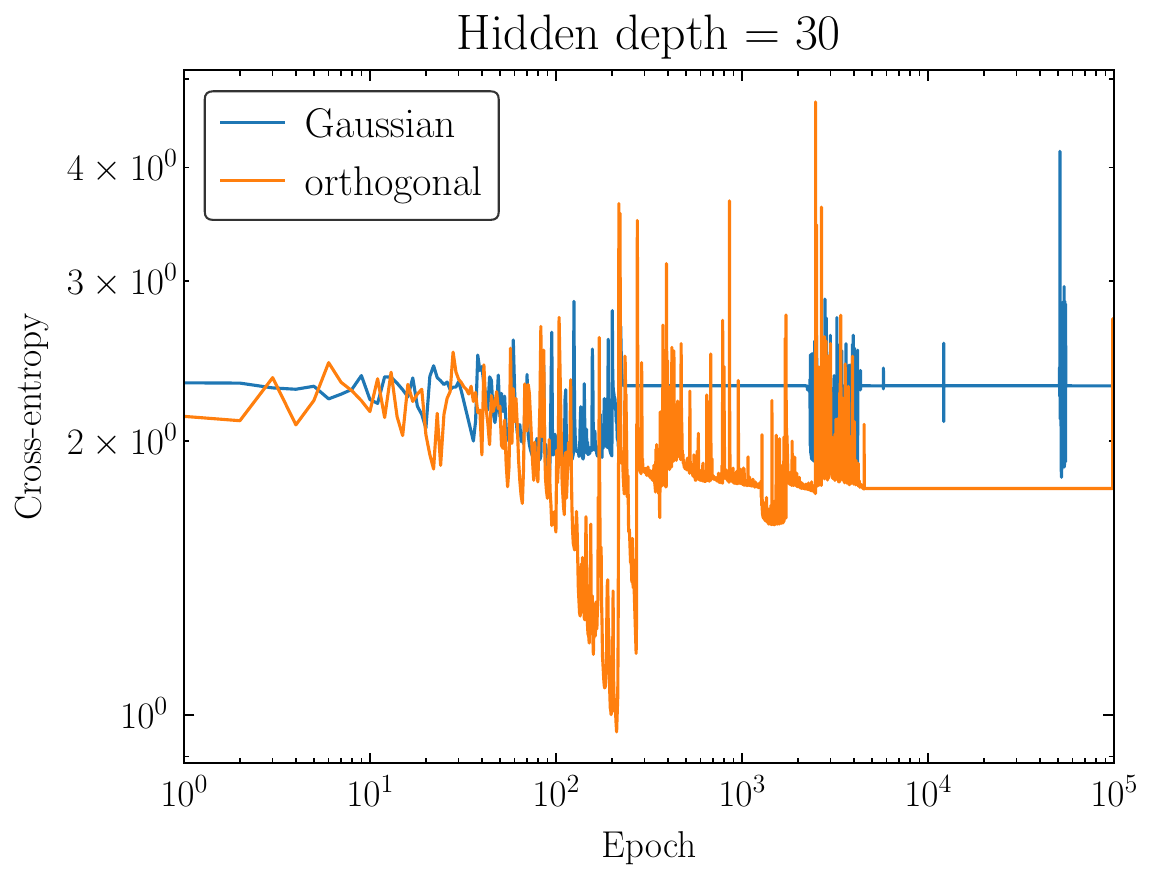}
\caption{Cross-entropy validation loss versus epoch for networks of width 30 and hidden depth 5 (left) and 30 (right), with tanh activation on MNIST data. Both initializations have nearly identical performance for depth 5, but orthogonal outperforms Gaussian in both training speed and generalization for depth 30.}
\label{fig:lossesCross}
\end{figure}

Finally, we consider the effect of changing the loss function to the more-typical cross-entropy loss paired with sigmoid activations on the output layer. Fig.~\ref{fig:lossesCross} (left) shows that Gaussian and orthogonal networks still have essentially identical performance in the $L/n \ll 1$ limit, though the GD oscillations are much more pronounced (as also noted in Ref.~\cite{cohen2021gradient}). Interestingly, the oscillations seem to damp out shortly after overfitting sets in, a phenomenon we intend to explore in future work. However, for $L/n \simeq 1$ (right panel), the Gaussian network completely fails to train, with a loss that only briefly surpasses that of a random 10-class classifier, $-\log(0.1) \approx 2.3$. The orthogonal network does train but overfitting occurs extremely fast, at epoch 200. Even so, the orthogonal network loss plateaus at a value below that of the Gaussian network, suggesting that somewhat better performance is still achieved with cross-entropy loss.

\end{appendix}

\vskip 0.2in

\bibliographystyle{JHEP}
\bibliography{OrthogonalInitBib}

\end{document}